\documentclass[11pt,letterpaper,english]{article}
\usepackage{microtype}
\usepackage{float}
\usepackage{amsmath} 
\usepackage{amsthm}
\usepackage{mathrsfs}
\usepackage{graphicx} 
\usepackage{caption}
\usepackage{setspace}
\usepackage{amssymb}
\usepackage{multirow}
\usepackage{rotating}
\usepackage{array}
\usepackage{booktabs}
\usepackage{multibib}
\usepackage{bibunits}
\usepackage{xr}
\usepackage{algorithm}
\usepackage{algpseudocode}
\usepackage{color}
\usepackage[small,compact]{titlesec} 
\DeclareMathAlphabet{\mathcalligra}{T1}{calligra}{m}{n}
\usepackage[sectionbib,round]{natbib}
\usepackage{mdframed}
\usepackage{tcolorbox}

\usepackage{listings}
\usepackage{geometry}
\usepackage{threeparttable}
\usepackage{tikz}
\usepackage{pgfplots}
\usepackage{tabularx} 
\usepackage{longtable} 
\usepackage{placeins}
\usepackage{url}

\usepackage{times}
\usepackage{abstract}

\usepackage[hang,flushmargin]{footmisc}
\usepackage{fullpage} 
\usepackage{hyperref}
\usepackage[compact]{titlesec}
\usepackage{epstopdf}
\usepackage{placeins}
\usepackage[toc,page]{appendix}
\usepackage{graphicx}
\usepackage{subcaption}
\usepackage{amsmath} 
\usepackage{amsthm}
\usepackage{graphicx} 
\usepackage{blkarray}
\usepackage{stackengine}
\usepackage{accents}
\usepackage{enumitem}
\usepackage[mathscr]{euscript}

\defaultbibliographystyle{plainnat}
\defaultbibliography{mybib}
\usepackage{tabularx}
\usepackage{footnote}

\hypersetup{
    colorlinks=true,      % false: boxed links; true: colored links
    linkcolor=MyDarkBlue,      % color of internal links
    citecolor=MyDarkBlue,      % color of links to bibliography
    filecolor=MyDarkBlue,      % color of file links
    urlcolor=MyDarkBlue        % color of external links
}
\definecolor{MyDarkBlue}{RGB}{158,0,0}

\makeatletter
\def\UrlAlphabet{%
      \do\a\do\b\do\c\do\d\do\e\do\f\do\g\do\h\do\i\do\j%
      \do\k\do\l\do\m\do\n\do\o\do\p\do\q\do\r\do\s\do\t%
      \do\u\do\v\do\w\do\x\do\y\do\z\do\A\do\B\do\C\do\D%
      \do\E\do\F\do\G\do\H\do\I\do\J\do\K\do\L\do\M\do\N%
      \do\O\do\P\do\Q\do\R\do\S\do\T\do\U\do\V\do\W\do\X%
      \do\Y\do\Z}
\def\UrlDigits{\do\1\do\2\do\3\do\4\do\5\do\6\do\7\do\8\do\9\do\0}
\g@addto@macro{\UrlBreaks}{\UrlOrds}
\g@addto@macro{\UrlBreaks}{\UrlAlphabet}
\g@addto@macro{\UrlBreaks}{\UrlDigits}
\makeatother

\theoremstyle{definition}

\newtheorem{corollary}{Corollary}

\newtheorem{prop}{Proposition}

\newtheorem{assumption}{Assumption}
\newtheorem{theorem}{Theorem}

\usepackage[utf8]{inputenc}
\usepackage{longtable}
\usepackage{graphicx}
\usepackage{epstopdf}
\usepackage{bbm}
\usepackage{dsfont}
\usepackage{comment}

\usepackage{xfrac}
\usepackage{xcolor}
\usepackage{siunitx}

\usepackage{pifont}% http://ctan.org/pkg/pifont

\usepackage{caption}
\newcommand{\quadstate}[1]{\Statex \quad #1}
\theoremstyle{plain}

\newcommand{\squishlist}{
   \begin{list}{$\bullet$}
    { \setlength{\itemsep}{0pt} \setlength{\parsep}{1pt}
      \setlength{\topsep}{1pt} \setlength{\partopsep}{1pt}
      \setlength{\leftmargin}{1.5em} \setlength{\labelwidth}{1em}
      \setlength{\labelsep}{0.5em} } }

\newcommand{\squishlisttwo}{
   \begin{list}{$\bullet$}
    { \setlength{\itemsep}{0pt} \setlength{\parsep}{0pt}
      \setlength{\topsep}{0pt} \setlength{\partopsep}{0pt}
      \setlength{\leftmargin}{1em} \setlength{\labelwidth}{1.5em}
      \setlength{\labelsep}{0.5em} } }

\newcommand{\squishend}{
    \end{list}  }

\title{Fair Document Valuation in LLM Summaries via Shapley Values}
\author{Zikun Ye \thanks{We thank Yizhuo Chang and Lei Wang for outstanding research assistance. We are also grateful to the participants of the Third Annual Research Roundtable on
Platform Dynamics, the WUSTL Junior Faculty Forum in Marketing 2025, the AIM Conference 2025, the 2025 INFORMS Marketing Science Conference, the UNC marketing seminar, and the MIT DSL seminar for feedback and comments, which have significantly improved this paper. We also thank Xiao Liu and Dennis Zhang for their detailed and thoughtful comments. Please address all correspondence to: zikunye@uw.edu and hemay@uw.edu.} \\ \textit{University of Washington} \and Hema Yoganarasimhan \\ \textit{University of Washington}}
\pgfplotsset{compat=1.18} 

\begin{document}

% ===== Response letter to the review team, compiled at the front of the combined PDF so it
% ===== can \ref the paper's labels. For the paper alone (final version), comment out the
% ===== next three lines.
% arXiv/SSRN public version: response letter removed (paper only).
% \input{letter}
% \clearpage
% \setcounter{page}{1}

\maketitle

\begin{abstract}
\begin{singlespace}

Large Language Models (LLMs) increasingly power search engines and AI assistants that retrieve and summarize content from many sources. By serving answers directly, these systems obscure the original content creators' contributions, threatening the compensation that sustains a healthy content ecosystem. We frame this as a problem of fair document valuation and compensation, and propose a framework based on the Shapley value. Because exact Shapley computation is prohibitively expensive at scale, we develop Cluster Shapley, an approximation that groups semantically similar documents via LLM embeddings and computes Shapley values at the cluster level, with formal bounds on both the approximation error and the induced revenue-attribution error. On Amazon product review data, off-the-shelf approximations such as Monte Carlo sampling and Kernel SHAP perform suboptimally in LLM settings, whereas Cluster Shapley substantially improves the efficiency--accuracy frontier. Simple attribution heuristics (e.g., equal or relevance-based allocation), though computationally cheap, yield highly unfair outcomes. Our approach is agnostic to the exact LLM used, the summarization process used, and the evaluation procedure, which makes it broadly applicable to a variety of summarization settings.

\end{singlespace}
\end{abstract}
\noindent \textbf{Keywords:} LLMs, Shapley Value, Digital Marketing, Search System, Retrieval Augmented Generation.

\thispagestyle{empty}
\newpage
\begin{bibunit}

\section{Introduction}
\label{sec:intro}

The advent of Large Language Models (LLMs) has reshaped how users search, process, and consume information. Unlike traditional search engines that return a list of links, LLM-powered search returns a concise summary with references to relevant documents, freeing users from manually navigating and aggregating multiple sources \citep{searchgpt}. To keep these summaries grounded in up-to-date content and to mitigate hallucination, such systems often use Retrieval-Augmented Generation (RAG) \citep{fan2024survey}: the system first retrieves the documents most relevant to a query and then uses them as context in the generative process. Thus, LLM-based summarization combines the generative strengths of LLMs with the retrieval strengths of search engines.

%The advent of Large Language Models (LLMs) has revolutionized how users search, process, and consume information. Today's LLM-based search and summarization engines combine the strengths of LLMs with those of traditional search engines. First, unlike traditional search engines that return a list of links in response to a query, LLM-powered search provides a concise summary with references to relevant documents/websites. This frees users from the cognitive load of manually navigating and aggregating information from multiple sources \citep{searchgpt}. Second, unlike regular LLMs that rely solely on static training data, LLM-based search engines augment their generative process with real-time retrieval and source grounding using a framework known as Retrieval-Augmented Generation (RAG) \citep{fan2024survey}. In the RAG framework, the system first retrieves a set of documents/articles that are most relevant to the query and then uses them as context in the generative process. This ensures that the LLM's responses are grounded in up-to-date, relevant content and do not suffer from limitations such as non-factual hallucinations and outdated knowledge that plague off-the-shelf LLMs. Thus, LLM-based search and summarization platforms combine the generative aspects of LLMs with the retrieval aspects of search engines by augmenting generation with context/information from documents most relevant to a query.

\begin{figure}[htbp]
    \centering
    \includegraphics[width=\textwidth]{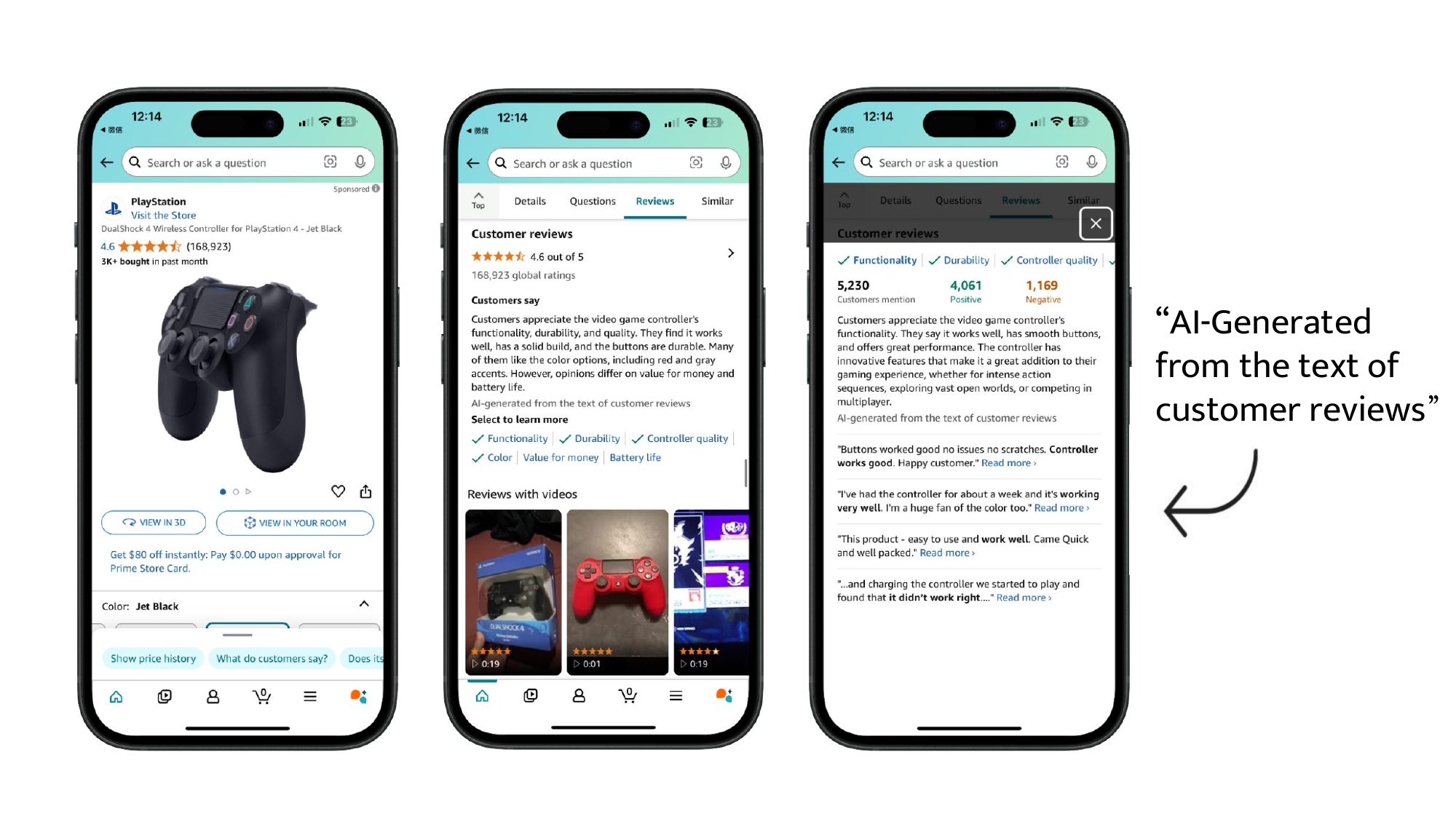}
    \caption{Amazon's AI-generated customer review for a \href{https://www.amazon.com/DualShock-Wireless-Controller-PlayStation-Black-4/dp/B01LWVX2RG}{wireless controller product} (snapshot taken on Dec 12, 2024): The left image shows the wireless controller product page on Amazon. The center image displays an Amazon AI-generated summary review of this product. Users can click ``Select to learn more'' to focus on specific aspects of interest. The right image shows AI-generated summaries for the selected aspect, displaying the source customer reviews with key information highlighted in bold.}
    \label{fig:amazon_review}
\end{figure}

This transformation is now ubiquitous: major search engines have integrated LLMs into their infrastructure (e.g., Microsoft's Bing AI \citep{microsoft2025copilotsearch} and Google's AI Overview \citep{google2025ai-overviews}), while new entrants such as OpenAI's ChatGPT Search \citep{searchgpt} and Perplexity AI \citep{perplexityai2025} have grown rapidly. The impact extends beyond web search to Q\&A sites (e.g., Reddit) and e-commerce platforms (e.g., Amazon, Best Buy). For example, Amazon now displays an LLM-based summary of a product's reviews on its product page, with the source reviews accessible on demand (Figure~\ref{fig:amazon_review}); consumers can also query Amazon's AI shopping assistant, Alexa for Shopping (formerly Rufus), for specific product information \citep{amazon_gen_ai_reviews, amazon_rufus, cnbc2026alexashopping}. Web Appendix~\S\ref{appsec:ai_generated_overviews} illustrates these applications.

%The main advantage of LLM-based search and summarization tools is that they simplify the search process for users, who no longer have to click on multiple links and aggregate information from individual sources. This can increase customer satisfaction and lead to higher platform usage \citep{xu2023chatgpt}. For example,  \cite{zhu2025generative} empirically shows that LLM-based search and summarization can significantly boost consumer purchases on an e-commerce platform through better articulation of consumer needs and product attributes. Furthermore, these approaches keep customers on the platform's own website or interface; that is, consumers don't need to leave the search engine and visit other websites to gather information. This, in turn, can lead to increased time spent on the platform, resulting in more eyeballs and potentially higher revenues \citep{Goodwin2024}. For these reasons, the rise of LLM-based search and summarization has generally been seen as beneficial to both platforms and consumers, at least in the short term. 

These tools benefit both users and platforms, at least in the short term: they simplify search and raise satisfaction and usage \citep{xu2023chatgpt, zhu2025generative}, and by keeping users on the platform rather than sending them to external sites, they increase time spent on the platform and potential revenues \citep{Goodwin2024}. This short-term benefit, however, masks a longer-term threat to the supply of content, which we frame as a \emph{compensation and incentive-design} problem. In a traditional search model, source documents (and their producers) gain traffic and reputation (both monetizable) by appearing in organic listings. LLM-powered search instead delivers a synthesized answer directly, letting users obtain information without visiting the sources. Website owners and recent research have accordingly reported losses in traffic and revenue following Google's AI Overview \citep{carroll2025aioverview, sherrer2025misleading, ipa2025google, khosravi2026impact}, and on review and Q\&A sites, where contributors earn reputation through badges, elite status, or perks,\footnote{For example, Reddit uses a ``Karma'' system capturing how helpful others found a user's responses \citep{reddit_karma}, and Yelp awards high-quality reviewers an ``Elite'' badge with associated perks \citep{chang2023yelp}.} reputation systems erode once users stop reading and rating individual contributions \citep{resnick_etal_2000}. The shift to LLM summaries thus threatens content creators' revenue streams and raises concerns about uncompensated use of their data.\footnote{Several major publishers such as {\it The New York Times} have begun to restrict or revoke AI access to their content, citing undervaluation and uncompensated use \citep{nyt_openai_lawsuit_2023}.} Absent fair compensation, creators have no incentive to supply content, degrading summary quality over time. Thus, while avoiding attribution may benefit platforms in the short run, it is not a sustainable equilibrium.

As a first step, AI firms have begun striking licensing deals with large content creators; e.g., OpenAI's deal with Murdoch-owned publications such as {\it The Wall Street Journal} and {\it The New York Post} \citep{robertson2024openai}. Such deals, however, scale poorly: they work for a few large publishers but not for the many small, niche contributors with whom individual contracting is too costly. A complementary line of industry practice instead applies simple heuristic attribution rules that are easy to implement at scale: for instance, splitting value equally across the sources of a summary, weighting sources by their query--document relevance scores from the retrieval step, or charging per scrape request (as in Cloudflare's pay-per-crawl marketplace~\citep{cloudflare2025}). While attractive for their simplicity, these rules share a basic flaw: they pay for visibility, volume, or access rather than for the value a document actually adds to the summary. An equal split ignores differences in usefulness across sources; relevance weighting scores each document's topical match in isolation, so it rewards redundant copies of the same information; and pay-per-crawl schemes charge for retrieval regardless of whether the content shapes the answer. Indeed, we show empirically that these rules deviate substantially from fair valuations ($\S$\ref{ssec:simple rule}).

Valuing a document is intrinsically hard: its contribution to a summary depends not only on its own relevance, quality, and reliability, but also on how unique its information is relative to the other retrieved documents, i.e., its marginal contribution. Further, a principled valuation must therefore capture a document's contribution both {\it within} a query and, when aggregating across queries, {\it across} them (e.g., one document may be essential for a niche query while another is moderately useful for a popular one). Finally, for a valuation approach to be broadly applicable and practical, it must satisfy three properties: it should be (1) summarization-procedure agnostic, since summarization LLMs and techniques vary across platforms and evolve over time; (2) evaluation-process agnostic, since platforms score summary usefulness differently (e.g., implicit dwell-time signals on search engines versus explicit helpfulness ratings on review and Q\&A sites); and (3) scalable and cost-effective, so that it can run on real platforms without prohibitive time or money costs.

This motivates the two research questions we study. First, \emph{how can a platform fairly value the contribution of each source document to an LLM-generated summary, in a way that rewards genuine marginal contribution rather than visibility or volume?} Second, \emph{how can such valuations be computed efficiently enough to scale to platforms serving large query volumes?} The first question concerns the economic fairness of the resulting compensation; the second, its practical feasibility.

In this paper, we present a framework for {\it fair document valuation} in the context of LLM summaries, building on the Shapley value from cooperative game theory \citep{shapley1953value}. Our framework addresses the challenges discussed and satisfies the three properties above. Given a query $q$ and a summary built from a set of documents, we use the Shapley value to distribute the summary's total value across those documents according to each one's marginal contribution, accounting for redundancy and overlapping information. It aligns compensation with the genuine value a document adds. Because the Shapley computation treats the summarization $A(\cdot)$ and evaluation $v(\cdot)$ procedures as black boxes, it is summarization- and evaluation-agnostic and its linearity property lets per-query values aggregate across queries, accommodating both per-query revenue (e.g., advertising, API calls) and aggregated subscription revenue.

However, a major hurdle in applying Shapley values to LLM summaries is its exponential computational complexity -- if we have $|S_q|$ relevant documents for query $q$, the number of summarizations and evaluations needed to calculate Shapley values is $2^{|S_q|}-1$. Even for modest-scale platforms that process millions of queries, this becomes computationally infeasible in practice. To address this problem, many approximation algorithms have been proposed, such as Monte Carlo \citep{mann1960values}, Truncated Monte Carlo \citep{ghorbani2019data}, and Kernel SHAP \citep{lundberg2017unified}. However, all these algorithms treat each document as an independent unit and do not leverage the textual information of documents. As a result, their empirical performance in LLM summarization settings is often suboptimal.

To solve this computation issue, we develop a simple yet effective approximation algorithm, Cluster Shapley, that leverages textual content for efficiency. The intuition is that documents with similar content make comparable contributions and should receive similar Shapley values. Instead of treating documents as independent, the algorithm uses LLM-generated embeddings to group semantically similar documents into clusters, treats each cluster as a single meta-document in the Shapley calculation, and distributes each cluster's value evenly across its documents, sharply reducing the number of combinations the LLM must summarize and evaluate at limited accuracy cost. These clusters are \emph{query-specific}: they are formed over the retrieved set $S_q$ of each query, so two documents grouped together for one query may fall into different clusters for another (indeed, a document may not even belong to $S_q$ for the other query), and the clustering is recomputed per query rather than fixed globally. A tunable clustering diameter $\epsilon$ trades off accuracy against cost: smaller $\epsilon$ yields more accurate but slower computation, larger $\epsilon$ the reverse. We also establish theoretical guarantees for Cluster Shapley. The algorithm reduces complexity from $O(2^n)$ to $O(n^2+C_{\mathcal M}(m))$, where $n=|S_q|$ is the number of retrieved documents, $m$ is the number of query-specific clusters, and $C_{\mathcal M}(m)$ is the cost of the cluster-level Shapley procedure; exact cluster-level computation gives $C_{\mathcal M}(m)=O(2^m)$, while Monte Carlo gives $C_{\mathcal M}(m)=O(Nm)$. We derive a general oracle-style bound that separates the deterministic clustering/coarsening bias from any cluster-level numerical approximation error, and the approximation is exact when the partition is refined to singleton documents; under an additional Lipschitz condition, the deterministic term admits an interpretable specialization in terms of within-cluster dispersion and meta-document aggregation. We also translate the bound into a corresponding bound on the revenue-attribution error. Together, these results show that Cluster Shapley is a principled, scalable solution deployable even at large scale.

We demonstrate our algorithm on products from the Amazon product review dataset \citep{hou2024bridging}, which closely mirrors real-world use since Amazon's AI review summaries are built on a similar repository. We use 24 products from diverse categories and, for each, construct two queries from frequently mentioned attributes to mimic real consumer information needs. Each query is answered via RAG: the eight most relevant reviews are retrieved by query--review embedding similarity, and the LLM generates a query-specific summary from this set. We benchmark Cluster Shapley against the two simple attribution rules introduced above (equal and relevance-weighted attribution) and three widely used approximations: Monte Carlo \citep{mann1960values}, Truncated Monte Carlo \citep{ghorbani2019data}, and Kernel SHAP \citep{lundberg2017unified}. Two insights emerge. (1) Off-the-shelf approximations are far from optimal here: designed for independent contributors, they ignore the semantic structure of text and deliver poor efficiency--accuracy trade-offs, whereas Cluster Shapley markedly improves the efficient frontier (reaching a given accuracy with significantly fewer evaluations). This points to a broader opportunity for structure-aware Shapley approximations in LLM applications. (2) Simple heuristic rules are highly biased: equal attribution yields an MAPE above 150\% and relevance-weighted attribution above 30\%, large deviations from the fair Shapley baseline that confirm cheap heuristics can substantially distort document values. Together, these findings show that structure-aware approximations make fairer attribution both feasible and affordable.

In summary, our paper makes two core contributions. First, we address the important and largely unsolved problem of source document valuation in the context of LLM-generated summaries. By leveraging the Shapley value framework, we propose a principled approach for fair document valuation. To the best of our knowledge, this is among the first applications of Shapley values to LLM-based summarization systems. Second, we introduce the Cluster Shapley algorithm, which improves the computational efficiency of Shapley value approximation by exploiting semantic similarity among documents. We provide theoretical guarantees on its performance and evaluate its empirical effectiveness using real-world data, showing that it substantially improves the efficiency–accuracy frontier. %Third, our results open up a new research direction by demonstrating that Shapley approximation algorithms can be significantly improved in LLM applications when additional structure, such as embedding-based similarity, is available. This highlights the potential for developing a broader class of structure-aware Shapley approximation algorithms for LLM applications. Fourth, we show that simple attribution rules, though computationally cheap, can lead to substantial unfair outcomes. This insight provides practical guidance for decision-makers: platforms relying on simplistic heuristics may systematically undervalue or overvalue individual contributors, underscoring the importance of adopting fair and theoretically grounded attribution frameworks.

\section{Related Literature}
\label{sec:related_lit}

Our research relates to and contributes to multiple streams of work, including game theory, LLMs, and business applications of AI systems. 

%First, our work relates to the growing literature on LLM-based summarization. Recent research on summarization has shown that summaries produced by LLMs like GPT-4 achieve comparable or superior factual accuracy, coherence, and overall quality compared to human annotators in news summarization tasks \citep{pu2023summarization}.  Indeed, text summarization has evolved from traditional extractive methods like TextRank \citep{mihalcea2004textrank} to more sophisticated abstractive approaches powered by LLMs. RAG frameworks \citep{lewis2020retrieval}, which combine information retrieval with LLM generation, have shown significant improvements in factual accuracy and information currency \citep{gao2023retrieval, jiang-etal-2023-active}. Recent developments include GraphRAG \citep{edge2024local}, which enhances retrieval performance using graph-based representations. While summarization technologies continue to evolve rapidly, our work addresses the fundamental challenge of document valuation that persists across summarization approaches. The document valuation framework we develop is designed to be agnostic to specific summarization techniques, ensuring its applicability even as LLM and RAG technologies advance. \hy{Is this para necessary?}

First, our work builds on the game-theoretic concept of Shapley values \citep{shapley1953value}, which has gained significant traction in machine learning for quantifying feature importance \citep{lundberg2017unified} and data valuation \citep{jia2019towards, ghorbani2019data}.\footnote{Apart from the Shapley value approach, alternative approaches for calculating the value of contributors in cooperative tasks have also been proposed. Some of the well-known approaches include the leave-one-out approach \citep{cook1977detection} and the influence function approach \citep{barshan2020relatif, han2020explaining, guo2020fastif}. However, these methods, neglecting high-order interactions, may result in undesirable data valuation. For example, when there are two or more similar contributors, these methods tend to assign nearly zero value to similar documents. For example, consider a scenario where we have three documents, A, B, and C, where A and B have the same high-quality content and C contributes nothing to the overall value. Intuitively, we should expect A and B to have the same valuation and C to have zero value. However, the leave-one-out approach will assign zero value to both A and B -- if we leave A out and calculate the value generated by B and C, the value would almost be the same as that with A (and vice-versa for B). Thus, if we believe this approach and assume that both A and B have no incremental value and drop them both from the set of contributors, the overall value generated by the set will go to zero! In sum, these approaches are unable to correctly account for the marginal contributions of each player. This is the reason why Shapley has emerged as the dominant paradigm for valuing contributors in cooperative settings.} While prior studies have applied Shapley values primarily to supervised learning tasks and feature attribution, our work is among the first to apply Shapley values to document valuation in LLM-based summarization systems. Our work also relates to the work on approximation algorithms for computing Shapley values since it is well-known that computing exact Shapley values is computationally challenging. Without additional structural assumptions on the value function beyond bounded range, any unbiased estimator must rely on sampling permutations or coalitions; simple Monte Carlo (random-permutation) estimators are unbiased and achieve concentration rates that match information-theoretic lower bounds up to constants \citep{castro2009polynomial,maleki2013bounding}. Specifically, with value range bounded by $[a,b]$, Hoeffding-style arguments yield a sample complexity of $O\!\left(\frac{(b-a)^2}{\varepsilon^2}\log\frac{1}{\delta}\right)$ to obtain an $\varepsilon$-accurate estimate with probability $1-\delta$, and \citet{maleki2013bounding} give matching lower bounds. Alternative speedups such as linear regression surrogates, kernel/coalitional weighting, or grouping assumptions can reduce computation but introduce modeling assumptions that may bias attribution when violated \citep{fatima2008linear,bachrach2010approximating,ghorbani2019data}. 

In our proposed Cluster Shapley algorithm, we explicitly account for the semantic similarity between documents, which allows us to achieve high accuracy at relatively low computation costs. Relative to the approximation literature above, existing methods fall into two broad families. The first consists of general-purpose, \emph{structure-unaware} approximations, such as Monte Carlo Shapley \citep{mann1960values, castro2009polynomial} and Kernel SHAP \citep{lundberg2017unified}, which treat documents as independent units and ignore their semantic content. The second consists of structure-exploiting methods that impose additional assumptions: pre-defined Group Shapley \citep[Section~3.2]{ghorbani2019data} exploits grouping but supplies no approximation-error guarantee; \citet{corder2019shapley} combine a hierarchical clustering approach with exact Shapley calculations under restrictive assumptions (e.g., K-means with two clusters); and, most recently, \citet{patel2025maxshapley} study Shapley-based attribution for generative search, where their MaxShapley algorithm assumes a decomposable (max-sum) utility structure to obtain polynomial-time attribution. Cluster Shapley combines the strengths of both families: it makes no structural assumption on the value function, treating the summarization and evaluation pipeline as a black box; it forms \emph{query-specific} groups from LLM embeddings, exploiting semantic structure; and it provides a formal error decomposition (Theorem~\ref{thm:cluster_shap_general}) that delivers a principled, tunable accuracy--efficiency trade-off. To our knowledge, we are the first to use LLM-generated text embeddings to guide the clustering, which better captures semantic relationships and is especially suited to our task. We expect that our general idea of using a clustering approach to capture the similarity between documents/observations in Shapley calculations can be extended beyond LLM summarization settings and applied more broadly to other cases where similarity measures between observations exist.

More broadly, recent works such as \citet{wang2024economic} have used the notion of Shapley value to quantify the value of large corpora to pretraining by approximating Shapley values through repeated fine-tuning (an expensive and often infeasible process). In contrast, our method targets applications such as LLM-driven summarization tasks, where the model is fixed and attribution occurs at inference time. Because our approach approximates Shapley values based on LLM-generated embeddings, it avoids retraining and scales more efficiently with the number of input documents. This distinction is crucial: \citet{wang2024economic} deals with corpus-level attribution during LLM training; we focus on attribution over documents presented as context to an already-trained LLM, making our method more practical for interactive, high-frequency applications. Relatedly, \citet{wang2025datashapley} propose In-Run Data Shapley, which computes data Shapley values within a single training run via gradient-based accumulation, thereby avoiding repeated retraining. As with \citet{wang2024economic}, their focus is training-time data attribution, whereas our method attributes value to documents supplied as context to a fixed model at inference time. In marketing, \citet{mohammadi2024wait} also applies Shapley values to LLMs, but for interpretability: attributing an LLM's output to individual prompt components to diagnose sensitivity to semantically irrelevant tokens. Our goal is instead economic -- attributing the value of a summary to its retrieved source documents to support compensation at scale.

Our work relates to the literature on news aggregators in marketing and economics. News aggregators function similarly to AI-based summarization, since users may consume content directly on the aggregator’s platform without visiting the publishers’ websites. A large stream of research has examined whether this substitutive effect reduces traffic to publishers or whether aggregators can also serve as a discovery channel that increases exposure using game theoretical models \citep{mayzlin2012link, dellarocas2013media, jeon2016news, amaldoss2023can}. \citet{calzada2020news} and \citet{athey2021impact} look at this question empirically and both find that news aggregators can exert a market-expansion effect, increasing visits to news outlets, especially for smaller publishers. 

%\citet{song2023does} empirically examine the effects of carrying news on user engagement with non-news content on social media. 

Relatedly, there is also a growing literature on the creator economy, which examines how platforms aggregate independent content providers through bundling and revenue-sharing mechanisms. For instance, \citet{bhargava2022creator} develops a model of platform design, demonstrating that revenue sharing is not merely a zero-sum game between platforms and creators; rather, an optimal sharing rule effectively incentivizes ecosystem supply and enhances total welfare. Complementing this, recent studies on bundle pricing \citep{zhang2024if} suggest that aggregating high-variety goods can reduce demand volatility and improve monetization efficiency. Recent empirical work by \citet{wu2022competition} highlights the critical role of incentive design and demonstrates that performance-based revenue-sharing contracts, as opposed to fixed compensation, significantly boost novelty. Given that AI-based summarization is functionally similar to content aggregators and bundling, our Cluster Shapley approach for valuing source documents can also be used by content aggregators and newspapers/content websites to formalize revenue-sharing arrangements.

\section{Problem Definition}
\label{sec:problem_def}
We define the problem from the perspective of a platform that has access to $D$ original documents generated by different producers. We do not make any distributional assumptions on $D$. These documents are not necessarily i.i.d. and may contain overlapping information and vary in the quantity and quality of content. Users arrive at the platform and query for some information from the platform using queries $q$, drawn from a distribution $g(q)$. The platform generates a response to each query $q$ based on the $D$ documents using an LLM-based summarization model $A(q, D)$. We can view ${A(q, D)}$ as a black box that takes a dataset $D$ of any finite size to generate a summary in response to query $q$. Note that in practice, the platform may choose to only use a subset of documents ($S_q \subseteq D$) that are most relevant to the query for the summarization process; that is, we allow for cases where all documents are not relevant to all queries. In such cases,  the summarization process is denoted by $A(q, S_q)$.

The quality or performance of a summarization is denoted by $v(q, A(q, S_q))$. Intuitively, this score captures the extent to which the user finds the summary useful or valuable. The performance score $v$ can be treated as a black-box that takes the query and summary as input and returns a score. In practice, $v(q, A(q, S_q))$ can be obtained in a multitude of ways. It could be actual scores collected from user surveys on how helpful they find a given summary to be (e.g., rating of helpfulness, fraction of upvotes). Alternatively, it could be helpfulness scores based on an LLM model, where an independent LLM agent does the scoring instead of human agents. This can be a viable option in settings where collecting user responses is costly and/or slow; indeed, recent research has shown that LLM ratings tend to align with user ratings in many situations \citep{kang2023llms, cheng_etal_2025}. It could also represent implicit helpfulness scores based on user behavior, which are commonly used in the information retrieval and search literature to measure the relevance of a given document/link, e.g., whether the user clicked on the summary, the time spent reading the summary \citep{liu_2009, yoganarasimhan_2020}.

The platform's goal is to determine how much each document \( i \in D \) contributes to the quality of summaries produced for user queries. Because both the summarization model \( A(\cdot) \) and the evaluation method \( v(\cdot) \) are fully controlled by the platform, content providers cannot directly influence the generation or evaluation of outputs. This limits direct manipulation of the valuation procedure itself, though strategic content creation and retrieval-stage gaming remain broader platform-design concerns that we discuss below.

%The platform's goal is to derive an equitable valuation for each document $i \in D$. In our setting, summarization and evaluation are fully under the platform's control, and content providers cannot manipulate these processes. Let $\phi_i(D, q, A(q, S_q), v(q, A(q, S_q))) \in \mathbb{R}$ denote the value of document $i$ in dataset $D$, for query $q$, given summarization $A(q, S_q)$ and score $v(q, A(q, S_q))$. Then, we can write the value of each document $i$ over all the queries as:
%\begin{equation}
%\rho_i\left(D, A(\cdot), v(\cdot), g(\cdot)\right) = \int \phi_i\left(D, q, A(q, S_q), v(q, A(q, S_q)\right) g(q) dq ,
%\end{equation}
%where the value of each document $i$ for a given query $q$ is integrated over the distribution of queries $g(q)$. 

We define \( \phi_i(q,S_q, A, v) \) as the value of document \( i \) for a given query \( q \). This value function depends on the subset of documents retrieved for query \( q \) (i.e., \( S_q \subseteq D \)), the summary function of documents \( A(q, S) \), and the resulting performance score function \( v(q, A(q, S)) \). Throughout the discussion, we use the shorthand $\phi_i(q) = \phi_i(q,S_q, A, v)$ whenever the context is clear. This setup allows the platform to aggregate document values across queries while respecting both the relative importance of documents and the frequency of the queries they support. For instance, a document that is essential for answering a rare query may be less valuable overall than a document that provides moderate value across many common queries.

We argue that it is important to conduct value attribution at the query level, especially for small publishers and niche queries. First, the economic impact of LLM-generated answers varies substantially across queries and sources. Second, large publishers can negotiate direct licensing agreements with LLM providers, but small or niche sites, whose value lies in specific, long-tail queries, cannot feasibly do so at scale. A query-level approach also ensures that marginal contributions are measured in the exact contexts where the content is used, leading to a more accurate and fair valuation. Thus, our goal is to develop a query-level document valuation approach that satisfies three properties:
\squishlist
\item Summarization Procedure Agnostic: The approach should be agnostic to the specifics of the summarization process, $A(\cdot)$, used by the platform. That is, it should generalize across RAG systems.  While we present a standard RAG implementation in $\S$\ref{ssec:summarization}, numerous alternatives exist, from simpler methods to more sophisticated frameworks like TextRank, GraphRAG, and DRAG \citep{mihalcea2004textrank, edge2024local, zhang2025lexical}. As LLM technologies rapidly evolve, our document valuation framework is designed to remain effective, regardless of advancements in summarization, ensuring broad applicability across current and future implementations.

\item Evaluation Process Agnostic: The approach should apply to any evaluation method, $v(\cdot)$. As discussed earlier, many explicit and implicit approaches for scoring summaries exist. Different business use cases may have access to different evaluation approaches. For example, search engines (e.g., Perplexity or Google) usually only have implicit feedback/evaluation, whereas question-answering websites or review aggregators may have more explicit feedback on the helpfulness of reviews. We would like our algorithm to be agnostic to the exact approach used. In our empirical context, we use a prompt-based LLM approach for evaluation; see $\S$\ref{ssec:eval_step}.

\item Scalable and Cost-Effective: The approach should scale to platforms that serve large query volumes without incurring prohibitive computational time or cost. This property is essential for practical deployment.

\squishend

\section{Solution Concept: Shapley Value for Document Valuation}
\label{sec:shapley_framework}

This section develops our document valuation framework. We introduce the Shapley value and its relevance to document valuation ($\S$\ref{ssec:shapley_intro}), apply it to revenue attribution ($\S$\ref{ssec:revenue_attribution}), and, because exact Shapley computation is intractable at scale, develop an efficient approximation, Cluster Shapley ($\S$\ref{ssec:cluster_shapley}), with theoretical guarantees on its approximation error and computational complexity ($\S$\ref{ssec:theory}).

\subsection{Shapley Value}
\label{ssec:shapley_intro}

The Shapley value $\phi_i(q)$ is the unique document-valuation rule satisfying four standard axioms \citep{shapley1953value}: \emph{Efficiency} (the summary's total value is fully distributed across the documents), \emph{Symmetry} (documents that contribute equally to every coalition are valued equally), \emph{Null Document} (a document with no marginal contribution receives zero value, so that $\phi_i(q)=0$ for any $i \notin S_q$), and \emph{Linearity} (per-query values are additive, so attributions can be aggregated across queries, which is central to the revenue application in $\S$\ref{ssec:revenue_attribution}). Because these axioms are standard, we state their formal characterization and adaptation to our setting in Web Appendix~\S\ref{appsec:axioms}.

%Based on $\S$\ref{sec:problem_def}, recall that our goal is to find a document valuation function $\rho_i\left(D, A(\cdot), v(\cdot), g(\cdot)\right) \in \mathbb{R}$ to quantify the value of document $i$ in set $D$. To obtain this valuation, we need to first accurately estimate the query-level document valuation function $\phi_i\left(D, q, A(q, S_q), v(q, A(q, S_q)\right)$. Henceforth, we denote this value function as $\phi_i(q)$ because the retrieval process $S_q$, the LLM-based summarization process $A(q, S_q)$, and the performance score function $v(q, A(q, S_q))$ are all uniquely defined by $q$.

The Shapley value \( \phi_i(q) \) for a document \( i \in S_q \subseteq D \) can be written as the expected marginal contribution of document \( i \) across all possible coalitions:
\begin{equation}
\label{eqn:shapley}
    \phi_i(q) = \frac{1}{|S_q|} \sum_{S \subseteq S_q \setminus \{i\}} \frac{v(q, A(q, S \cup \{i\})) - v(q, A(q,S))}{ \binom{|S_q|-1}{|S|}}.
\end{equation}
This can be stated equivalently as $ \phi_i(q) = \frac{1}{|S_q|!} \sum_{\pi \in \Pi(S_q)} \left[ v(q,A(q,P^\pi_i \cup \{i\})) - v(q, A(q,P^\pi_i)) \right]$
where $\pi\in \Pi(S_q)$ is a permutation of $S_q$, and $P^\pi_i$ is the set of documents which precede document $i$ in the permutation $\pi$. 
%\begin{equation}
%\label{eqn:shapley_permu}\end{equation}

Note that for $i \notin S_q$, the Shapley value of $i$ is $\phi_i(q) = 0$ because only documents in $S_q$ are used for summarization. Documents outside $S_q$ have no marginal contribution and thus receive a Shapley value of zero. By simple calculation, the Shapley value formula can thus be equivalently written over the original document set $D$ as:
\begin{equation}
\phi_i(q) = \frac{1}{|D|} \sum_{S \subseteq D \setminus \{i\}} \frac{v(q, A(q, S \cup \{i\})) - v(q, A(q, S))}{\binom{|D|-1}{|S|}}.
\end{equation}
%However, many of these evaluation score calculations are redundant since any document outside $S_q$ does not increase the performance score. In fact, this formula defined on $D$ is equivalent to the formula defined only on $S_q$ in Equation~\eqref{eqn:shapley}. 
The equivalence follows directly from the permutation-based definition, as all permutations of $D \setminus S_q$ do not affect the performance score.\footnote{$\phi_i(q)$ is document $i$'s marginal contribution and can in principle be negative if a document degrades the summary; our framework handles this directly, since the computed value is simply the document's actual contribution. In practice such negative values are rare, and a negative value signals that the document should not have been retrieved for this query. A platform can therefore handle it with a simple rule, e.g., removing the document from that query's retrieved set $S_q$ (so it receives no payout for that query). By Linearity, an isolated negative value also averages against the document's positive contributions across other queries.}

We now discuss a few points related to the Shapley attribution framework as applied to our setting:
\squishlist

\item \textbf{Comparison to simple heuristics.} Our query-level Shapley allocation rule offers several advantages compared to naive/simple heuristics such as splitting the value equally among all relevant documents in $S_q$ or relevance-weighting, where the value of a document is proportional to its relevance to the query. These kinds of heuristics largely reward \emph{visibility and volume}: a prolific contributor who floods the platform with similar documents is rewarded as much as, or more than, a niche expert. In contrast, query-level Shapley rewards \emph{contextual usefulness}: a document's value depends on what it adds given the other documents retrieved for that specific query. Moreover, unlike these heuristics, query-level Shapley \emph{benefits the long tail}: a niche document that is the sole source able to answer a rare query receives high value for that query, even if its creator contributes few documents overall. Finally, our approach also \emph{penalizes redundancy}: when many documents carry the same information, each has low marginal contribution and hence low value, so creators are incentivized to produce unique rather than duplicative content. Together, these properties steer the ecosystem toward the kind of original, high-quality contributions that purely volume- or visibility-based rules fail to reward. We formally show the performance of such heuristics in our empirical setting in $\S$\ref{ssec:simple rule}.

\item \textbf{Extension to Multiple Evaluation Metrics.} Platforms often assess summaries along several dimensions at once (for example, information coverage, factual accuracy, or engagement signals such as dwell time and likes). Our framework extends to this setting directly. Each metric $m$ defines its own value function $v^{(m)}(q, A(q, S))$ and hence its own per-document Shapley value $\phi_i^{(m)}(q)$; by the linearity property ($\S$\ref{ssec:shapley_intro}), these can be combined into an overall valuation $\sum_m w_m\, \phi_i^{(m)}(q)$ under any platform-chosen weights $w_m$. Computationally, the clustering step of our Cluster Shapley algorithm depends only on the document embeddings and is therefore shared across all metrics; only the cluster-level Shapley computation is repeated per metric. Each metric still incurs its own evaluation cost (e.g., a separate LLM judge, human rating, or behavioral signal), but the expensive clustering step is performed once.

\item \textbf{Gaming concerns.} A natural concern in attribution systems is whether document providers can strategically manipulate their content to inflate their assigned value. This issue is not new: even in traditional search engines like Google, publishers have long engaged in ``mimicking the winner'' strategies, which create content that resembles highly ranked pages to improve visibility. While such behavior is theoretically possible, it is difficult to execute in practice. Modern LLM-powered platforms (e.g., Perplexity, ChatGPT, Google’s AI Overview) do not disclose their specific RAG systems to external document providers. Moreover, these systems do not present explicit ranked lists but instead synthesize information into summaries, making it harder for publishers to infer their position or influence outcomes. Retrieval in such systems is governed by proprietary, black-box algorithms that incorporate a range of dynamic signals, including semantic embeddings, recency, diversity, and user engagement. This level of opacity is a deliberate design choice, observed not only in LLM systems but also in recommender platforms and digital advertising ecosystems, where hiding algorithmic details is a known strategy to deter gaming \citep{kroll_etal_2017, Sax_Wang_2025}.

\squishend

Finally, relating this framework to the three ideal properties of $\S$\ref{sec:problem_def}: because the Shapley value in Equation~\eqref{eqn:shapley} treats the summarization $A(\cdot)$ and evaluation $v(\cdot)$ procedures as black boxes, it is summarization- and evaluation-agnostic by construction, satisfying the first two properties. It does not, however, satisfy the third, scalability -- a computational challenge we detail and address in $\S$\ref{ssec:cluster_shapley}.\footnote{It is important to note that the Shapley framework offers one axiomatic view of fairness and uses it for value attribution. Nevertheless, it is not the only notion of fairness. Other ways to quantify value/fairness exist, e.g., economic value, which can be a function of not just the value generated for the platform, but also the cost associated with creating the content.}

In sum, our query-level Shapley value framework offers a transparent and theoretically grounded mechanism for sharing platform revenue with document providers. 

\subsection{Revenue Attribution}
\label{ssec:revenue_attribution}

An important application of our Shapley-based document valuation framework is enabling fair and economically grounded revenue sharing between platforms and document owners. In practice, platforms generate revenue either at the {\it query level}, for instance through per-query advertising or API charges, or through {\it subscription-based models} that aggregate revenue over a large number of queries. Our framework can accommodate both settings and provide a principled method for revenue allocation based on each document’s marginal contribution to the LLM-generated output.

To illustrate how such a framework can be operationalized in practice, consider platforms running LLM-based summarization services. Such platforms can determine a revenue share rate $\beta \in [0,1]$ for distribution among documents while retaining $1-\beta$ as platform revenue. For example, ProRata.AI provides an attribution solution in AI search and shares revenue 50/50 with its sources~\citep{prorata2025}. However, it uses a proprietary algorithmic approach to score and determine attribution~\citep{prorataUMG2025}, the details of which are not publicly known.

\noindent \textbf{Subscription-Based Revenue.}  
Let queries be drawn from a distribution \( g(q) \), and let the total subscription revenue over a billing cycle (e.g., a month) be denoted by \( R \). Then, the total value of document \( i \) over the query distribution can be expressed as the expectation:
\[
\mathbb{E}_{q \sim g(q)}[\phi_i(q)].
\]
The platform can use this  to allocate revenue proportionally among documents. Specifically, document \( i \)'s payout is given by:
\[
\beta \cdot R \cdot \frac{\mathbb{E}_{q \sim g(q)}[\phi_i(q)]}{\sum_{j \in D} \mathbb{E}_{q \sim g(q)}[\phi_j(q)]}.
\]
This ensures that documents are compensated according to their aggregate value across the full set of queries served, aligning economic incentives with contribution.

\noindent \textbf{Query-Level Revenue.}  
In settings where revenue is generated on a per-query basis, such as through API billing or advertising impressions, the allocation is straightforward. For a given query \( q \) that yields revenue \( r_q \), the payment to document \( i \) is:
\[
\beta \cdot r_q \cdot \frac{\phi_i(q)}{\sum_{j \in S_q} \phi_j(q)}.
\]

\noindent \textbf{Combined Revenue.} When both subscription-based and query-level revenue are present, the total expected payout to document $i$ is:

\[
\text{Total Payout}_i = \beta \cdot \bigg( R \cdot \frac{\mathbb{E}_{q \sim g(q)}[\phi_i(q)]}{\sum_{j \in D} \mathbb{E}_{q \sim g(q)}[\phi_j(q)]} + \mathbb{E}_{q \sim g(q)}\big[r_q \cdot \frac{\phi_i(q)}{\sum_{j \in S_q} \phi_j(q)} \big] \bigg).
\]

In real-world deployments, platforms often serve millions of queries, making exact evaluation of \( \mathbb{E}_{q \sim g(q)}\) computationally expensive. A practical solution is to use query sampling. Drawing \( k \) queries independently from \( g(q) \), we obtain a consistent sample-average estimator for the $\text{Total Payout}_i$, enabling efficient large-scale implementation.

%\hy{This para is confusing. }
%\rev{The Shapley axioms justify this allocation rule only \emph{conditional} on a chosen value function $v$: given $v$, the summary's value is fully allocated, interchangeable documents are treated symmetrically, documents that do not improve the summary receive zero, and query-level attributions aggregate consistently across queries. Whether the resulting payments are economically desirable is a separate question, depending on whether $v$ captures the platform's intended notion of user value and whether the retrieval and evaluation pipeline is sufficiently robust to gaming. Under these conditions, the marginal-contribution interpretation gives the allocation its incentive content: a document is paid for the incremental improvement it makes in the context where it is used, not for being long, frequently crawled, highly ranked, or one of many redundant sources.}

%Unlike flat licensing or arbitrary heuristics, such as attributing value based solely on user clicks, our framework accounts for each document’s marginal utility within the LLM summarization process. This avoids two key issues with click-based attribution: (1) many users do not click through to original documents, and (2) click rates are strongly influenced by the position or prominence of document presentation. In contrast, Shapley values capture each document’s actual contribution to the generated output. This enables fair compensation aligned with true informational value and supports both real-time (query-level) and batch (aggregate) attribution.

\subsection{Approximating Shapley Value: Cluster Shapley Approach}
\label{ssec:cluster_shapley}

\noindent \textbf{Computational Challenge:} While Shapley valuation is a theoretically appealing construct, evaluating Shapley values for source attribution presents a significant computational challenge even in moderate-sized settings. The computational cost associated with Shapley calculation exhibits exponential complexity -- if we have $|S_q|$ relevant documents for query $q$, the number of summarizations and evaluations needed to compute Shapley values scales as $2^{|S_q|}-1$, where $|S_q|$ is the number of relevant documents. This rapid growth in the number of calculations makes exact Shapley computation infeasible for large datasets, as the number of summarizations and evaluations quickly becomes overwhelming. Essentially, for each combination of documents, we need the LLM to generate a new summary and then perform an evaluation of that summary. These constraints suggest that exact Shapley methods in large-scale applications (e.g., document valuation for large platforms using LLMs) are infeasible, and we therefore need efficient approximation algorithms.% While parallelization and batch processing can reduce latency, the overall computational burden remains substantial. As discussed in a later section $\S$\ref{sssec:exac_shap_costs}, even for a query with eight relevant documents, exact Shapley computation involves processing 255 subsets, leading to significant API costs and delays in LLM settings. 

Researchers have proposed a number of algorithms designed to address the computational challenge associated with Shapley calculations. These approaches typically adopt a variety of sampling techniques to reduce the computational cost associated with Shapley calculation. One widely used method is Monte Carlo algorithm \citep{mann1960values}, which estimates Shapley values by randomly sampling permutations and computing marginal contributions across these samples. While this approach reduces computational costs compared to exact Shapley values, it still requires a large number of samples to achieve reasonable accuracy. Truncated Monte Carlo \citep{ghorbani2019data} improves efficiency by stopping the calculation early when additional samples provide diminishing returns below a threshold, significantly cutting down computational overhead. Another popular approach, Kernel SHAP \citep{lundberg2017unified}, employs a regression-based approximation to estimate Shapley values. However, none of these approaches leverage the textual content of documents when approximating Shapley values, treating them purely as independent units. 

\noindent \textbf{Key Idea:} Motivated by this limitation, we propose the Cluster Shapley algorithm. Instead of treating documents as independent, our method utilizes LLM-generated embeddings to identify and group similar documents, reducing redundant evaluations. Our core idea is intuitive: documents with similar content should have comparable contributions to the final summary and, therefore, should receive similar Shapley values. Our approach leverages the textual content of the documents and the LLM's numerical representation of this textual content (i.e., text embedding) to help approximate and simplify Shapley calculations. 

Text embedding techniques convert large chunks of text, such as sentences, paragraphs, or documents, into numerical vectors that capture semantic information. Earlier embedding methods, such as Word2Vec \citep{mikolov2013efficient} and GloVe \citep{pennington2014glove}, are based on shallow neural networks and co-occurrence statistics, learning word-level embeddings by predicting surrounding context words or factorizing word co-occurrence matrices. These embeddings typically represent each word with a fixed vector, independent of context. In contrast, modern LLM-based embeddings, such as those produced by OpenAI’s latest text-embedding models, are generated using Transformer-based architectures and are pretrained on massive text corpora. These newer embeddings are contextualized, meaning the vector for a word or sentence depends on its surrounding context, and are typically high-dimensional (e.g., 3072 dimensions in OpenAI’s \texttt{text-embedding-3-large} model). Unlike generative LLMs designed for tasks like chat or text generation, embedding models are optimized to produce semantically meaningful representations suitable for a wide range of downstream tasks. LLM-based embedding vectors have been successfully applied to a wide variety of discriminative tasks, including text classification, document retrieval, sentiment analysis, and predicting the attractiveness of news headlines \citep{patil2023survey, ye2025lola}. In our setting, we leverage these embeddings to cluster similar documents before computing Shapley values, allowing us to reduce redundant calculations.

\noindent \textbf{Our Approach:} We outline our proposed Cluster Shapley Algorithm in Algorithm \ref{alg:cluster shap}. Cluster Shapley begins with a preprocessing step, where for a given query $q$, we first determine the set of relevant documents $S_q \subseteq D$. This retrieval step ensures that only contextually relevant documents are considered; e.g., if the query pertains to political news, unrelated sports articles will be excluded from summarization. In $\S$\ref{ssec:summarization}, we discuss the retrieval and summarization steps, and related literature in further detail. For each document $i \in S_q$, we also obtain its embedding vector $e_i$, using an LLM embedding model. This step can be performed using proprietary models like OpenAI and Gemini or open-source alternatives such as Llama/Alpaca. 

Because similar documents tend to have similar embeddings, we can use the text embeddings to cluster the documents into similar groups. Specifically, after getting the embeddings, we cluster the embeddings of $S_q$ based on their distance, as outlined in Step 1. To achieve this, we first need to quantify the similarity between any two documents $i$ and $j$ in $S_q$. We employ cosine similarity, a widely used metric for measuring the closeness of embeddings, for this purpose. Cosine similarity is defined as:
\begin{equation}
\text{cosine\_similarity}(e_i, e_j) = \frac{e_i \cdot e_j}{\|e_i\| \|e_j\|}.
\end{equation}
This metric measures the cosine of the angle between two vectors in an inner product space, capturing how similar their directional components are. Higher cosine similarity values indicate greater textual similarity, meaning the embeddings of semantically similar documents are more aligned. To facilitate clustering, we define a corresponding distance measure, $d(e_i, e_j)$, which lies in \([0,1]\) for the review embeddings in our application (whose pairwise cosine similarities are non-negative; in general $1-\cos\in[0,2]$, or one may use the rescaled distance $(1-\cos)/2$), given by:
\begin{equation}\label{eqn:distance}
d(e_i, e_j) = 1 - \text{cosine\_similarity}(e_i, e_j) = 1 - \frac{e_i \cdot e_j}{\|e_i\| \|e_j\|}.
\end{equation}
This definition ensures that similar documents, which have high cosine similarity, are assigned a smaller distance value. Documents with lower distance values are more likely to be grouped together in the clustering process, allowing us to reduce redundancy and improve computational efficiency in Shapley value estimation.

Algorithm~\ref{alg:cluster shap} writes Cluster Shapley with a generic cluster-level Shapley procedure $\mathcal M$. Formally, for an $m$-player cooperative game $w:2^{[m]}\to\mathbb R$, $\mathcal M(w)\in\mathbb R^m$ returns a vector of cluster-level Shapley values or estimates. This notation covers both exact enumeration and approximate cluster-level computation; in our main empirical implementation, $\mathcal M$ is exact enumeration.

\begin{algorithm}[ht]
\small{
\caption{General Cluster Shapley Algorithm}
\label{alg:cluster shap}
\label{alg:cluster_shap_general}
\begin{algorithmic}
\State \textbf{Step 0: Inputs and preprocessing.} 
\quadstate Given a query $q$, retrieve the set of relevant documents $S_q$ and let $n=|S_q|$.
\quadstate For each document $i \in S_q$, obtain its embedding vector $e_i$.
\quadstate Set the clustering diameter $\epsilon>0$ and choose a cluster-level Shapley procedure $\mathcal M$.

\State \textbf{Step 1: Document clustering with the distance constraint.}
\quadstate Partition $S_q$ into clusters $\mathcal G=\{G_1,\ldots,G_m\}$ such that for all $k\in[m]$ and all $i,j\in G_k$, $d(e_i,e_j)\le \epsilon$.
\quadstate Let $k(i)$ denote the unique cluster index such that $i\in G_{k(i)}$.

\State \textbf{Step 2: Cluster-level Shapley computation.}
\quadstate Define the cluster-level game
\begin{equation}\nonumber
v_{\mathcal G}(T)=v\left(q,A\left(q,\bigcup_{G_k\in T}G_k\right)\right),\qquad T\subseteq \mathcal G.
\end{equation}
\quadstate Apply $\mathcal M$ to the cluster-level game $v_{\mathcal G}$ to obtain the cluster-level Shapley values $(\tilde\phi^{\mathcal G}_{G_1},\ldots,\tilde\phi^{\mathcal G}_{G_m})=\mathcal M(v_{\mathcal G})$.

\State \textbf{Step 3: Document-level allocation.}
\quadstate For each document $i\in G_{k(i)}$, assign
\begin{equation*}
    \tilde{\phi}_i = \frac{\tilde\phi^{\mathcal G}_{G_{k(i)}}}{|G_{k(i)}|}.
\end{equation*}
\end{algorithmic}}
\end{algorithm}

The goal of Step 1 is to partition the retrieved documents into $m$ non-overlapping clusters, i.e., $S_q=\bigcup_{k=1}^m G_k$ with $G_k\cap G_\ell=\emptyset$ for $k\neq \ell$. The key requirement is the pairwise diameter constraint: every pair of documents within a cluster must have embedding distance at most $\epsilon$. We can employ standard clustering algorithms such as K-Means \citep{lloyd1982least} or Density-Based Spatial Clustering of Applications with Noise (DBSCAN) \citep{ester1996density}, but many off-the-shelf procedures do not enforce this uniform intra-cluster proximity. To address this limitation, we use an adaptive DBSCAN algorithm (Algorithm~\ref{alg:dbscan}) detailed in Web Appendix~\S\ref{appsec:step1}. For comparison, we also empirically evaluate standard DBSCAN in \S\ref{ssec:robustness}, which performs worse in our setting.

The clustering diameter $\epsilon$ is a key hyperparameter that governs the accuracy--cost trade-off. Smaller values of $\epsilon$ produce finer clusters and therefore reduce within-cluster heterogeneity, but they increase the number of clusters $m$ and hence the cost of the cluster-level Shapley procedure. Larger values of $\epsilon$ yield coarser clusters and lower cost, but at the expense of greater approximation error. In practice, one can perform standard hyperparameter tuning: use a separate dataset to select the value of $\epsilon$ that best balances approximation accuracy and computational efficiency. Later in \S\ref{ssec:numerical_results}, we empirically show how $\epsilon$ governs this trade-off in our application setting.

In Step 2, we consolidate the documents within each cluster by concatenating them into a single \emph{meta-document}, which serves as the representative unit for that cluster in the cluster-level game. The Shapley procedure $\mathcal M$ can be exact Shapley, in which case the algorithm computes the cluster values by evaluating all $2^m-1$ non-empty cluster subsets, or it can be an approximation method such as Monte Carlo when $m$ remains large. Thus, the exact cluster-level implementation used in our main empirical analysis is a special case of Algorithm~\ref{alg:cluster shap} with $\mathcal M$ equal to exact Shapley. Finally, in Step 3, we allocate each cluster's value equally across its member documents, assigning $\tilde\phi_i=\tilde\phi^{\mathcal G}_{G_{k(i)}}/|G_{k(i)}|$.

\subsection{Theoretical Analysis of Cluster Shapley Algorithm}
\label{ssec:theory}

Our analysis of Algorithm~\ref{alg:cluster shap} delivers three results. First, Theorem~\ref{thm:cluster_shap_general} provides a general, oracle-style decomposition of the approximation error that applies to any query-specific partition and any cluster-level Shapley procedure. Second, we analyze the algorithm's computational complexity and instantiate the general bound for a Monte Carlo cluster-level procedure (Corollary~\ref{cor:error_mc}), yielding explicit error and cost expressions. Third, we translate the approximation-error bound into a guarantee on the revenue attribution of \S\ref{ssec:revenue_attribution} (Corollary~\ref{cor:revenue}), directly linking the algorithm's accuracy to its economic application.

\noindent \textbf{Error Decomposition.} The bound below is an oracle-style decomposition that applies to any query-specific partition $\mathcal G$ and any cluster-level Shapley procedure $\mathcal M$; it requires no smoothness assumption on the value function. For document $i$, let $k(i,q)$ index the cluster containing it, and let $\phi^{\mathcal G}_{G_k}(q)$ denote the exact Shapley value of cluster $G_k$ in the cluster-level game $v_{\mathcal G}$. The \emph{fixed-partition coarsening bias} for document $i$ is
\[
    b_i(\mathcal G,q)
    =
    \left|\frac{\phi^{\mathcal G}_{G_{k(i,q)}}(q)}{|G_{k(i,q)}|}-\phi_i(q)\right|,
\]
where $\phi_i(q)$ is the exact document-level Shapley value in the original game over $S_q$; this is the error that would remain even if Step 2 were solved exactly. The \emph{realized cluster-level numerical error} of $\mathcal M$ is
\[
    \delta_{\mathcal M}(q)
    =
    \max_{k\in[m]}\left|\tilde\phi^{\mathcal G}_{G_k}(q)-\phi^{\mathcal G}_{G_k}(q)\right|,
\]
with $\delta_{\mathcal M}(q)=0$ when $\mathcal M$ computes exact cluster-level Shapley values. We also write $B_1(\mathcal G,q)=\frac{1}{n}\sum_{i\in S_q}b_i(\mathcal G,q)$, with $n=|S_q|$, for the mean coarsening bias that enters the aggregate (mean-absolute-error) bounds below.

\begin{theorem}[General Cluster Shapley Oracle Bound]
\label{thm:cluster_shap_general}
For every document $i\in S_q$,
\begin{equation}
\label{eq:general_error_bound}
\left|\tilde\phi_i(q)-\phi_i(q)\right|
\le
b_i(\mathcal G,q)
+
\frac{\delta_{\mathcal M}(q)}{|G_{k(i,q)}|}.
\end{equation}
Consequently, if $\mathcal M$ admits a high-probability error bound $\Delta_{\mathcal M}(\eta)$, that is, $\Pr\{\delta_{\mathcal M}(q)\le\Delta_{\mathcal M}(\eta)\}\ge 1-\eta$, then with probability at least $1-\eta$,
\[
\left|\tilde\phi_i(q)-\phi_i(q)\right|
\le
b_i(\mathcal G,q)
+
\frac{\Delta_{\mathcal M}(\eta)}{|G_{k(i,q)}|}
\qquad\text{for all } i\in S_q.
\]
The computational complexity of Algorithm~\ref{alg:cluster shap} is $O(n^2+C_{\mathcal M}(m))$, where $C_{\mathcal M}(m)$ is the cost of the cluster-level procedure $\mathcal M$.
\end{theorem}

See Web Appendix~\S\ref{appssec:proof_3} for the proof. Theorem~\ref{thm:cluster_shap_general} separates the two forces that govern approximation quality. The first term, $b_i(\mathcal G,q)$, is the deterministic error induced by clustering, meta-document construction, and equal within-cluster allocation; it is zero for the singleton partition and can be evaluated empirically on validation queries for which exact Shapley values are computed. The second term is the cluster-level numerical approximation error, divided by the size of the cluster that receives the estimated cluster value. If the cluster-level procedure is exact Shapley, then $\delta_{\mathcal M}(q)=0$, and the error is entirely the fixed-partition coarsening bias. Aggregate $\ell_\infty$ and mean-$\ell_1$ (MAE) versions of the bound are recorded in the proof in Web Appendix~\S\ref{appssec:proof_3}. If the partition is refined to singleton documents and Step 2 is exact, then $b_i(\mathcal G,q)=0$ for all $i$ and Cluster Shapley recovers exact document-level Shapley values.

The clustering diameter $\epsilon$ affects Theorem~\ref{thm:cluster_shap_general} through the deterministic bias $b_i(\mathcal G,q)$: smaller $\epsilon$ produces finer partitions that typically reduce coarsening bias but increase $m$ and therefore the cost of the cluster-level procedure. Web Appendix~\S\ref{appssec:lipschitz_specialization} gives one interpretable specialization under the Lipschitz condition validated in Web Appendix~\S\ref{appssec:assumption}: in that case, $b_i(\mathcal G,q)$ is bounded by a within-cluster dispersion term and a meta-document aggregation term. We keep this specialization in the appendix because the oracle bound above is more general and does not rely on smoothness in the embedding space.

\noindent \textbf{Computational Complexity.} We now analyze the computational complexity of Algorithm~\ref{alg:cluster shap}. The clustering step requires computing embeddings and pairwise distances for $n=|S_q|$ documents. Generating embeddings scales linearly in $n$, while computing the pairwise distance matrix costs $O(n^2)$ in the worst case. The adaptive DBSCAN procedure used in Step 1 is also polynomial and is dominated by the distance-matrix computation in our setting. Thus, the clustering phase is $O(n^2)$, which is modest relative to exhaustive Shapley computation.

The remaining cost is the cluster-level procedure $\mathcal M$. If $\mathcal M$ computes exact Shapley values over the $m$ clusters, it evaluates $2^m-1$ non-empty cluster subsets and has complexity $O(2^m)$. The resulting exact-cluster implementation therefore has complexity $O(n^2+2^m)$, reducing the exponential problem size from $n$ documents to $m$ clusters. Cluster Shapley does not eliminate the combinatorial nature of exact Shapley computation; rather, it exploits semantic redundancy to reduce the number of effective players.

\noindent \textbf{Monte Carlo Instantiation.} When $m$ remains large, Algorithm~\ref{alg:cluster shap} can use an approximation algorithm for $\mathcal M$. Theorem~\ref{thm:cluster_shap_general} applies directly to any cluster-level procedure with a known error bound and computational cost. The following corollary instantiates the result for Monte Carlo Shapley.

\begin{corollary}[Monte Carlo Cluster Shapley Error and Complexity]
\label{cor:error_mc}
Suppose $\mathcal M$ estimates cluster-level Shapley values by averaging marginal contributions over $N$ random permutations, and suppose each cluster-level marginal contribution has range at most $V_{\max}$. Then, for every document $i\in S_q$, with probability at least $1-\eta$,
\begin{equation}
\label{eq:error_mc_shapley}
\left|\tilde\phi_i(q)-\phi_i(q)\right|
\le
b_i(\mathcal G,q)
+
\frac{V_{\max}}{|G_{k(i,q)}|}\sqrt{\frac{\log(2m/\eta)}{2N}}.
\end{equation}
Aggregating across documents, the mean absolute error satisfies
\begin{equation*}
\frac{1}{n}\|\tilde{\boldsymbol\phi}(q)-\boldsymbol\phi(q)\|_1
\le
B_1(\mathcal G,q)
+
\frac{m}{n}V_{\max}\sqrt{\frac{\log(2m/\eta)}{2N}}.
\end{equation*}
For a target document-level error $\epsilon_{\mathrm{total}}>b_i(\mathcal G,q)$, it suffices to choose
\[
N\ge
\frac{V_{\max}^2\log(2m/\eta)}{2|G_{k(i,q)}|^2\left(\epsilon_{\mathrm{total}}-b_i(\mathcal G,q)\right)^2}.
\]
The total computational complexity is $O(n^2+Nm)$.
\end{corollary}
\noindent See Web Appendix~\S\ref{appssec:proof_4} for the proof. This corollary shows that the cluster-level Monte Carlo component scales at the standard $N^{-1/2}$ rate, while clustering affects accuracy through the deterministic bias $B_1(\mathcal G,q)$ and computation through the number of clusters $m$.

\noindent \textbf{From Approximation Error to Revenue Attribution.} We close by translating the approximation error of Theorem~\ref{thm:cluster_shap_general} into a guarantee on the \emph{revenue attribution} of \S\ref{ssec:revenue_attribution}, directly linking the algorithm's accuracy to its economic application.

\begin{corollary}[Revenue Attribution Error]
\label{cor:revenue}
Let $V_q=v(q,A(q,S_q))$. Consider the raw query-level Shapley revenue allocation
\[
r_i(q)=\beta r_q\frac{\phi_i(q)}{\sum_{j\in S_q}\phi_j(q)}.
\]
Let $\tilde r_i(q)$ denote the analogous allocation computed from Cluster Shapley values $\tilde\phi_i(q)$. If the cluster-level procedure $\mathcal M$ is efficiency-preserving, so that $\sum_{k=1}^m\tilde\phi^{\mathcal G}_{G_k}(q)=V_q$, then, with probability at least $1-\eta$,
\begin{equation*}
\left|\tilde r_i(q)-r_i(q)\right|
\le
\frac{\beta r_q}{V_q}
\left(
    b_i(\mathcal G,q)
    +
    \frac{\Delta_{\mathcal M}(\eta)}{|G_{k(i,q)}|}
\right).
\end{equation*}
The analogous subscription-revenue bound is
\begin{equation*}
\left|\tilde r_i-r_i\right|
\le
\frac{\beta R}{\mathbb E_q[V_q]}
\mathbb E_q\left[
    b_i(\mathcal G,q)
    +
    \frac{\Delta_{\mathcal M}(\eta)}{|G_{k(i,q)}|}
\right].
\end{equation*}
\end{corollary}

See Web Appendix~\S\ref{appssec:proof_revenue} for the proof. The key observation is that efficiency-preserving Cluster Shapley values and exact Shapley values have the same query-level denominator, $V_q$. Therefore, the revenue-attribution error is the Shapley approximation error scaled by the revenue paid per unit of summary value.

In sum, our theoretical analysis shows that Cluster Shapley is a flexible algorithm with a transparent accuracy--cost trade-off. The empirical results below directly measure the realized error against exact document-level Shapley values in our application.

\section{Application Setting: Amazon Review Dataset}
\label{sec:amazon_data}
We now present an application of our algorithm to a real setting. We use the publicly available Amazon Product Reviews dataset as the empirical context to demonstrate the performance of our document valuation approach. This dataset was collected by \citet{hou2024bridging} and has been used extensively in recent research studies on a variety of topics, including sentiment analysis \citep{haque2018sentiment}, sequential product search and recommendation \citep{hou2024bridging}, fine-tuning of LLMs \citep{zhang2024llasa}, and evaluation of LLM alignment \citep{shankar2024validates}. The dataset spans from May 1996 to September 2023, featuring over 571.54 million reviews from 54.51 million users and covering 48.19 million unique items. It is organized into 33 distinct categories, including electronics, household goods, clothing, and books. This user review data set consists of textual feedback provided by users that captures their opinions, ratings, and experiences with products. %This component contains 30.14 billion review tokens. 
A comprehensive analysis of review categories, basic statistics, and detailed data field information is available in \citet{hou2024bridging}. 

While it is well-established that review valence and content can help consumers make better decisions \citep{chevalier_mayzlin_2006}, it is also well-understood that it is hard for consumers to process the large amounts of information/text in reviews. For example, the most popular products in the data have hundreds or even thousands of reviews. Thus, consumers must often sift through hundreds of reviews to extract relevant insights. This information overload makes it difficult for users to efficiently locate specific details (e.g., product quality, value for money, durability, ease of return).

To help consumers navigate this vast amount of information, online platforms typically rank reviews by helpfulness votes and allow searching for specific information. While these solutions can aid consumers in their quest for information, they nevertheless require users to sift through a large volume of irrelevant information and expend significant time and effort on the task. As such, it often leads to inefficient searches and potentially uninformed purchasing decisions. To that end, many e-commerce platforms (including Amazon) have started adopting LLMs to retrieve and summarize the most relevant information for a consumer's specific query from the available set of reviews/user-generated content (see Figure~\ref{fig:amazon_review}). Customers can either see a summary from all reviews or query the system for a specific piece of information (e.g., ease of return) through Amazon's Alexa for Shopping assistant (formerly the Rufus chatbot). For our analysis, we focus on query-based summaries, though our framework is quite general and can also be applied to the general summarization settings.

For our numerical experiments, we select 24 products from different categories to ensure a diverse representation of consumer goods (Table \ref{tab:designed_queries_full} in Web Appendix~\S\ref{appssec:query_list}). These products span a variety of domains, including video games, beauty products, and personal care items, with review counts varying widely. The number of reviews per product ranges from 323 to 15,594, with a mean of 2,075 and a standard deviation of 3,216. Even the product with the fewest reviews presents a significant information overload for consumers, making it impractical to read through all reviews manually. While our methodology can be applied to a larger set of products, our empirical findings do not fundamentally change with more products. Therefore, we focus on this smaller subset of products for expositional and computational ease. %The quartile distribution is as follows: the 25th percentile is at 565.5 reviews, the median at 1,232 reviews, and the 75th percentile at 1,570.75 reviews. 

To compute the Shapley value of each individual review (within the context of a given product), we need to first specify the distribution of queries, $g(q)$, that consumers use when requesting summaries for this product. While this query distribution is not publicly available, for each product, we observe the set of popular attributes frequently mentioned by customers. For instance, in Figure~\ref{fig:amazon_review}, for the wireless controller product, attributes such as ``Functionality'' and ``Controller quality'' are among the most common aspects of the product that users are concerned about. This information allows us to craft a proxy distribution of user queries for each product that mimics the real distribution of queries. Specifically, for each product, we design two queries based on the top two attributes most frequently mentioned in its reviews, as shown in the last column of Table~\ref{tab:designed_queries_full}. Each query is assigned an equal probability, contributing equally to the overall valuation.\footnote{In practice, the platform will have data on the true distribution of queries and can directly use that empirical distribution in its analyses. As discussed earlier, our approach is agnostic to the exact distribution of queries.}

\section{Implementation Details}
\label{sec:implementation_details_shapley_value}

%Recall that the two key inputs to our document valuation framework are a summarization method $A(\cdot)$ and an evaluation method $v(\cdot)$. In $\S$\ref{ssec:rag}, we first introduce the RAG architecture, which describes how pre-trained LLMs can be augmented for search tasks with domain-relevant documents/articles. Next, in $\S$\ref{ssec:summarization} and $\S$\ref{ssec:eval_step}, we detail our summarization $A(\cdot)$ and evaluation $v(\cdot)$, respectively. In $\S$\ref{ssec:stochasticity}, we discuss the challenges associated with stochasticity in summarization and evaluation in real settings and how we handle them. Finally, we discuss the implementation details for the exact Shapley and Cluster Shapley approaches in $\S$\ref{ssec:shap_cal_exac} and $\S$\ref{ssec:clu_shap_cal}.

The proposed document valuation framework is built on two key components: a summarization method $A(\cdot)$ and an evaluation method $v(\cdot)$, which are detailed in $\S$\ref{ssec:summarization} and $\S$\ref{ssec:eval_step} respectively. Then, we discuss the implementation details for the exact Shapley and Cluster Shapley approaches in $\S$\ref{ssec:shap_cal_exac} and $\S$\ref{ssec:clu_shap_cal}.

\subsection{Summarization of Relevant Amazon Reviews via RAG}
\label{ssec:summarization}

AI search engines are designed to provide real-time, contextually relevant responses to user queries. A key enabling technique behind many of these systems is retrieval-augmented generation (RAG), which integrates pre-trained LLMs with information retrieval modules to augment model outputs with domain-relevant documents \citep{lewis2020retrieval}. A detailed introduction to the RAG architecture is provided in Web Appendix~\S\ref{ssec:rag}.

Next, we construct the RAG-based search and summarization tool that finds the relevant documents $S_q$ from Amazon reviews and produces the summary $A(q, S_q)$ for any given query $q$; Figure~\ref{fig:architecture_of_AI_engine_for_amz} in Web Appendix~\S\ref{appssec:rag_pipeline} shows the architecture, and the appendix details each step. In brief, we first pre-compute embeddings for all reviews of a product using OpenAI's \texttt{text-embedding-3-large} model (excluding reviews shorter than ten words, which tend to be uninformative). Given a user query $q$, we extract its key semantic content using an LLM (\texttt{GPT-4o-2024-08-06}), embed it with the same embedding model, and retrieve the eight most relevant reviews ($|S_q|=8$) by the cosine similarity between the query and review embeddings, a retrieval depth in line with current industry practice \citep{perplexity_citation_count, google_citation_count}. Finally, we combine the retrieved reviews with the query in a structured prompt (Web Appendix~\S\ref{appssec:prompt_design}), and GPT-4o generates a query-specific summary that cites its supporting reviews in square brackets (e.g., ``[2]'') and explicitly marks irrelevant reviews as such.

\subsection{Evaluation of Summarized Amazon Reviews}
\label{ssec:eval_step}

Next, we describe the implementation of $v(\cdot)$, a function that evaluates the quality of generated summaries. To operationalize this, we design a prompt for GPT-4o that serves as the performance evaluation method. The full prompt is shown in Figure~\ref{fig:complete_evaluation_task_prompt} in Web Appendix $\S$\ref{appssec:prompt_design}. This prompt takes summaries as inputs and outputs a performance score. The LLM evaluates each summary’s informativeness based on its “Information Coverage,” reflecting how well the summary captures key aspects of the product reviews.\footnote{Nevertheless, as discussed in $\S$\ref{sec:problem_def}, our framework is agnostic to the exact evaluation tool used, and other approaches can be used.} Each summary is rated on a scale from 0 to 10,\footnote{We chose a 0 to 10 scale to offer sufficient granularity for distinguishing levels of information coverage, as smaller scales (e.g., 0 to 5) lack subtlety, while larger scales (e.g., 0 to 100) add unnecessary complexity. We tested alternative ranges to confirm this choice for optimal consistency in evaluation.} with higher scores indicating a more comprehensive and accurate reflection of relevant information. The LLM is instructed to prioritize clarity and relevance, emphasizing key details. 

Note that LLM outputs are inherently stochastic due to the probabilistic nature of token generation. As a result, both the summarization process $A(q, S)$ and the evaluation process $v(q, A(q, S))$ can produce non-deterministic outputs, even under identical inputs. To quantify and manage this stochasticity, we analyze its impact on the variance of Shapley values. As shown in Web Appendix $\S$\ref{appssec:var_analysis}, we find that the summarization step contributes approximately 53\% of the total output variance, while the evaluation step contributes about 47\%. To reduce the effect of randomness in evaluation, we run four independent evaluations per summary and use their average score. This simple averaging method substantially reduces evaluation variance while maintaining computational efficiency. Please see Web Appendix $\S$\ref{appsssec:reduce_var_shap} for additional analysis and results.

\subsection{Exact Shapley Value Calculation}
\label{ssec:shap_cal_exac}
We now discuss the implementation details for the exact Shapley value calculation and its scalability and costs. 

%\subsubsection{Exact Shapley Implementation and Example}
%\label{sssec:exac_shap_example}

For each product, we calculate the exact Shapley values using the formula in Equation \eqref{eqn:shapley}. Table~\ref{tab:filtered_reviews_shapley_value} presents the Shapley values for the top eight most relevant reviews in response to the query, ``How is the quality of the wireless controller?'' for the first product. Other reviews not contributing to this query receive a Shapley value of zero. In this example, we see that Review \#3 has the highest Shapley value ($1.83$), as it directly compares the controller's quality to other versions and emphasizes functionality, aligning well with the prompt’s emphasis on “Information Coverage” for quality details. Similarly, Review \#7 ($1.61$) and Review \#2 ($1.58$) score highly for addressing quality explicitly: \#7 in a positive tone and \#2 by highlighting durability compared to off-brand controllers. Review \#5 ($1.44$) also performs well by underscoring the superior quality of the original controller versus knockoff brands. Review \#4 ($1.25$) is somewhere in the middle, highlighting the good quality but without additional information relevant to the query. The lower-scoring reviews, including Review \#1 ($0.59$) and Review \#6 ($0.53$), just generally mention the great quality and cheap price, lacking specific details. Review \#8 has the lowest value of $0.17$, as it emphasizes aspects like shipping and being good for gifts, which are less relevant to the query's focus on controller quality, though the title mentions the ``Quality'', which makes it relevant to the query.

\begin{table}[htbp]
\centering
\small
\begin{tabular}{@{}c>{\raggedright}p{4cm}p{9cm}c@{}}
\toprule
\textbf{No.} & \textbf{Title} & \textbf{Main Text} & \textbf{Shapley} \\ \midrule
1 & Cheaper price, same great quality& This product stands as a testament to the reason I go to the store to find the product then buy it online at a cheaper price. & 0.59\\ 
2 & Quality & It's worth the price. Controllers last much longer than off brand. & 1.58\\ 
3 & Great Quality and Price & Great price and product and unlike others this one worked. Ordered one from ebay and it was garabe but this seller is legit 5 stars. & 1.83\\ 
4 & Great buy and Product is exactly what I expected! & I liked the red color and that the product quality was exactly what I needed!& 1.25\\ 
5 & Five Stars & I only recommend the original makers product, pay more but better then the knockoffs. & 1.44\\ 
6 & great product & Great product and so much cheaper than buying it in store. & 0.53\\ 
7 & Nice, new and crispy & Nice new and crispy! Very happy with the quality, the vendor and the price 10/10 would recommend.& 1.61 \\ 
8 & Quality & Very nice and the shipping was very quick. My grandson loved it for Christmas.& 0.17\\ \bottomrule
\end{tabular}
\caption{Shapley values of Top 8 relevant Amazon reviews for the query ``How is the quality of the wireless controller?''.}
\label{tab:filtered_reviews_shapley_value}
\end{table}

%\subsubsection{Computational Costs and Scalability}
%\label{sssec:exac_shap_costs}
Next, we discuss the cost of implementing the exact Shapley calculation. For each query with 8 relevant reviews, we must process $2^8 - 1 = 255$ distinct subsets, with each subset requiring a summarization and four evaluations to ensure reliable scoring. We use OpenAI's API to process both summarization and evaluation with GPT-4o, and it costs about \$1.30 in OpenAI API fees per query.

%Our experiments indicate that processing a single query takes 15 minutes (on average)\footnote{This time includes the full process for both summarization and evaluation of Python-based API calls, network latency, time to first token, and all computational overheads. Processing time depends significantly on the \href{https://platform.openai.com/docs/guides/rate-limits}{OpenAI API tier level}; our experiments use Tier 2 access.} and costs about \$1.30 in OpenAI API fees per query. As we can see, this can become prohibitive in both time and money as the number and variety of queries scale up.\footnote{Batch processing, i.e., simultaneous API calls to OpenAI, can effectively reduce the processing time from 15 minutes to around 3.5 seconds by parallelizing the 255 summarizations and evaluations. However, the total computation time and cost remain unchanged. Alternatively, open-source LLMs for summarization and evaluation can further reduce both time and costs. For simplicity, we report the total computation time based on GPT-4o throughout the paper.} 

To illustrate the scalability challenge, consider Perplexity AI, a leading LLM-based search engine, which serves over 400 million queries per month \citep{srinivas2025}. Applying the exact Shapley algorithm to each query for this volume would imply over \$520 million in monthly compute costs (\$1.3 per query × 400 million queries). Switching to a lower-priced open-source LLM model such as \texttt{Llama-3-8B-Instruct-Lite} can reduce API costs by approximately 98.4\%.\footnote{GPT-4o costs \$1.313 per million tokens compared to just \$0.021 per million tokens for Llama 3 \citep{aimlapi2025}.} However, even with Llama 3, the cost to calculate exact Shapley values would still exceed \$8.3 million (\$520 million per month $\times$ (1 - 98.4\%)) per month, which is still too high compared to Perplexity's annualized revenue of \$100 million as of March 2025~\citep{srinivas2025}. The scale of these computational expenses creates a significant gap between theoretical document valuation frameworks and practical implementation, highlighting the need for an efficient approximation algorithm like our proposed Cluster Shapley algorithm.

\subsection{Cluster Shapley Implementation}
\label{ssec:clu_shap_cal}

For the implementation of our proposed Cluster Shapley algorithm, as described in Algorithm~\ref{alg:cluster shap}, we use exact Shapley computation for the cluster-level procedure $\mathcal M$ and first specify the clustering diameter hyperparameter \(\epsilon\). In our numerical comparison in $\S$\ref{ssec:numerical_results}, we evaluate a spectrum of \(\epsilon\) values to illustrate the trade-off between approximation error and computational time. Specifically, we explore \(\epsilon\) values ranging from 0 to 1 in increments of 0.025. As discussed in Web Appendix~$\S$\ref{appsec:step1}, one can apply standard hyperparameter tuning procedures to select an appropriate \(\epsilon\). To assess the robustness of this choice, we conduct an additional experiment in $\S$\ref{ssec:robustness} (detailed in Web Appendix $\S$\ref{appssec:split_sample}), where we randomly split the test dataset into two subsets, one for tuning \(\epsilon\) and the other for implementation and evaluation. The results demonstrate that the Cluster Shapley algorithm is fairly robust to the choice of \(\epsilon\).

After specifying the clustering diameter \(\epsilon\), we apply Algorithm~\ref{alg:dbscan} to perform document clustering. Note that the distance matrix used in the algorithm is computed based on Equation~\eqref{eqn:distance}. To obtain document embeddings, we extract query-relevant information from each comment individually, using the same LLM prompt as in the summarization step but applied to one document at a time. This extraction adds $n$ lightweight LLM calls (one per document) as preprocessing. The computational cost of our adaptive clustering algorithm is negligible in our setting, especially when compared to the cost of LLM-based summarization and evaluation, whose $2^{m}-1$ cluster-level summary and evaluation calls dominate the $O(n)$ preprocessing. We analyze and report the computation time of the clustering step using Algorithm~\ref{alg:dbscan} in Web Appendix~\S\ref{appssec:cluster_epsilon_iteration_time}.

\begin{figure}[htbp]
    \centering
    \includegraphics[width=0.6\textwidth]{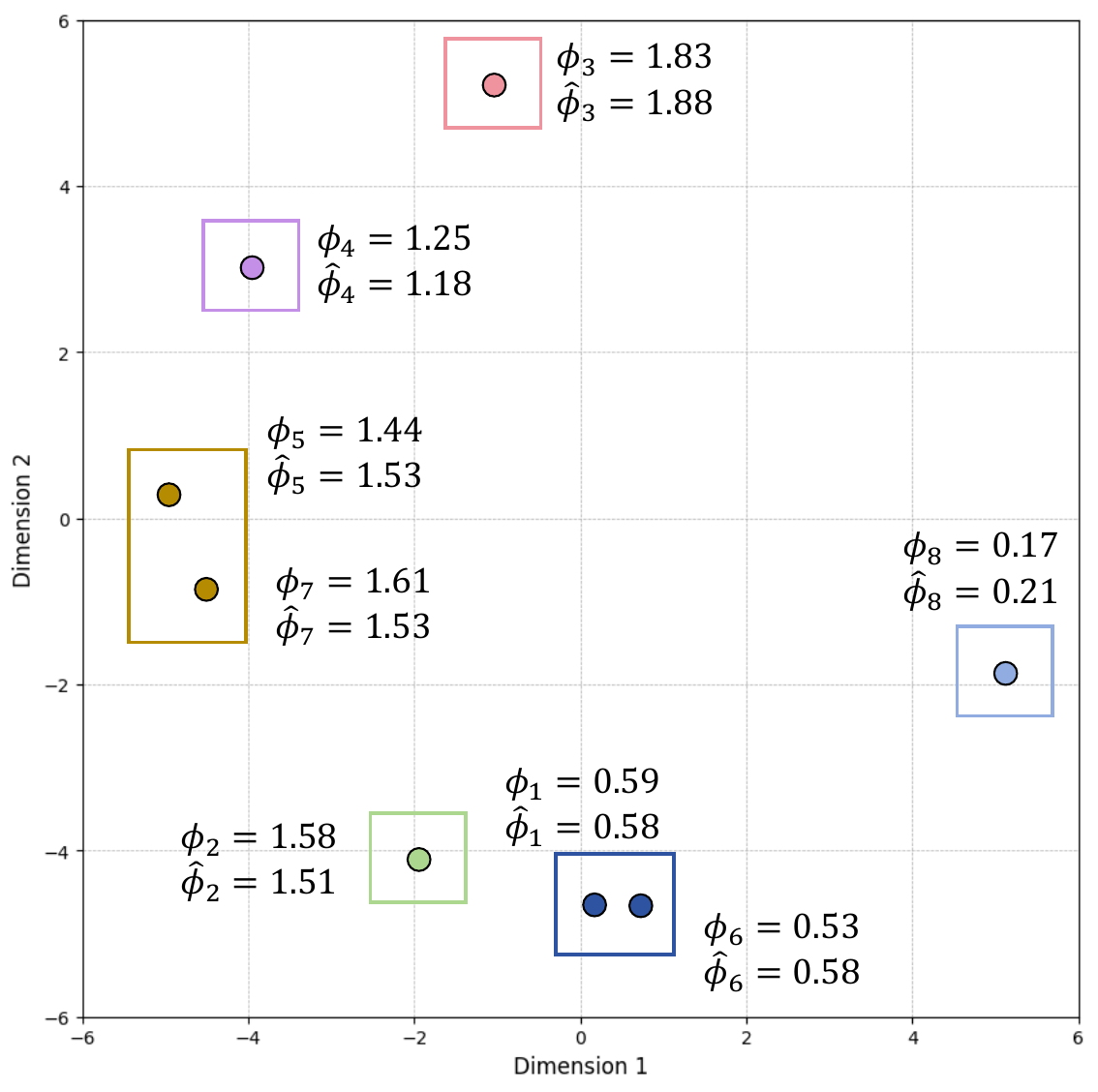}
    \caption{Clustering result of Top 8 relevant Amazon reviews for the query ``How is the quality of the wireless controller?'' We use 3072-dimensional OpenAI embeddings for the clustering. However, we use PCA to reduce the embedding dimension to 2 for better visualization. Dots represent the reviews, and squares represent clusters. ${\phi}_i$ is the exact Shapley value while $\hat{\phi}_i$ is the approximated Shapley value by the Cluster Shapley algorithm.}
    \label{fig:clustering_vis}
\end{figure}

Figure~\ref{fig:clustering_vis} presents the clustering results using Algorithm \ref{alg:dbscan} for a sample query. In this instance, $\epsilon = 0.05$, which yields six clusters. Increasing $\epsilon$ results in fewer clusters, further reducing computational cost, but may introduce higher approximation error. Even with six clusters, the computational complexity is significantly reduced-shrinking from $2^8-1=255$ to $2^6-1=63$, representing a fourfold improvement in efficiency.

In Step 2, we append all the documents within a cluster, treat each cluster as a meta-document, and obtain cluster-level exact Shapley values. At this stage, the summarization and evaluation steps follow exactly the same prompts as those used in the exact Shapley calculation, except that we use the meta-documents at the cluster level as the input for summarization. Finally, we distribute the cluster-level Shapley values equally back to the individual documents within the clusters. As shown in the example in Figure~\ref{fig:clustering_vis}, documents within the same cluster have similar exact Shapley values, and our Cluster Shapley approximation achieves high accuracy. For instance, reviews $1$ and $6$ are in the same cluster with similar exact Shapley values of $\phi_1 = 0.59$ and $\phi_6 = 0.53$, and the same approximated Shapley value $\hat{\phi}_1 =\hat{\phi}_6= 0.58$. As shown in Table~\ref{tab:filtered_reviews_shapley_value}, both reviews $1$ and $6$ emphasize that the price is cheap but only mention that the product is great without details. The mean absolute error (MAE) between the exact and approximated Shapley values across all reviews in this example is only 0.04, demonstrating that our algorithm successfully approximates the Shapley value with a low error and reduces the computation cost. It also validates our theory that semantically similar documents tend to have comparable marginal contributions to the summary, and thus similar Shapley values.

\section{Numerical Results}
\label{sec:results}
We now present the results from the application of our Cluster Shapley algorithm to the Amazon review setting. First, in $\S$\ref{ssec:numerical_results}, we report the numerical results of our proposed Cluster Shapley approach and compare its performance with three other off-the-shelf Shapley approximation algorithms. Next in $\S$\ref{ssec:simple rule}, we further contrast Cluster Shapley with the simple attribution rules (with minimal computational cost) and discuss the corresponding findings. In $\S$\ref{ssec:robustness}, we present a series of robustness checks and extensions. 

%\subsection{Benchmark Algorithms}
%\label{ssec:benchmarks}

\subsection{Comparison with Off-the-shelf Shapley Approximation Algorithms}
\label{ssec:numerical_results}

We now introduce three widely used Shapley value approximation algorithms that serve as off-the-shelf benchmarks and compare the performance of our proposed algorithm against them.

\squishlist

\item \textbf{Monte Carlo:} The Monte Carlo algorithm (or permutation sampling) is a popular approach for approximating Shapley values \citep{mann1960values}. This method randomly samples permutations from the $|S_q|!$ possible combinations of documents and then for each document $i$ and one permutation $P^\pi_i$, calculates its marginal contribution, i.e., $v(q,A(q,P^\pi_i \cup \{i\})) - v(q, A(q,P^\pi_i))$. Shapley value can then be approximated using the sample average of marginal contributions over all sampled permutations. As the number of permutation samples increases, the approximation error decreases, but the computational cost grows linearly. In our numerical experiments, we progressively increase the number of permutations to show the trade-off between accuracy and efficiency.

\item \textbf{Truncated Monte Carlo:} The algorithm accelerates Shapley value calculation by adaptively reducing the number of evaluated samples. This method operates under the idea that marginal contributions tend to diminish as the coalition grows, i.e., $v(q,A(q, S_1 \cup \{i\})) - v(q,A(q, S_1)) \leq v(q,A(q, S_2 \cup \{i\})) - v(q,A(q, S_2))$ if $S_2\subseteq S_1$, i.e., the marginal contribution of document $i$  decreases when more documents come into the permutation. This is because, with a larger set of permutations, document $i$ is more likely to have higher overlapping information with other documents, reducing its marginal contribution.

We briefly summarize the algorithm here and refer readers to \citet{ghorbani2019data} for details. This algorithm randomly samples a permutation of reviews and sequentially calculates performance scores, $v$, by adding reviews in the permutation order. Under this diminishing-returns heuristic, the algorithm truncates the computation by assigning zero marginal contributions to the remaining reviews when the gap between the current score and the maximum score (10 in our setting) is smaller than a pre-specified threshold, called {\it performance tolerance}. In effect, the algorithm treats the remaining reviews' marginal contributions as negligible relative to this threshold. Thus, this algorithm simply assigns zero marginal contribution instead of calculating the negligible marginal value. 

The performance tolerance parameter, which governs the allowable change in Shapley values before truncation occurs, is tuned over a range of values -- \{0.1, 0.2, 0.3, 0.5, 0.7, 1, 2, 3\} -- to balance estimation accuracy and computational efficiency. Smaller values reduce the effectiveness of truncation-causing Truncated Monte Carlo to behave similarly to standard Monte Carlo-while larger values result in early truncation, degrading estimation accuracy. Based on the tuning, we select 0.5 for our main experiments. After fixing the performance tolerance at 0.5, we vary the number of sampled permutations to construct the efficient frontier, which illustrates the trade-off between computational cost and approximation error.

\item \textbf{Kernel SHAP:} Kernel SHAP is a model-agnostic approach to approximating Shapley values based on weighted least squares regression \citep{lundberg2017unified}. Our implementation uses Python's SHAP package, which we adapt specifically for our LLM-based summarization task by implementing a custom mapping function between subset compositions and their corresponding summarization scores. The method employs KernelExplainer with an identity link function and L1 regularization to enhance numerical stability. We tested increasing numbers of samples to evaluate the performance of the method under different computational budgets. Kernel SHAP has been widely used across domains, including NLP for transformer-based model interpretation \citep{kokalj2021bert}, finance for credit risk analysis \citep{fama2024explainable}, healthcare for clinical decision support \citep{li2022unified}, and marketing for optimizing content engagement \citep{kong2023neural}. Recent work has pointed out some limitations of Kernel SHAP, especially when interactions are strong \citep{gosiewska2019not, ragodos2024model}. Nevertheless, given that it is one of the most widely used Shapley approximations used in practice, we include it in our evaluation for benchmarking purposes.

\squishend

%\footnote{Note that the issues pointed out by these papers relate to the approximation process used in the Kernel SHAP algorithm, and not the Shapley concept itself.}
We now present the results from the empirical exercise comparing the performance of our algorithm with the above three algorithms. Our experiments include the 48 test queries described in $\S$\ref{sec:amazon_data}. Each query comprises the eight most relevant reviews selected from the Amazon review dataset, forming the foundation for our comparative analysis of various Shapley value approximation algorithms.

To establish a stable evaluation baseline and reduce variance introduced by the summarization and evaluation steps, we standardize the process as follows: for each query, we generate a single summary for each subset (for all 255 possible subsets) and fix the evaluation score for each summary by averaging four evaluations, as detailed in $\S$\ref{ssec:eval_step}. By fixing sample paths, we mitigate the inherent randomness in LLM outputs, ensuring consistent baseline measurements across different approximation methods.

\begin{figure}[htbp]
    \centering
    \includegraphics[width=0.8\textwidth]{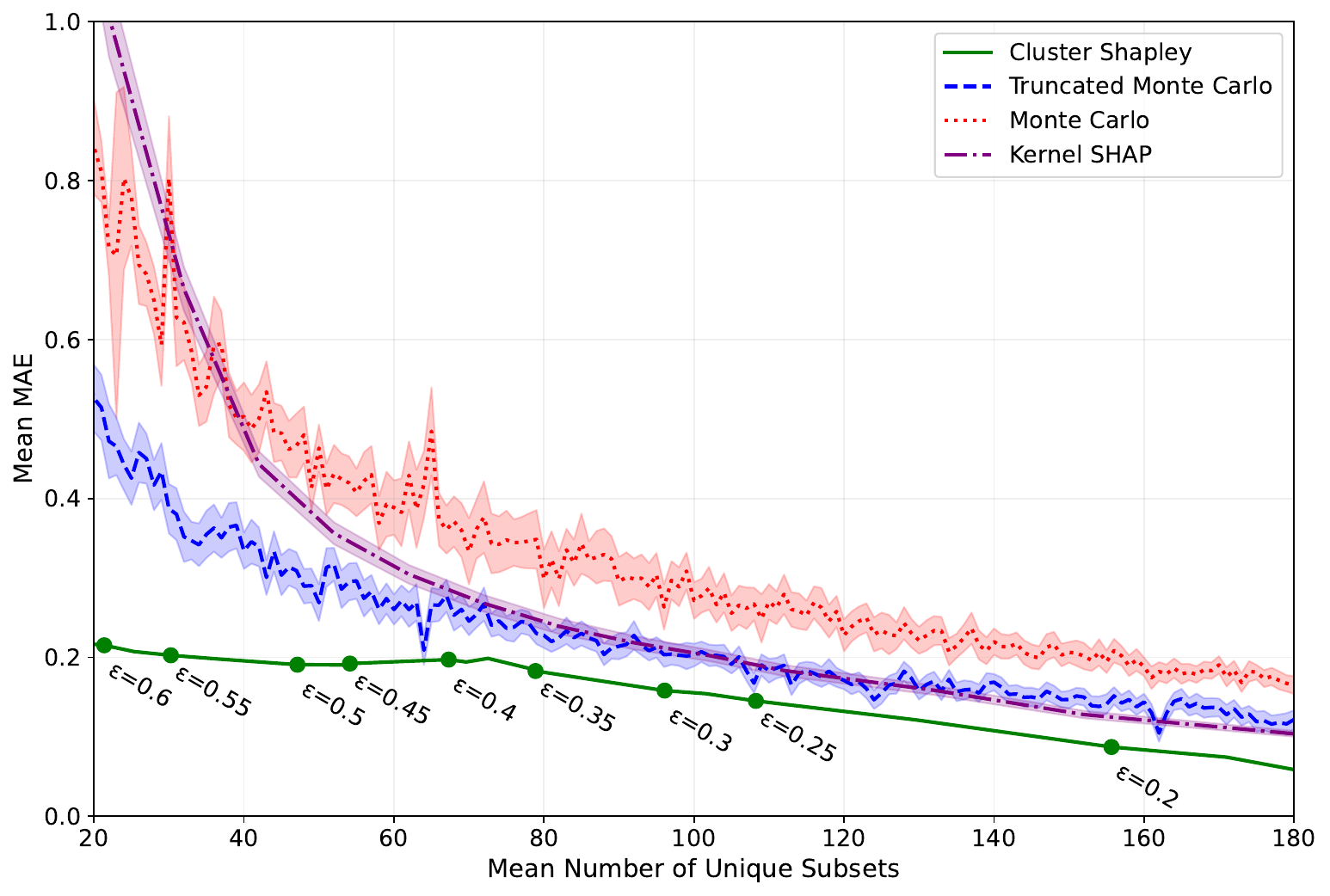}
    \caption{Efficient frontiers of Shapley approximation algorithms. The $x$-axis represents the number of unique subsets used by the algorithms, averaged across all test queries and reviews. The $y$-axis represents the Mean Absolute Error (MAE) of the Shapley values, averaged across all test queries and reviews. The points on the Cluster Shapley curve correspond to different clustering diameters $\epsilon$. For reference on the size of MAE, the average Shapley value over all test samples is 1.084, indicating that 0.2 MAE is around a 20\% percentage error. 95\% CIs for Monte Carlo, Truncated Monte Carlo, and Kernel SHAP are computed through 10 replications of the algorithms.}
    \label{fig:benchmark_comparison}
\end{figure}

We visualize the efficient frontier of different Shapley value approximation methods in Figure~\ref{fig:benchmark_comparison}. The $y$-axis represents the Mean Absolute Error (MAE) of the Shapley values, averaged across all test instances and reviews, which serves as a measure of the approximation error for each algorithm. Results under performance metrics, including Mean Squared Error (MSE) and Mean Absolute Percentage Error (MAPE), exhibit similar trends and can be found in Web Appendix $\S$\ref{appssec:mse_mape}. The $x$-axis represents the number of unique subsets used by the algorithms. Here, a ``unique subset'' refers to a distinct (non-replicated) subset of reviews used in the algorithm. For Cluster Shapley, this number represents the distinct cluster subsets that emerge after clustering, i.e., $2^m-1$ averaged over 48 test instances. For Monte Carlo, Truncated Monte Carlo, and Kernel SHAP, while these methods can sample the same subset multiple times, we count only the unique subsets encountered during sampling to ensure fair computational comparison. For example, if the same subset appears multiple times in these algorithms, we only evaluate it once and cache its evaluation score for reuse. Because the computation time for clustering in the Cluster Shapley algorithm, the additional time for evaluating larger meta-reviews, and the regression step in Kernel SHAP are all negligible compared to the cost of summarization and evaluation prompts, the overall computation cost of all algorithms effectively scales linearly with the number of unique subsets. Thus, this figure highlights the cost-effectiveness of the various algorithms, where cost is represented by the $x$-axis and effectiveness by the $y$-axis. The lower left region indicates more desirable outcomes-lower cost and higher effectiveness. For clarity, Figure~\ref{fig:benchmark_comparison} truncates the $x$-axis at 180 subsets rather than showing the full range up to 255. This choice reflects two considerations. First, all algorithms perform similarly beyond this point, and extending the axis would not add meaningful insight. Second, larger subset sizes imply higher computation costs, which are less relevant to practical applications. As such, our analysis focuses on the range where performance differences are most informative.

We now discuss the main takeaways from Figure~\ref{fig:benchmark_comparison}. First, we see that Cluster Shapley achieves the best overall performance across all algorithms. Its error curve consistently lies below those of other algorithms, indicating that it achieves lower approximation error for a given computation cost or, equivalently, requires less computation to reach the same level of accuracy. Truncated Monte Carlo ranks second, benefiting from its early stopping mechanism, which limits unnecessary evaluations when performance plateaus. Second, Cluster Shapley’s advantage is particularly pronounced when the computation budget is small, i.e., the number of unique subsets is relatively small. In this regime, it achieves an MAE of around 0.2, whereas other methods exceed 0.4. This efficiency arises from clustering's ability to capture core similarity patterns with relatively few clusters, which effectively reduces the dimensionality of the approximation problem. However, as the number of subsets increases (i.e., with smaller $\epsilon$), the advantage gradually diminishes. When the subset count exceeds 150, Cluster Shapley’s performance converges with other methods as single-review clusters emerge and the benefit of clustering is lost. Third, all algorithms show decreasing MAE as the number of subsets increases, consistent with theoretical expectations. However, their stability varies substantially. Monte Carlo and Truncated Monte Carlo exhibit noticeable fluctuations and wide confidence intervals due to their inherent sampling variability. In contrast, Cluster Shapley produces deterministic results across the computation range because we calculate the exact Shapley of clusters in Step 2. Kernel SHAP also uses sampling but reduces variance through its linear regression-based estimation, resulting in moderate stability. %Within Cluster Shapley, a plateau occurs when large $\epsilon$ values create coarse clusters that limit fine-grained grouping. As $\epsilon$ decreases, the clustering refines and the MAE steadily declines-until eventually reaching the diminishing returns phase described above.

In summary, our empirical results show that the proposed Cluster Shapley algorithm achieves substantially higher accuracy at lower computational costs by leveraging semantic similarity in LLM-based document embeddings. Unlike other approximation methods, such as Monte Carlo, Truncated Monte Carlo, and Kernel SHAP, that rely on random sampling without exploiting intrinsic semantic relationships, Cluster Shapley groups semantically similar documents to approximate cooperative contributions more efficiently. This scalability provides practitioners with a flexible toolkit to balance accuracy and computation cost according to platform constraints. More broadly, these findings highlight the value of incorporating semantic representations from LLMs into document valuation frameworks, enabling practical and principled approaches to fair value attribution at scale.

\subsection{Comparison with Simple Attribution Rules}
\label{ssec:simple rule}

%As shown earlier, the proposed Cluster Shapley method achieves the best efficiency–accuracy frontier among the tested approximation algorithms. 

So far, we have seen that the Cluster Shapley method achieves the best efficiency–accuracy frontier among all Shapley approximation algorithms. Nevertheless, one might argue that simple attribution rules that are not based on Shapley can be both cheap and effective. To that end, we now consider two simple attribution rules -- equal attribution and relevance-weighted attribution, and benchmark the performance of our approach against them.

\squishlist

\item \textbf{Equal Attribution:} This baseline splits the full-set value uniformly across documents, i.e., all documents in $S_q$ get equal attribution $\hat{\phi}^{\mathrm{Equ}}_i(q)=v(q,A(q,S_q))/|S_q|$. This baseline is simple to apply in practice, similar in spirit to Cloudflare’s pay-per-crawl approach~\citep{cloudflare2025}.

\item \textbf{Relevance-Weighted Attribution:}
This baseline allocates $v(q,A(q,S_q))$ proportionally to the relevance score between the query and each document:
\[
s_i(q)\;=\;\text{cosine\_similarity}(e_i, e_q),\qquad
w_i(q)\;=s_i(q)/\sum_{j\in S_q}s_j(q)\
\]
\[
\hat{\phi}^{\mathrm{Rel}}_i(q)\;=\;v(q,A(q,S_q))\,w_i(q),\qquad i\in S_q.
\]

\squishend

These two simple rules are costless to compute but can introduce substantial bias, as their assigned document values deviate widely from the fair values defined by Shapley. Such comparisons are crucial because platforms must weigh fairness against computational feasibility. In practice, a simple rule such as equal attribution can appear attractive for its negligible cost, as in Cloudflare’s pay-per-crawl model~\citep{cloudflare2025}, yet understanding the magnitude of the fairness loss is essential for responsible design.

Table~\ref{tab:accuracy_efficiency} quantifies this fairness–efficiency trade-off. Panel A reports attribution accuracy relative to the exact Shapley baseline, together with normalized computation cost (where 1 corresponds to the cost of computing all 255 subsets). Panel B displays the corresponding results for Cluster Shapley across different clustering diameters $\epsilon$, listed from coarse to fine (high $\epsilon$ to low), highlighting how accuracy improves as computational investment increases.  

The results exhibit a clear pattern: simpler attribution rules incur negligible computational cost but suffer from substantial bias, whereas more accurate methods require greater computation. In Panel~A, equal attribution yields an MAPE exceeding 150\%, indicating a large deviation from fair allocation, while relevance-weighted attribution achieves a lower but still significant 31.93\%. By contrast, even a very coarse Cluster Shapley with $\epsilon = 0.70$ achieves a smaller MAPE of 26.49\% at an extremely low cost, only 0.02 times that of the exact Shapley (about five of the 255 subset evaluations). Panel~B further illustrates the fairness--efficiency trade-off within Cluster Shapley: as $\epsilon$ decreases, computational cost increases while accuracy improves overall, though not strictly monotonically (e.g., $\epsilon=0.40$ is marginally worse than $\epsilon=0.50$), reflecting the stochasticity of LLM-based evaluation. For instance, with $\epsilon = 0.20$, Cluster Shapley achieves an MAPE of 11.85\% at 0.60 times the computational cost of the exact Shapley. These findings quantify a tangible ``cost of fairness'': achieving fair value attribution demands meaningful, though not prohibitive, computation.

For platform decision-makers, these results provide actionable guidance. When computational budgets are tight, the framework helps quantify how much fairness must be traded off for efficiency. Conversely, when fairness toward content contributors is a design priority, the same analysis clarifies the computational investment required. Although our experiments rely on limited or synthetic query data, the observed bias is likely heterogeneous across query types; once real user-level query distributions become available, the same framework can readily quantify this heterogeneity at scale.

Finally, Cluster Shapley should be viewed as a proof of concept rather than a prescriptive algorithm for deployment. Adapting it to specific LLM-based summarization architectures, or replacing it with improved approximation algorithms, may further enhance the efficiency frontier. The central insight is conceptual: the Shapley-based framework offers a principled way to evaluate and compare the computational cost of fairness in document-value attribution.

\begin{table}[htbp]
\centering
\small
\begin{threeparttable}
\begin{tabular}{cccrc}
\toprule
\multicolumn{5}{c}{\textbf{Panel A: Accuracy–Efficiency Comparison of Simple Attribution Rules and Exact Shapley}}\\
\midrule
\textbf{Method} & \textbf{MAE} & \textbf{MSE} & \textbf{MAPE} & \textbf{Computation Cost} \\
\midrule
Exact Shapley & 0 & 0 & 0 & 1.00 \\
Equal Attribution & 0.3284 & 0.3955 & 157.86\% & 0.00 \\
Relevance-Weighted & 0.3374 & 0.3893 & 31.93\% & 0.00 \\
\midrule
\multicolumn{5}{c}{\textbf{Panel B: Cluster Shapley's Accuracy–Efficiency Frontier Across $\epsilon$}}\\
\midrule
\textbf{Cluster Shapley $\epsilon$} & \textbf{MAE} & \textbf{MSE} & \textbf{MAPE} & \textbf{Computation Cost} \\
\midrule
%0.80 & 0.2259 & 0.2123 & 26.33\% & 0.01 \\
0.70 & 0.2305 & 0.1636 & 26.49\% & 0.02 \\
0.60 & 0.2152 & 0.1038 & 24.43\% & 0.08 \\
0.50 & 0.1908 & 0.1074 & 21.05\% & 0.18 \\
0.40 & 0.1972 & 0.1499 & 21.35\% & 0.26 \\
0.30 & 0.1617 & 0.1723 & 17.16\% & 0.38 \\
0.20 & 0.0913 & 0.0441 & 11.85\% & 0.60 \\
0.10 & 0.0507 & 0.0184 & 8.62\% & 0.73 \\
0.01 & 0.0381 & 0.0148 & 7.47\% & 0.77 \\
\bottomrule
\end{tabular}
\caption{Approximation error (averaged over all documents) and computation cost of simple attribution rules (Panel A) and Cluster Shapley under varying $\epsilon$ (Panel B). The last column, computation cost, is normalized relative to the exact Shapley, which requires evaluating all 255 unique subsets (assigned value 1). Other methods’ costs are expressed as a fraction of this baseline. Equal and relevance-weighted rules only require $v(q,A(q,S_q))$, hence their cost is close to 0 ($1/255\approx 0.00$). Note that when calculating MAPE, we add a small constant (0.1, approximately 10\% of the mean Shapley value) to the denominator to prevent near-zero Shapley values from inflating the error.}
\label{tab:accuracy_efficiency}
\end{threeparttable}
\end{table}

%\begin{table}[htbp]
%\centering
%\small
%\begin{threeparttable}
%\begin{tabular}{ccccc}
%\toprule
%\textbf{Clustering Diameter ($\epsilon$)} & \textbf{MAE} & \textbf{MSE} & \textbf{MAPE} & \textbf{Cost Reduction} \\
%\midrule
%0.01 & 0.0381 & 0.0148 & 7.47\% & 23.01\% \\
%0.10 & 0.0507 & 0.0184 & 8.62\% & 26.67\% \\
%0.20 & 0.0913 & 0.0441 & 11.85\% & 40.00\% \\
%0.30 & 0.1617 & 0.1723 & 17.16\% & 62.39\% \\
%0.40 & 0.1972 & 0.1499 & 21.35\% & 73.61\% \\
%0.50 & 0.1908 & 0.1074 & 21.05\% & 81.52\% \\
%0.60 & 0.2152 & 0.1038 & 24.43\% & 91.62\% \\
%0.70 & 0.2305 & 0.1636 & 26.49\% & 98.63\% \\
%0.80 & 0.2259 & 0.2123 & 26.33\% & 99.13\% \\
%\bottomrule
%\end{tabular}
%\caption{Approximation error (averaged over all documents) and computation time reduction of Cluster Shapley under varying $\epsilon$. The last column, cost reduction, is calculated as the percentage reduction in the number of unique subsets used compared to all 255 subsets used by the exact Shapley. Note that when calculating MAPE, we add a small constant (0.1, approximately 10\% of the mean Shapley value) to the denominator to prevent near-zero Shapley values from inflating the error.
%}
%\label{tab:efficiency_gains}
%\end{threeparttable}
%\end{table}

\subsection{Robustness Checks and Extensions}
\label{ssec:robustness}

We now present a series of robustness checks on various aspects of our approach and an extension that combines Cluster Shapley with other approximation algorithms. 

\squishlist
\item In the main analysis, we use GPT-4o for both summarization and for evaluating the summaries. This can potentially introduce bias because LLMs tend to give higher scores to their own summaries. To address this, we conduct an analysis where we use a different LLM, Claude, for evaluation. We find that Claude yields similar evaluation results and therefore similar Shapley values. See Web Appendix $\S$\ref{appssec:claude_evaluation} for details. These findings are consistent with a growing literature documenting substantial, though context-dependent, agreement between LLM judges and human preferences \citep{zheng2023judging, shankar2024validates}; a dedicated human-subject validation would further strengthen the case and is a natural direction for future work. Since the framework treats $v(\cdot)$ as a black box, a platform that already collects human feedback can use it directly in place of the LLM judge.

\item Document valuations are computed conditional on the retrieved set $S_q$, which depends on the top-$K$ retrieval depth ($K=8$ in the main analysis). We verify robustness to this choice (and, more broadly, to the retrieval step that selects $S_q$) by recomputing exact Shapley values at $K \in \{4,6,8\}$ and comparing the resulting document rankings. We find that these are highly stable across retrieval depths (median rank correlation of $1.00$). See Web Appendix~$\S$\ref{appssec:vary_k} for details.

\item The main analysis uses GPT-4o as the summarizer. To ensure that the results are not dependent on the summarizer choice, we re-run the pipeline with a much lighter summarizer (Gemini~2.5~Flash-Lite) while holding the evaluator fixed (Gemini~2.5~Flash scores the summaries from both summarizers). We find that the document Shapley values are broadly preserved in magnitude. See Web Appendix~$\S$\ref{appssec:vary_summarizer} for details.

\item For completeness, we also compare the approximation error of different algorithms using alternative metrics, including MSE and MAPE, and find that the results are consistent with those shown in Figure~\ref{fig:benchmark_comparison}. See Web Appendix~\S\ref{appssec:mse_mape} for additional details of this robustness check.

\item In the main analysis, we use our proposed adaptive DBSCAN (Algorithm~\ref{alg:dbscan}) to enforce tight distance constraints within clusters. To examine the impact of this design choice, we conduct a robustness check using the standard (non-adaptive) DBSCAN algorithm. We find that while standard DBSCAN performs reasonably well, its performance is consistently inferior to our proposed adaptive version. Details of this comparison are provided in Web Appendix~\S\ref{appssec:standard_DBSCAN}.

\item In the main analysis, we use all 48 test queries and report average performance across them. To assess robustness to query selection and hyperparameter tuning, we conduct a sample-splitting check by randomly dividing the queries into two halves and replicating the analysis for each split. The results, shown in Web Appendix~\S\ref{appssec:split_sample}, confirm consistent comparative performance across splits and stable hyperparameter choices.

\item As discussed in \S\ref{ssec:theory}, in large data settings, we can integrate Monte Carlo sampling into our Cluster Shapley to improve efficiency. We present additional experiments in Web Appendix~\S\ref{appssec:cluster_approx} demonstrating the performance of our approach for such settings. The first experiment considers a setting where the number of relevant documents $|S_q| = 10$. This setup reflects more computationally intensive, yet still feasible, scenarios in which exact Shapley values can be computed.  All other settings are consistent with the main analysis. The second experiment involves a setting where the number of relevant documents is extremely large, $|S_q| = 30$. This number is unlikely to be commonly used in practice; however, we consider this case as a proof of concept to demonstrate the scalability of our approach. For example, \citet{google_citation_count} reports that Google AI Overviews include 4–5 citations on average, with a maximum of 33 in rare cases. Since computing exact Shapley values is intractable at this scale, we synthetically construct ground-truth Shapley values and simulate the evaluation function $v(q, A)$. Across both experiments, the Monte Carlo–based Cluster Shapley demonstrates favorable accuracy–cost trade-offs, reinforcing its practicality in realistic and large-scale applications.
\squishend

\section{Conclusion}
\label{sec:conclusion}

The rapid integration of LLM-based summarization into digital platforms has reshaped how information is consumed and monetized, raising pressing concerns around attribution, compensation, and sustainability for content creators. This paper addresses a critical and timely challenge in this ecosystem: how to fairly and efficiently value the contribution of individual documents to LLM-generated summaries.

Our work makes two core contributions. First, we address the important and largely unsolved problem of source document valuation in the context of LLM-generated summaries: framing it as a compensation and incentive-design problem, we propose a principled, query-level valuation framework based on the Shapley value, which is fair by construction and agnostic to the summarization method and the evaluation mechanism, making it broadly applicable across search, review, and Q\&A platforms. Second, we introduce the Cluster Shapley algorithm, which improves the computational efficiency of Shapley value approximation by exploiting semantic similarity among documents; it admits a flexible efficiency--accuracy trade-off via a tunable clustering parameter, comes with theoretical guarantees, and substantially improves the empirical efficiency--accuracy frontier. Beyond these contributions, our results carry two broader messages: simple heuristic attribution rules, though computationally cheap, produce substantially unfair allocations, which offers concrete guidance for decision-makers weighing the cost of fairness; and Shapley approximation can be improved significantly in LLM applications when additional structure, such as embedding-based similarity, is available, pointing to a broader class of structure-aware approximation algorithms.

Our work serves as a foundational first step for fair content attribution in the generative AI ecosystem, helping balance platform goals with content creator incentives. Many avenues for extensions and future work remain open. One avenue is to adapt it to real-world LLM summarization systems with richer retrieval stacks, reranking, tool use, and reasoning, and to develop more dynamic, fine-grained valuation that accounts not only for document content but also for metadata such as authorship, credibility, and temporal relevance, as well as multi-turn conversational contexts where queries evolve over time. A second direction is to connect document valuation with reputation mechanisms, which are central to sustaining two-sided platforms \citep{resnick_etal_2000, cripps_etal_2004, yoganarasimhan_2013} yet hard to maintain without direct feedback on contributor performance: document value could augment or replace existing reputation signals. A third direction is to combine Shapley value with other notions of fairness. For instance, Shapley is agnostic to the cost associated with creating content. In practice, a news article from a war-zone can be much costlier to write than covering a local movie screening. One possibility is to extend our framework to incorporate economic costs, and other notions of fairness (e.g., ensuring coverage of certain subsets of topics). Finally, while Cluster Shapley is an effective first approximation, future work could investigate hybrid methods that combine semantic clustering with adaptive sampling or reinforcement learning to further improve efficiency and attribution fidelity.

\section*{Funding and Competing Interests Declaration}
Author(s) have no competing interests to declare.

%\clearpage
{\small  % \footnotesize: smaller
% ==== arXiv/SSRN: bibliography inlined from bu1.bbl (no bibtex / no .bbl needed) ====

% ==== end inlined bu1 ====
}
\end{bibunit}

\newpage
%\begin{refsection}
\setcounter{table}{0}
\setcounter{figure}{0}
\setcounter{equation}{0}
\setcounter{page}{1}
\renewcommand{\thetable}{A\arabic{table}}
\renewcommand{\thefigure}{A\arabic{figure}}
\renewcommand{\theequation}{A\arabic{equation}}
\renewcommand{\thepage}{\roman{page}}
\singlespacing
\renewcommand{\thesection}{\Alph{section}}
\pagenumbering{roman}

\begin{bibunit}

\begin{appendices}

\section{Illustration Examples of AI-Generated Summaries}
\label{appsec:ai_generated_overviews}

We now present more examples of AI/LLM-generated summaries, showing how it is changing the traditional search industry, Q\&A, and e-commerce websites.

\begin{figure}[htbp]
    \centering
    \includegraphics[width=1\textwidth]{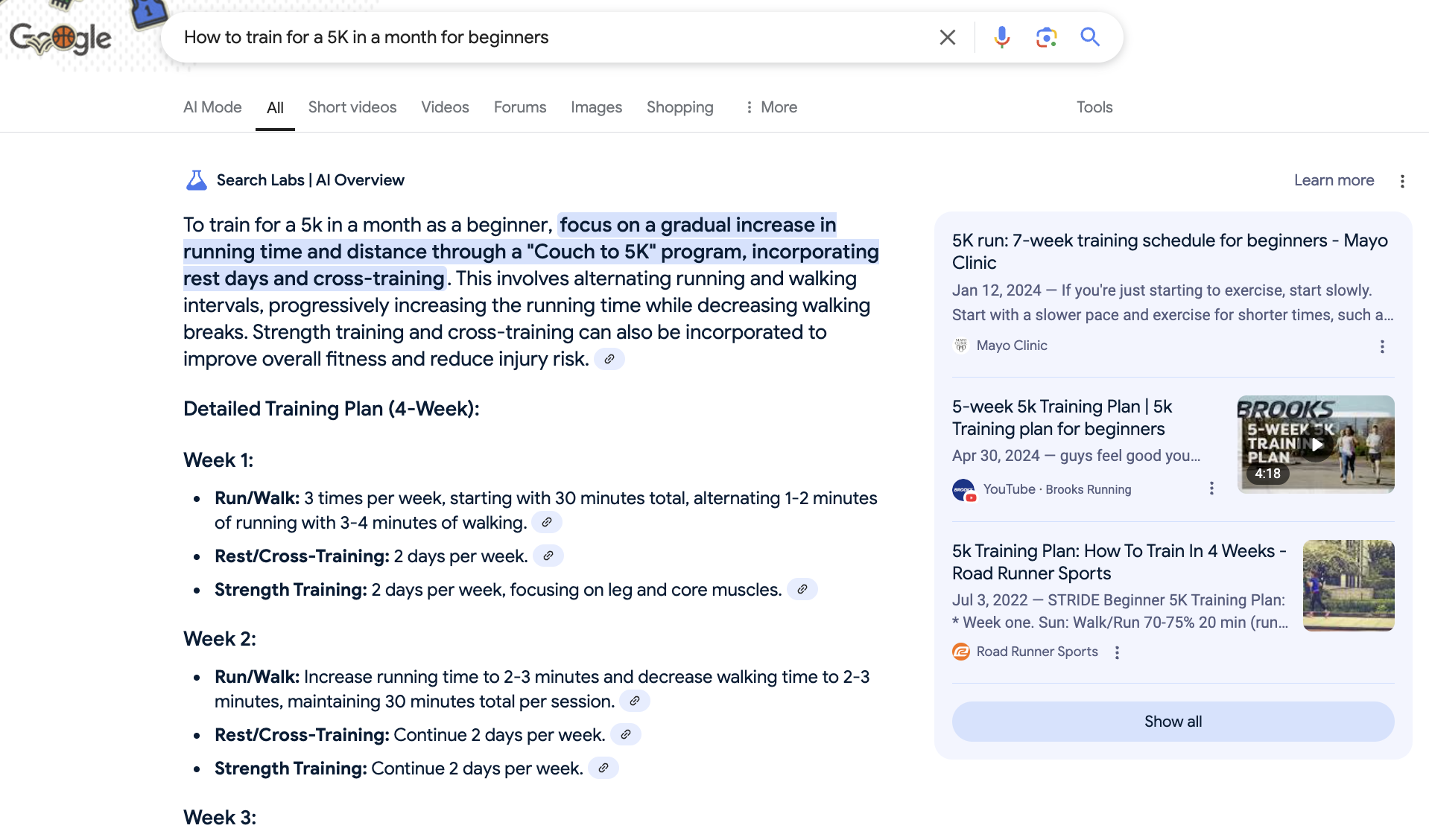}
    \caption{Google AI Overview}
    \label{fig:google_overview}
\end{figure}

\begin{figure}[htbp]
    \centering
    \includegraphics[width=1\textwidth]{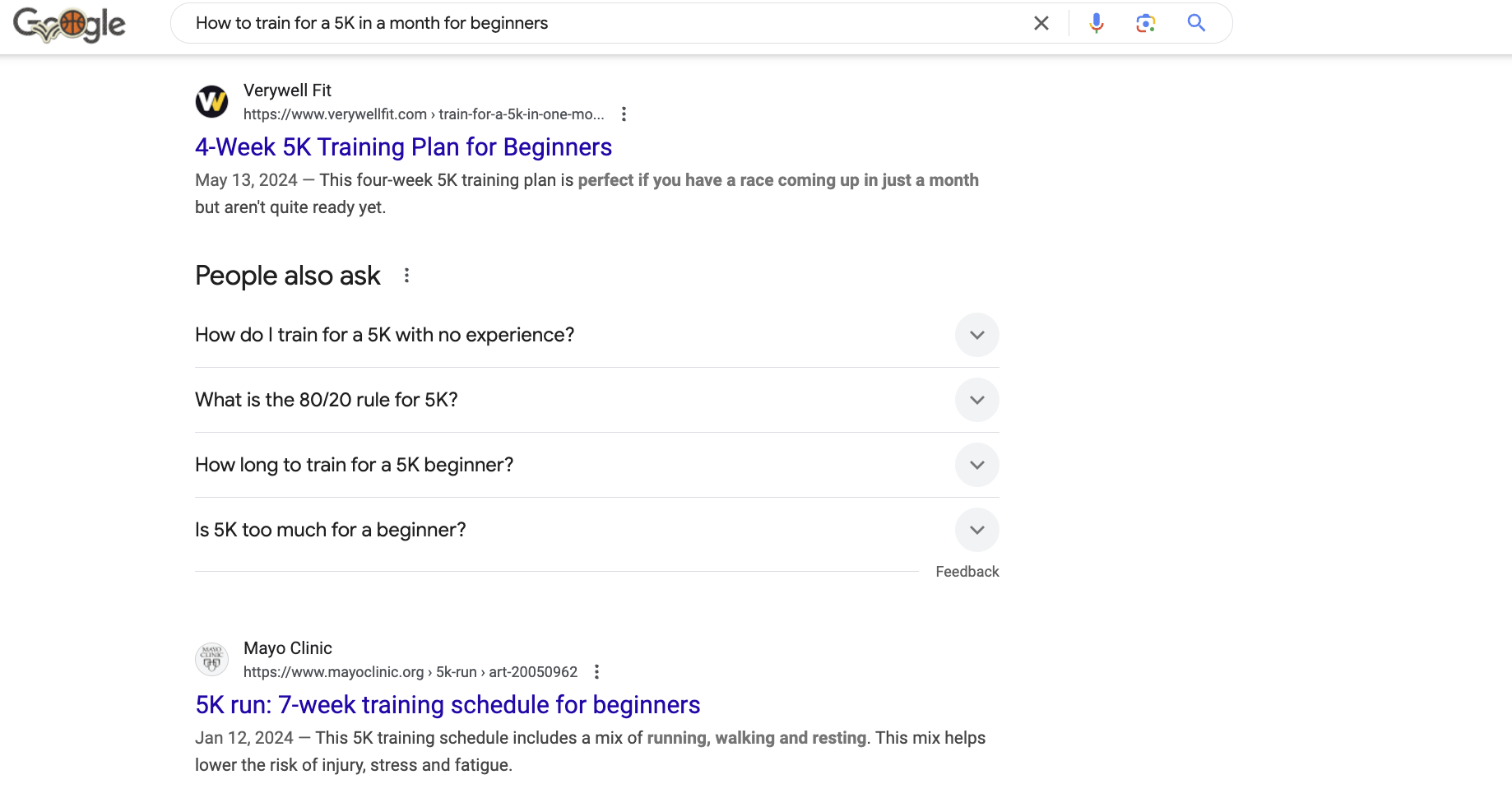}
    \caption{Google Search without AI Overview}
    \label{fig:google_search}
\end{figure}

Figure~\ref{fig:google_overview} illustrates Google’s AI Overview in response to the user query ``How to train for a 5K in a month for beginners.'' The system returns a detailed 4-week training plan synthesized from multiple web pages, with reference links displayed on the right-hand side. Each step in the summary also includes links to the original sources, enabling users to verify the information and credit the content providers. In contrast, Figure~\ref{fig:google_search} shows a traditional Google Search results page, where users must manually click through a ranked list of relevant websites to extract and compile information on their own. 

\begin{figure}[htbp]
    \centering
    \includegraphics[width=\textwidth]{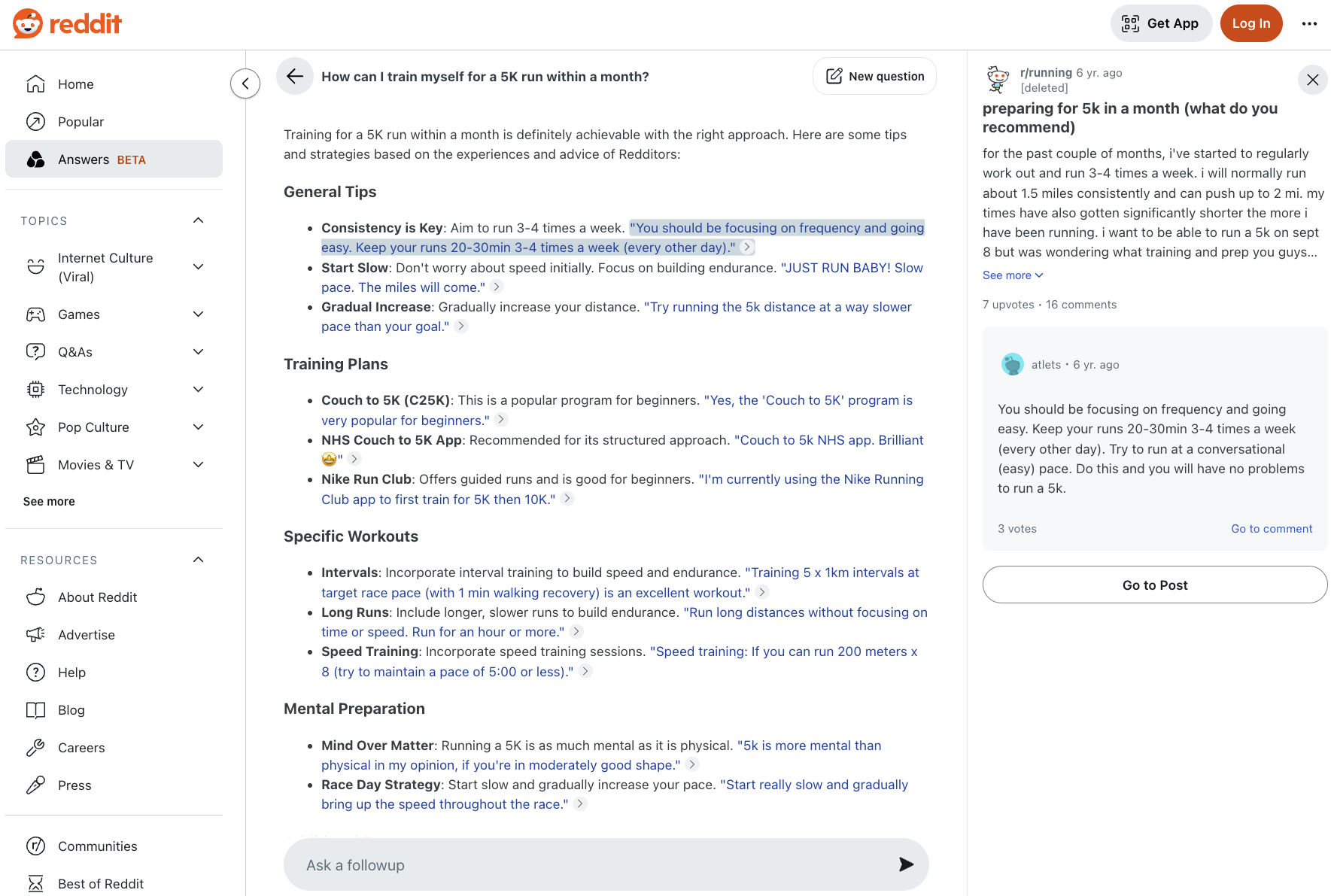}
    \caption{Reddit AI-Generated Answer}
    \label{fig:reddit_demo}
\end{figure}

The next example, shown in Figure~\ref{fig:reddit_demo}, comes from Reddit, a Q\&A forum. Reddit is currently piloting a new RAG-enhanced Q\&A assistant called \textit{Reddit Answers} \citep{reddit2025answers}. Using the same query, ``How to train for a 5K in a month for beginners?'', Reddit Answers generates a summarized response based on user-generated content from relevant Reddit threads. Similar to Google’s AI Overview, the summary is accompanied by links to the original responses, though the content is limited to the Reddit platform.

\begin{figure}[htbp]
    \centering
    \includegraphics[width=0.9\textwidth]{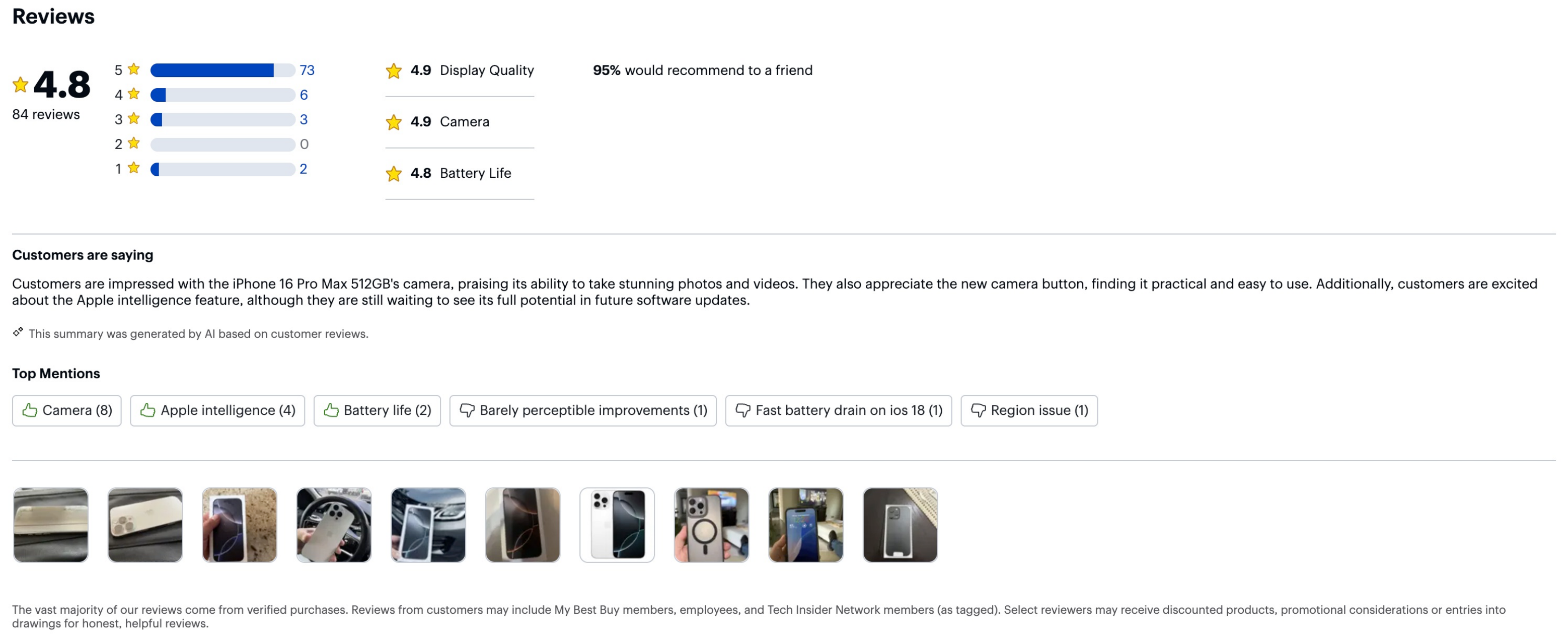}
    \caption{Best Buy AI-Generated Product Review}
    \label{fig:bestbuy_demo}
\end{figure}

\begin{figure}[htbp]
    \centering
    \includegraphics[width=1\textwidth]{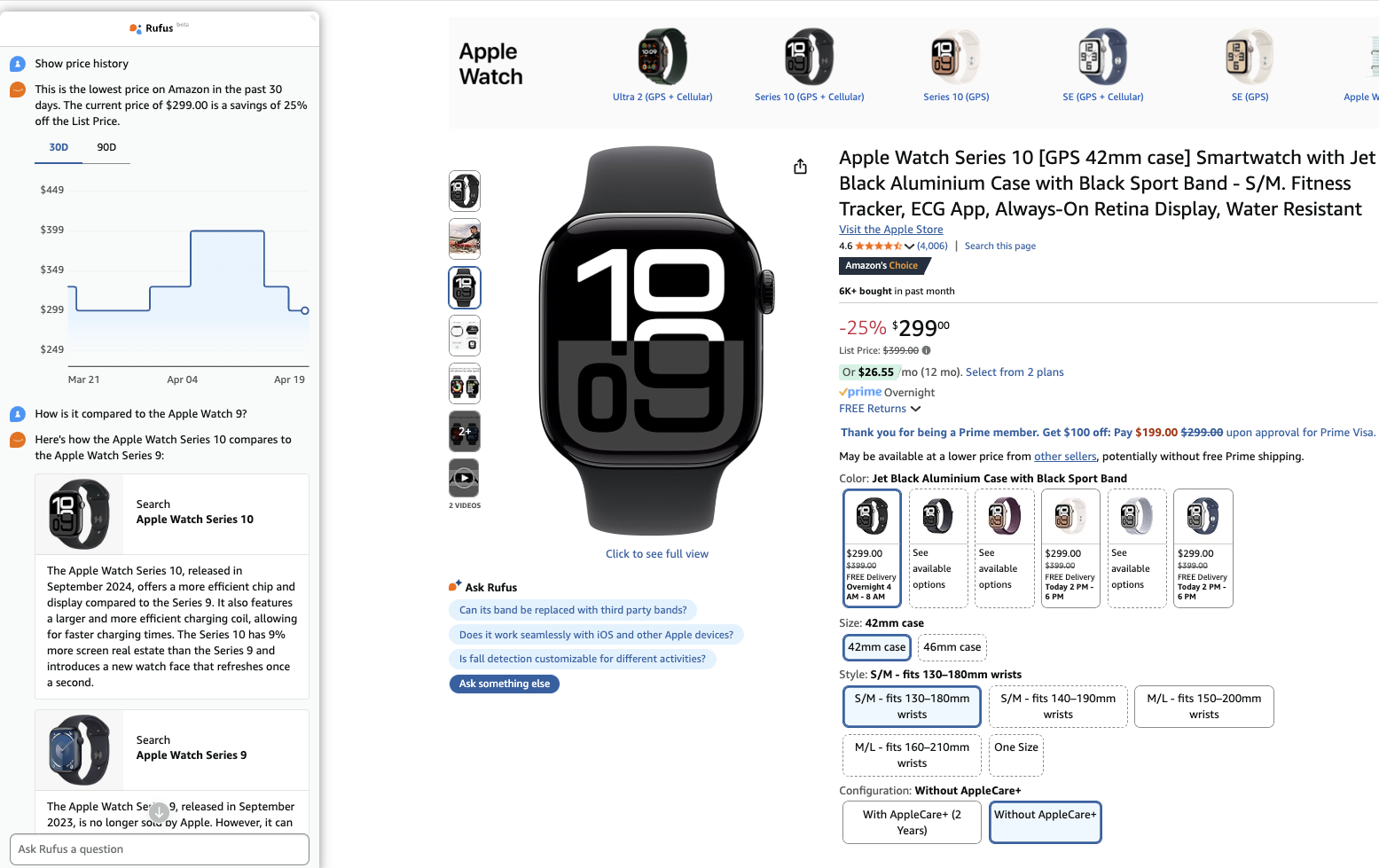}
    \caption{Amazon AI-Powered Shopping Assistant - Rufus (since renamed Alexa for Shopping)}
    \label{fig:rufus}
\end{figure}

AI assistants are also increasingly being integrated into e-commerce platforms. For example, Amazon now provides summarized product reviews directly on the product page, as shown in Figure~\ref{fig:amazon_review}. Similarly, Best Buy offers the same function, as illustrated in Figure~\ref{fig:bestbuy_demo}. In this iPhone example, the summary highlights user mentions of the improved camera, the new camera button, and anticipation for future updates related to Apple Intelligence. In addition to overall review summaries, Amazon's AI shopping assistant, Alexa for Shopping (formerly Rufus), enables users to ask specific questions about products. As shown in Figure~\ref{fig:rufus} (the screenshot shows the earlier Rufus interface), users can inquire about price history, product features, customer reviews, or comparisons with other products, and receive responses grounded in the information available on the product page.

\section{Supplementary Materials to Section~\ref{sec:shapley_framework}}
\label{appsec:cluster_shapley}

\subsection{Formal Axiomatic Characterization of the Shapley Value}
\label{appsec:axioms}

The following formal statement of the four Shapley axioms, adapted to our LLM-summarization setting, complements the concise treatment in $\S$\ref{ssec:shapley_intro}. Following the standard Shapley literature, we present the four properties of $\phi_i(q)$ that together guarantee \textbf{fair document valuation}:

\begin{enumerate}
    \item \textbf{Efficiency.} Efficiency ensures that the total value generated by the summarized document is fully distributed among the documents, with no surplus or deficit. Mathematically, this is represented as:
    \begin{equation}
    \sum_{i \in S_q} \phi_i(q) = v(q,A(q,S_q)).
    \end{equation}
    Because documents outside $S_q$ are null players for query $q$ (with $\phi_i(q)=0$), this is equivalent to summing over all documents in $D$, $\sum_{i\in D}\phi_i(q)=v(q,A(q,S_q))$. If the efficiency property is not enforced, the value function $\phi$ is only determined up to a proportional constant (see Theorem 2.1 in \cite{ghorbani2019data}). We adopt the standard normalization $v(q,A(q,\emptyset))=0$: the empty document set is not among the $2^{n}-1$ evaluated subsets and is assigned value zero by convention (there is no summary to score). Efficiency then equates the sum of the document values to the full summary value $V_q=v(q,A(q,S_q))$; we assume $V_q>0$ (the retrieved set yields a summary of positive value, which holds whenever at least one retrieved document is relevant), so it is well defined as the normalizing denominator for the revenue allocation in $\S$\ref{ssec:revenue_attribution} and Corollary~\ref{cor:revenue}.

    \item \textbf{Symmetry.} Symmetry ensures that documents with equal contributions are valued equitably. That is, two documents $i$ and $j$ have the same valuation if they contribute equally to every possible coalition. Formally:
    \begin{equation}
    \phi_i(q) = \phi_j(q), \;\; \textrm{if} \;\;  v(q, A(q, S \cup \{i\})) = v(q, A(q,S \cup \{j\})) \;\;  \forall \;\;  S \subseteq D \setminus \{i,j\}.
    \end{equation}

    \item \textbf{Null Document.} A null document implies that if a document provides no marginal value to any subset of documents, its value is zero. Formally, a document $i$ in a query $q$ is called null if $v(q, A(q, S \cup \{i\})) = v(q, A(q,S))$ for subsets \( S \subseteq D \setminus \{i\} \). If document $i$ for query $q$ is null, then the value $\phi_i(q)=0$.

    In our setting, any document $i$ not used in the summarization process for query $q$ has zero value for that query. That is, $\phi_i(q)=0$, if $i \notin S_q$.

    \item \textbf{Linearity.} The values of document $i$ under two separate queries \( q_1 \) and \( q_2 \), sum up to its value when evaluated using a performance score function that combines the individual performance score functions. Formally:
    \begin{equation}
    \phi_i(q_1 + q_2) = \phi_i(q_1) + \phi_i(q_2),
    \end{equation}
    where $q_1+q_2$ represents a combination of two queries, and the performance score function for this combined query is naturally defined as $v(q_1, A(q_1, S))+v(q_2, A(q_2, S))$, reflecting the aggregate contributions of \( q_1 \) and \( q_2 \).

    Note that the combination of two queries $q_1 + q_2$ does not imply that the two queries are merged into a single new query. Instead, it represents a setting where there are two distinct queries being processed (this definition naturally extends to any finite number of queries, not just two). For example, consider two queries: one on quality ($q_1$) and another on price ($q_2$). The value of document $i$ under the quality query is denoted by $\phi_i(q_1)$, calculated using the performance score function $v(q_1, A(q_1, S))$. Similarly, $\phi_i(q_2)$ represents the value of document $i$ under the price query, based on the performance score function $v(q_2, A(q_2, S))$. The linearity property asserts that if we sum the values obtained from these separate queries, i.e., $\phi_i(q_1) + \phi_i(q_2)$, the result is equivalent to the value of document $i$ calculated under a new, combined performance score function $v(q_1, A(q_1, S)) + v(q_2, A(q_2, S))$. This property ensures that value attributions are invariant to the merging of multiple queries, which has important implications for revenue attribution, as discussed in \S\ref{ssec:revenue_attribution}.
\end{enumerate}

While other desirable properties are worth discussing, these four -- Efficiency, Symmetry, Null Document, and Linearity -- are sufficient to uniquely determine the document value function $\phi_i(q)$, known as Shapley value.\footnote{These four properties (axioms) are independent of any specific summarization $(A)$ or evaluation $(v)$, which are laid out in our problem definition. These properties originate from Shapley's work on cooperative game theory \citep{shapley1953value}.} This is a foundational result in cooperative game theory. We refer interested readers to Shapley's seminal paper for a formal proof.

\subsection{Document Clustering with the Distance Constraint}
\label{appsec:step1}

We now discuss the document clustering step (Step 1). The general oracle bound in Theorem~\ref{thm:cluster_shap_general} is valid for any query-specific partition. The distance constraint is nevertheless useful because it makes the deterministic coarsening bias interpretable and controllable when semantically similar documents have similar marginal contributions: under the Lipschitz condition stated and validated in Web Appendix~\S\ref{appssec:assumption}, the specialization in Web Appendix~\S\ref{appssec:lipschitz_specialization} bounds the within-cluster dispersion of exact Shapley values by $L\epsilon$, the product of the Lipschitz constant and the clustering diameter. Therefore, unlike the standard clustering algorithm, we impose a distance constraint that documents within the same cluster should be strictly close to each other, with a distance less than \( \epsilon \); that is, if \( i, j \in G_k \), then \( d(e_i, e_j) \leq \epsilon \). We also find that this constraint can improve the empirical performance of the Cluster Shapley algorithm; see more details in Web Appendix $\S$\ref{appssec:standard_DBSCAN}. Intuitively, a smaller \( \epsilon \) results in more clusters (larger \( m \)), leading to a more accurate Shapley estimation at the cost of increased computation. In the extreme case, we can set \( \epsilon \) to be the smallest distance between any pair of documents, \( \epsilon < \min_{i, j \in S_q, i \neq j} d(e_i, e_j) \), which yields clusters where each cluster contains only one document. In this case, our proposed algorithm reduces to the exact Shapley calculation. Therefore, tuning \( \epsilon \) appropriately is essential, as it balances the trade-off between computational efficiency and the approximation error induced by clustering. We formalize this statement in $\S$\ref{ssec:theory}.

To achieve this clustering goal, we propose Algorithm \ref{alg:dbscan}, which is essentially an adaptive version of the DBSCAN algorithm. The main advantage of DBSCAN is that it is non-parametric and clusters documents based on density rather than requiring parameters like the number of clusters (which algorithms such as K-Means need). This makes DBSCAN particularly suitable for our task, where the number of clusters is not predetermined. The standard DBSCAN operates through a density-based clustering mechanism, utilizing two key hyperparameters: $r$ (the neighborhood radius) and $\texttt{MinPts}$ (minimum points required to form a dense region). The algorithm identifies core points as those having at least $\texttt{MinPts}$ points within their $r$-neighborhood and constructs clusters through density-reachability -- a property where points are connected through a chain of core points. Points that fall within the $r$-neighborhood of a core point but do not qualify as core points themselves are classified as border points, while points that fulfill neither criterion are designated as noise. However, the standard DBSCAN does not guarantee that any two documents within the same cluster have strictly smaller distances than $r$ because DBSCAN forms clusters based on local density connectivity rather than enforcing global distance constraints. Consider three documents $i$, $j$, and $k$: if $d(e_i, e_j) \leq r$ and $d(e_j, e_k) \leq r$, DBSCAN will assign all three points to the same cluster through density-reachability, even if $d(e_i, e_k) > r$. This transitive clustering property can result in clusters where some document pairs exceed the $r$ threshold. It means that if we set the radius the same as our clustering diameter $\epsilon$, the distance between two documents within the same cluster may exceed $\epsilon$. This limitation necessitates modifications to the standard DBSCAN algorithm to enforce a global distance constraint for accurate Shapley value estimation in our context.

In our proposed Algorithm~\ref{alg:dbscan}, we calculate the distance matrix using document embeddings and a predefined distance function \( d \), and input this matrix into the DBSCAN algorithm. We set the minimum number of points per cluster to 1 (i.e., a DBSCAN hyperparameter $\texttt{MinPts} = 1$) to ensure no document is excluded as noise. However, standard DBSCAN does not guarantee that all documents within a cluster are within \( \epsilon \) (the distance constraint we want to satisfy), nor does it ensure sufficient separation between clusters. To address this, we distinguish between two thresholds: the clustering diameter \( \epsilon \), which defines our desired upper bound on intra-cluster distances, and the DBSCAN neighborhood radius \( r \), which governs the clustering procedure. Initially, we set \( r \gets \epsilon \), run DBSCAN, and then check whether all document pairs within each cluster satisfy the global diameter constraint \( d(e_i, e_j) \leq \epsilon \). If any cluster violates this condition, we adaptively tighten the local neighborhood by reducing \( r \) by a factor of \( \alpha = 0.95 \), i.e., \( r \leftarrow 0.95r \), and rerun DBSCAN until all clusters satisfy the diameter constraint.

\begin{algorithm}[ht]
\small{
\caption{Adaptive Distance-Constrained DBSCAN}
\label{alg:dbscan}
\begin{algorithmic}
\State \textbf{Input:} Distance matrix $M$ with $M_{ij}=d(e_i,e_j)$. Hyperparameters: clustering diameter $\epsilon$ defined in Cluster Shapley Algorithm, neighborhood radius $r$, $\texttt{MinPts} = 1$, scaling factor $\alpha = 0.95$
\State Initialize the neighborhood radius $r \gets \epsilon$ \Comment{Start with the original $\epsilon$}

\While{true} \Comment{Iterate until all clusters satisfy the distance constraint}
    \State \textbf{Run the standard DBSCAN with $\texttt{MinPts} = 1$:} 
    \begin{itemize}
        \item For each document $i$: find $r$-neighborhood $N_{r}(i) = \{j: M_{ij} \leq r\}$.
        %\item If $|N_{r}(i)| \geq \texttt{MinPts}$, mark $i$ as a core point.
        \item Connect points that are within $r$ distance.
        %\item Assign non-core points to clusters of nearby core points.
    \end{itemize}
    \State \textbf{Check the distance constraint:}
   \begin{itemize}
    \item Check all clusters: $d(e_i, e_j) \leq \epsilon$ for all documents $i, j$ in the same cluster.
    \item If all clusters satisfy the distance constraint, exit the loop. Otherwise, update $r \gets \alpha \cdot r$  and continue the loop.
\end{itemize}
\EndWhile

\State \textbf{Output:} Return clusters such that $d(e_i, e_j) \leq \epsilon$ for all documents $i, j$ in the same cluster.
\end{algorithmic}}
\end{algorithm}

\subsection{Lipschitz Continuity Assumption and Its Empirical Validation}
\label{appssec:assumption}

The oracle bound in Theorem~\ref{thm:cluster_shap_general} does not require smoothness in the embedding space. For the interpretable specialization in Web Appendix~\S\ref{appssec:lipschitz_specialization}, we use the following condition.

\begin{assumption}[Lipschitz continuity in embedding space]
\label{assump:lipschitz}
There exists a constant $L>0$ such that for any two documents $i,j \in S_q$ satisfying $d(e_i,e_j)\le \epsilon$, the difference in their marginal contributions is bounded by their embedding distance:
\begin{equation*}
\Big|\big(v(S \cup \{i\}) - v(S)\big) - \big(v(S \cup \{j\}) - v(S)\big)\Big| \le L\,d(e_i, e_j),
\end{equation*}
for any coalition $S \subseteq S_q \setminus \{i, j\}$.
\end{assumption}

\begin{figure}[htbp]
    \centering
    \includegraphics[width=0.9\textwidth]{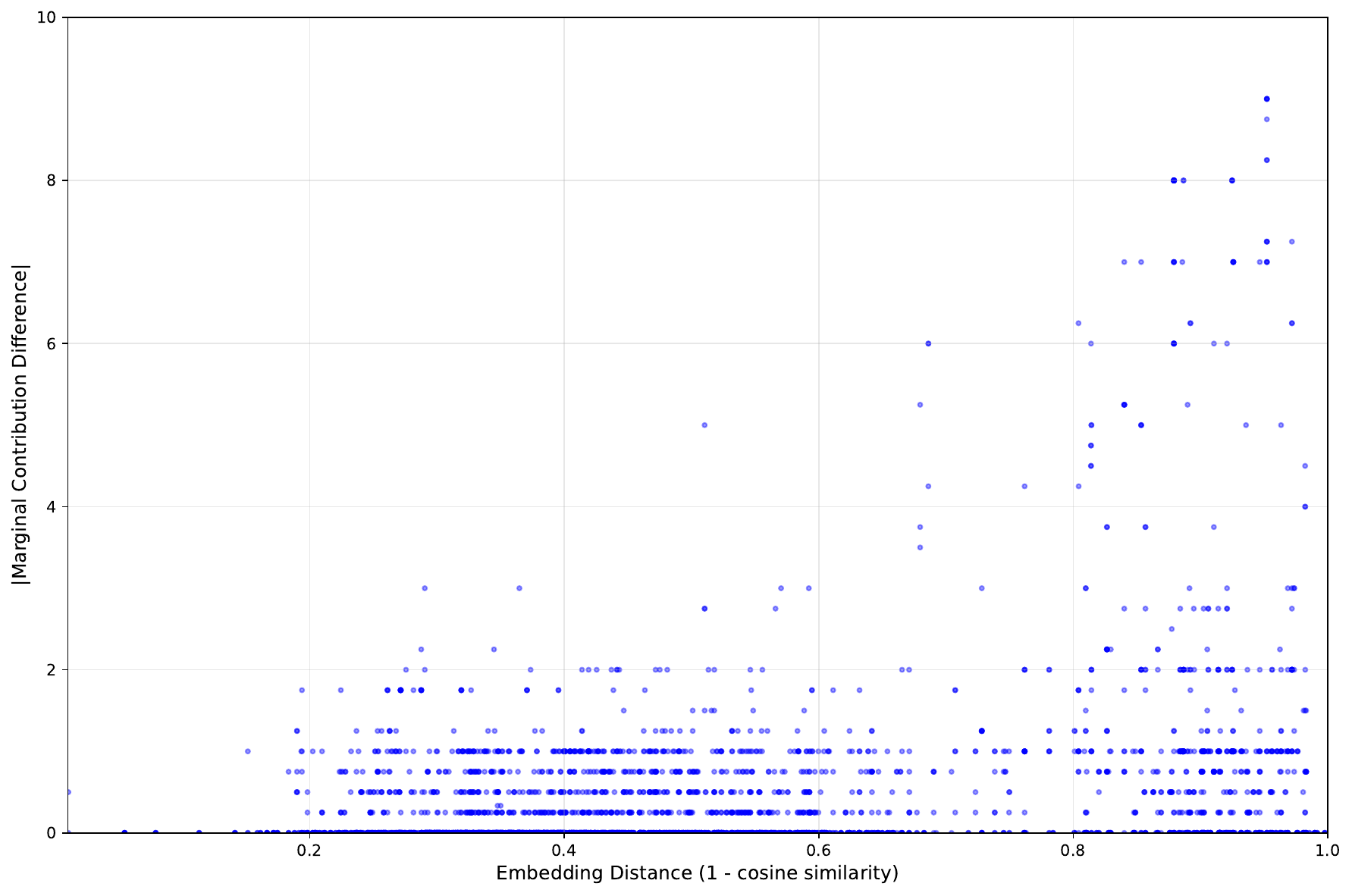}
    \caption{Empirical validation of Lipschitz continuity assumption. Each point represents a document pair $(i,j)$ and a coalition $S$ combination. The data encompasses all 48 queries in our dataset, and each query has 8 documents. Note that we sampled 5,000 data points for better visualization when the total number of points is $48\times 2^{(8-2)} \times \binom{8}{2} = 86,016$.}
    \label{fig:lipschitz_validation}
\end{figure}

In this appendix, we empirically examine Assumption~\ref{assump:lipschitz} (Lipschitz continuity in embedding space). The results are shown in Figure \ref{fig:lipschitz_validation}. The x-axis represents the embedding distance $d(e_i, e_j) = 1 - \text{cosine similarity}(e_i, e_j)$ between document pairs $i$ and $j$, where $e_i$ and $e_j$ are their respective text embeddings. The y-axis shows the absolute difference in marginal contributions $|v(S \cup \{i\}) - v(S) - (v(S \cup \{j\}) - v(S))|$ across all possible coalitions $S \subseteq S_q$ that do not contain documents $i$ or $j$.

The scatter plot indicates that marginal contribution differences are approximately linearly bounded by embedding distances, providing empirical support for the Lipschitz continuity assumption in the local, high-similarity regime (small embedding distance) most relevant to clustering. The data reveals a clear pattern: document pairs with small embedding distances (high similarity) consistently exhibit small differences in their marginal contributions across coalitions. Conversely, document pairs with large embedding distances show a wider range of marginal contribution differences, from small to large values. This asymmetric relationship aligns with our theoretical expectations and practical intuition. When two documents have similar embeddings, they naturally contribute similarly to any coalition, resulting in consistently small marginal contribution differences. However, when documents have dissimilar embeddings, their marginal contribution differences can vary significantly depending on the specific aspects they emphasize. For instance, two very different documents describing distinct aspects of a product (e.g., technical specifications versus user experience) may still provide comparable value to certain coalitions, leading to small marginal contribution differences despite large embedding distances.

Next, we further quantify this relationship. Recall that the y-axis represents the difference in marginal contributions, defined as:
\begin{equation}\nonumber
    \Delta_{i,j}(S) = \big|\big(v(S \cup \{i\}) - v(S)\big) - \big(v(S \cup \{j\}) - v(S)\big)\big|.
\end{equation}
Since our evaluation prompt restricts the performance score $v(\cdot)$ to the integer range $[0,10]$, $\Delta_{i,j}(S)$ is bounded by $10$, and the observed cosine distances lie in $[0,1]$ (review embeddings have non-negative pairwise cosine similarity). A bounded score therefore yields a constant ceiling on $\Delta_{i,j}(S)$, but such a constant bound is uninformative at small distances; what Assumption~\ref{assump:lipschitz} requires is a \emph{linear} bound $\Delta_{i,j}(S)\le L\,d(e_i,e_j)$ that also controls the small-distance regime. However, the validity of Assumption \ref{assump:lipschitz} hinges on the \textit{local} behavior of this difference as the embedding distance approaches zero. If the ratio of differences to distances were unbounded, we would observe data points filling the upper-left region of the plot (i.e., small distance but large difference).

Instead, Figure \ref{fig:lipschitz_validation} shows a dense concentration of points near the origin. To quantify this, we examine the data points within the region of high semantic similarity (specifically, embedding distance $d(e_i, e_j) \in [0, 0.4]$). In this range, we find that more than 95\% of the data points satisfy the condition:
\begin{equation}\nonumber
    \Delta_{i,j}(S) \le 2.5 \times d(e_i, e_j).
\end{equation}
This indicates that for the vast majority of cases, the effective Lipschitz constant is approximately $L \approx 2.5$, which is significantly tighter than the uninformative constant ceiling. A small fraction of outliers exists due to the inherent stochasticity of LLM scoring (e.g., randomness in generation or evaluation).

%The observed upper bound behavior-where embedding distance provides a ceiling for marginal contribution differences-validates the Lipschitz continuity assumption while capturing the nuanced relationship between semantic similarity and functional contribution in our evaluation framework. This empirical evidence supports the use of embedding-based distances as meaningful constraints in Shapley value computations.

\subsection{Proof of Theorem~\ref{thm:cluster_shap_general}}
\label{appssec:proof_3}

This and the following subsections present the proofs and supporting results for the theoretical analysis of \S\ref{ssec:theory}: \S\ref{appssec:lipschitz_specialization} develops the Lipschitz specialization of the coarsening bias, \S\ref{appssec:proof_exact_cluster} records the exact cluster-level special case, and \S\ref{appssec:representative_variant} presents a representative-imputation variant.

\begin{proof}
Fix a query $q$ and suppress $q$ from the notation. Let $i\in G_k$. By Step 3 of Algorithm~\ref{alg:cluster shap},
\[
\tilde\phi_i=\frac{\tilde\phi^{\mathcal G}_{G_k}}{|G_k|},
\qquad
\psi_i^{\mathcal G}=\frac{\phi^{\mathcal G}_{G_k}}{|G_k|}.
\]
Therefore,
\[
|\tilde\phi_i-\phi_i|
\le
\left|\frac{\tilde\phi^{\mathcal G}_{G_k}-\phi^{\mathcal G}_{G_k}}{|G_k|}\right|
+
|\psi_i^{\mathcal G}-\phi_i|.
\]
The first term is at most $\delta_{\mathcal M}(q)/|G_k|$ by the definition of $\delta_{\mathcal M}(q)$, and the second term equals $b_i(\mathcal G,q)$ (with $\psi_i^{\mathcal G}=\phi^{\mathcal G}_{G_k}/|G_k|$). This proves the deterministic bound~\eqref{eq:general_error_bound}; the high-probability version follows because $\delta_{\mathcal M}(q)\le\Delta_{\mathcal M}(\eta)$ on an event of probability at least $1-\eta$.

We also record the aggregate versions of the bound. With $B_\infty(\mathcal G,q)=\max_{i\in S_q}b_i(\mathcal G,q)$ and $s_{\min}=\min_{k\in[m]}|G_k|$, taking the maximum over $i$ in the per-document bound gives the $\ell_\infty$ bound:
\[
\|\tilde{\boldsymbol\phi}-\boldsymbol\phi\|_\infty
\le
\max_i b_i(\mathcal G,q)+\max_i \frac{\Delta_{\mathcal M}(\eta)}{|G_{k(i)}|}
=
B_\infty(\mathcal G,q)+\frac{\Delta_{\mathcal M}(\eta)}{s_{\min}}.
\]
For the mean $\ell_1$ bound,
\[
\frac{1}{n}\sum_{i\in S_q}|\tilde\phi_i-\phi_i|
\le
B_1(\mathcal G,q)
+
\frac{\Delta_{\mathcal M}(\eta)}{n}\sum_{k=1}^m\sum_{i\in G_k}\frac{1}{|G_k|}
=
B_1(\mathcal G,q)+\frac{m}{n}\Delta_{\mathcal M}(\eta).
\]
For complexity, Step 1 requires $O(n^2)$ time to compute the pairwise distance matrix and perform the constrained clustering. Step 2 applies $\mathcal M$ to the $m$-cluster game and costs $C_{\mathcal M}(m)$. Step 3 is linear in $n$. Hence the total complexity is $O(n^2+C_{\mathcal M}(m))$.
\end{proof}

\subsection{Lipschitz Specialization of the Coarsening Bias}
\label{appssec:lipschitz_specialization}

The oracle bias $b_i(\mathcal G,q)$ in Theorem~\ref{thm:cluster_shap_general} is deliberately defined without imposing assumptions on the value function. Under Assumption~\ref{assump:lipschitz}, it can be further decomposed into a within-cluster dispersion term and a meta-document aggregation term, yielding an explicit bound in terms of the clustering diameter.

For each cluster $G_k$, define
\begin{equation*}
\Gamma_k(q)=\left|\phi^{\mathcal G}_{G_k}(q)-\sum_{j\in G_k}\phi_j(q)\right|.
\end{equation*}
This term measures how much the Shapley value of the cluster in the cluster-level game differs from the sum of its members' exact Shapley values in the original document-level game. It is zero for singleton clusters, but in general it captures the deterministic effect of replacing a set of documents by a single meta-document.

\begin{prop}[Lipschitz specialization]
\label{prop:lipschitz_specialization}
Under Assumption~\ref{assump:lipschitz}, for every document $i\in S_q$,
\begin{equation*}
    b_i(\mathcal G,q)
    \le
    L\epsilon+
    \frac{\Gamma_{k(i,q)}(q)}{|G_{k(i,q)}|}.
\end{equation*}
Consequently, on the event of Theorem~\ref{thm:cluster_shap_general},
\begin{equation}
\label{eq:lipschitz_full_bound}
\left|\tilde\phi_i(q)-\phi_i(q)\right|
\le
L\epsilon
+
\frac{\Gamma_{k(i,q)}(q)}{|G_{k(i,q)}|}
+
\frac{\Delta_{\mathcal M}(\eta)}{|G_{k(i,q)}|}.
\end{equation}
\end{prop}

\begin{proof}
Fix a query $q$ and suppress $q$ from the notation. For any two documents $a,b\in G_k$, the pairwise Shapley-difference identity gives
\[
\phi_a-\phi_b
=
\sum_{S\subseteq S_q\setminus\{a,b\}}
\omega(S)
\left[
\big(v(S\cup\{a\})-v(S)\big)-\big(v(S\cup\{b\})-v(S)\big)
\right],
\]
where $\omega(S)=\frac{|S|!\,(n-1-|S|)!}{n!}+\frac{(|S|+1)!\,(n-2-|S|)!}{n!}$, which satisfies $\omega(S)\ge 0$ and $\sum_{S\subseteq S_q\setminus\{a,b\}}\omega(S)=1$. Since the clustering step ensures $d(e_a,e_b)\le \epsilon$ for all $a,b\in G_k$, Assumption~\ref{assump:lipschitz} implies
\begin{equation}
\label{eq:within_cluster_shapley_dispersion}
|\phi_a-\phi_b|\le L\epsilon,\qquad \forall a,b\in G_k.
\end{equation}
Let $\bar\phi_k=|G_k|^{-1}\sum_{j\in G_k}\phi_j$. Equation~\eqref{eq:within_cluster_shapley_dispersion} implies
\begin{equation*}
|\bar\phi_k-\phi_i|\le L\epsilon,
\qquad \forall i\in G_k.
\end{equation*}
Now decompose the oracle bias:
\[
\begin{aligned}
 b_i(\mathcal G,q)
 &=
 \left|
 \frac{\phi^{\mathcal G}_{G_k}}{|G_k|}-\phi_i
 \right| \\
 &\le
 \left|
 \frac{\phi^{\mathcal G}_{G_k}}{|G_k|}
 -
 \frac{1}{|G_k|}\sum_{j\in G_k}\phi_j
 \right|
 +
 |\bar\phi_k-\phi_i| \\
 &\le
 \frac{\Gamma_k}{|G_k|}+L\epsilon.
\end{aligned}
\]
Combining this inequality with Theorem~\ref{thm:cluster_shap_general} proves Equation~\eqref{eq:lipschitz_full_bound}.
\end{proof}

The proof invokes Assumption~\ref{assump:lipschitz} only to bound the dispersion term $|\bar\phi_k-\phi_i|$. Without any assumption, this term is at most $\delta_k=\max_{a,b\in G_k}|\phi_a-\phi_b|$, the within-cluster dispersion of exact Shapley values, which gives the assumption-free bound $b_i(\mathcal G,q)\le \delta_k+\Gamma_k(q)/|G_k|$. In particular, under exact cluster-level computation, Cluster Shapley is exact for any cluster that is homogeneous ($\delta_k=0$) and aggregation-stable ($\Gamma_k=0$).

\subsection{Special Case: Exact Cluster-Level Computation}
\label{appssec:proof_exact_cluster}

The exact cluster-level implementation used in the main empirical analysis is the special case in which $\mathcal M$ computes exact Shapley values in the cluster-level game. Writing $\hat\phi_i(q)$ for the output of Algorithm~\ref{alg:cluster shap} in this case, we have $\delta_{\mathcal M}(q)=0$ and the output coincides with the oracle coarsened allocation $\hat\phi_i(q)=\phi^{\mathcal G}_{G_{k(i,q)}}(q)/|G_{k(i,q)}|$, so the approximation error equals the coarsening bias:
\begin{equation*}
\left|\hat\phi_i(q)-\phi_i(q)\right|
=
b_i(\mathcal G,q),
\qquad
\frac{1}{n}\|\hat{\boldsymbol\phi}(q)-\boldsymbol\phi(q)\|_1
=
B_1(\mathcal G,q).
\end{equation*}
Under Assumption~\ref{assump:lipschitz}, Proposition~\ref{prop:lipschitz_specialization} further yields
\begin{equation*}
\left|\hat\phi_i(q)-\phi_i(q)\right|
\le
L\epsilon+
\frac{\Gamma_{k(i,q)}(q)}{|G_{k(i,q)}|}.
\end{equation*}
Thus, with exact cluster-level computation, the realized empirical error in \S\ref{ssec:numerical_results} is exactly the deterministic clustering/coarsening bias of the chosen partition.

\subsection{Representative-Imputation Variant}
\label{appssec:representative_variant}

The main Cluster Shapley algorithm treats each cluster as a meta-document and then divides the cluster-level Shapley value equally among member documents. An alternative theoretical variant uses clustering only to decide which documents can share an imputed value, while estimating representatives' Shapley values in the original document-level game.

For each cluster $G_k$, choose a medoid representative
\begin{equation*}
    r_k
    =
    \arg\min_{r\in G_k}\max_{j\in G_k}d(e_r,e_j),
    \qquad
    \rho_k=\max_{j\in G_k}d(e_{r_k},e_j)\le \epsilon.
\end{equation*}
Then estimate the original-game Shapley value $\theta_k(q)=\phi_{r_k}(q)$ of the representative, for example by permutation sampling over the original documents in $S_q$, and impute $a_i=\hat\theta_{k(i,q)}$ to every document in the cluster. If exact efficiency is desired, one can apply the additive correction
\begin{equation}
\label{eq:representative_efficiency_correction}
    \tilde\phi_i^{\mathrm{rep}}
    =
    a_i+
    \frac{V_q-\sum_{j\in S_q}a_j}{n},
    \qquad
    V_q=v(q,A(q,S_q)).
\end{equation}
This correction preserves the total value even when the imputed representative values do not sum to $V_q$.

This variant removes the meta-document aggregation term because it estimates Shapley values in the original document-level game. If Assumption~\ref{assump:lipschitz} holds and the representative estimates satisfy $|\hat\theta_k-\phi_{r_k}|\le \Delta_{\mathrm{rep}}(\eta)$ simultaneously for all $k$, then before the efficiency correction,
\begin{equation*}
    |a_i-\phi_i|
    \le
    L\rho_{k(i,q)}+\Delta_{\mathrm{rep}}(\eta).
\end{equation*}
With the additive correction in Equation~\eqref{eq:representative_efficiency_correction},
\begin{equation*}
    |\tilde\phi_i^{\mathrm{rep}}-\phi_i|
    \le
    L\rho_{k(i,q)}+L\bar\rho+2\Delta_{\mathrm{rep}}(\eta),
    \qquad
    \bar\rho=\frac{1}{n}\sum_{j\in S_q}\rho_{k(j,q)}.
\end{equation*}
For Monte Carlo estimation with $N$ permutations and marginal-contribution range bounded by $V_{\max}$, $\Delta_{\mathrm{rep}}(\eta)=V_{\max}\sqrt{\log(2m/\eta)/(2N)}$ by the same union-bound argument used in Corollary~\ref{cor:error_mc}. Thus the representative-imputation variant has the cleaner theoretical form $O(\epsilon)+O(N^{-1/2})$ without a meta-document aggregation term.

This improvement is primarily in the bias decomposition rather than in the sampling order. The computation remains $O(n^2+Nm)$ for $N$ original-game permutation samples over $m$ representatives, and each marginal contribution is evaluated in the original document game rather than in the reduced cluster game. In our Amazon-review experiments, this variant did not improve the empirical accuracy--cost frontier over the Cluster Shapley implementation reported in \S\ref{ssec:numerical_results}. We therefore keep it as a useful theoretical and implementation variant, while retaining Algorithm~\ref{alg:cluster shap} as the main algorithm.

\subsection{Proof of Corollary~\ref{cor:error_mc}}
\label{appssec:proof_4}

\begin{proof}
For each cluster $G_k$, let $X_{t,k}$ denote its marginal contribution in the $t$-th random permutation of the $m$ clusters. The Monte Carlo estimator is
\[
\tilde\phi^{\mathcal G}_{G_k}=\frac{1}{N}\sum_{t=1}^N X_{t,k},
\]
with $\mathbb E[X_{t,k}]=\phi^{\mathcal G}_{G_k}$, because a uniformly random permutation realizes each coalition $T\subseteq\mathcal G\setminus\{G_k\}$ as the set preceding $G_k$ with probability equal to its Shapley weight. If each cluster-level marginal contribution has range at most $V_{\max}$, Hoeffding's inequality gives
\[
\Pr\left(\left|\tilde\phi^{\mathcal G}_{G_k}-\phi^{\mathcal G}_{G_k}\right|\ge \delta\right)
\le
2\exp\left(-\frac{2N\delta^2}{V_{\max}^2}\right).
\]
Applying a union bound over the $m$ clusters, with probability at least $1-\eta$ the deviation is at most
\[
\delta
=
V_{\max}\sqrt{\frac{\log(2m/\eta)}{2N}}
\]
for every cluster. Thus $\Delta_{\mathcal M}(\eta)=\delta$ in Theorem~\ref{thm:cluster_shap_general}, which gives Equation~\eqref{eq:error_mc_shapley}. The MAE bound follows by averaging the per-document bound~\eqref{eq:error_mc_shapley} over the $n$ documents, as in the mean-$\ell_1$ derivation in the proof of Theorem~\ref{thm:cluster_shap_general}. The stated sufficient condition for $N$ follows by solving Equation~\eqref{eq:error_mc_shapley} for a target document-level error. Finally, each Monte Carlo permutation evaluates $m$ cluster-level marginal contributions, so Step 2 costs $O(Nm)$ and the total complexity is $O(n^2+Nm)$.
\end{proof}

\subsection{Proof of Corollary~\ref{cor:revenue}}
\label{appssec:proof_revenue}

\begin{proof}
Fix a query $q$ and write $V_q=v(q,A(q,S_q))$. Let $\Phi=\sum_{j\in S_q}\phi_j(q)$ and $\tilde\Phi=\sum_{j\in S_q}\tilde\phi_j(q)$. By Shapley efficiency in the original document-level game, $\Phi=V_q$. If the cluster-level procedure is efficiency-preserving, then
\[
\sum_{k=1}^m \tilde\phi^{\mathcal G}_{G_k}(q)=V_q.
\]
Because Step 3 allocates each cluster value equally among its members,
\[
\tilde\Phi
=
\sum_{k=1}^m\sum_{i\in G_k}\frac{\tilde\phi^{\mathcal G}_{G_k}(q)}{|G_k|}
=
\sum_{k=1}^m\tilde\phi^{\mathcal G}_{G_k}(q)
=V_q=\Phi.
\]
Hence the raw exact and Cluster Shapley revenue allocations have the same denominator, and
\[
|\tilde r_i(q)-r_i(q)|
=
\beta r_q\left|\frac{\tilde\phi_i(q)}{\tilde\Phi}-\frac{\phi_i(q)}{\Phi}\right|
=
\frac{\beta r_q}{V_q}|\tilde\phi_i(q)-\phi_i(q)|.
\]
Applying Theorem~\ref{thm:cluster_shap_general} proves the query-level bound. For subscription revenue, the same efficiency argument applied query by query gives
\[
\sum_{j\in D}\mathbb E_q[\phi_j(q)]
=
\sum_{j\in D}\mathbb E_q[\tilde\phi_j(q)]
=
\mathbb E_q[V_q].
\]
Using Jensen's inequality and Theorem~\ref{thm:cluster_shap_general},
\[
\left|\mathbb E_q[\tilde\phi_i(q)]-\mathbb E_q[\phi_i(q)]\right|
\le
\mathbb E_q\big[|\tilde\phi_i(q)-\phi_i(q)|\big]
\le
\mathbb E_q\left[
    b_i(\mathcal G,q)
    +
    \frac{\Delta_{\mathcal M}(\eta)}{|G_{k(i,q)}|}
\right].
\]
Multiplying by $\beta R/\mathbb E_q[V_q]$ gives the subscription bound. When $\mathcal M$ is randomized, the bound inside the expectation is applied on the event of Theorem~\ref{thm:cluster_shap_general} for each query (for a finite set of queries, a union bound over queries delivers all per-query events simultaneously); for exact cluster-level computation, $\Delta_{\mathcal M}(\eta)=0$ and both bounds are deterministic.
\end{proof}

\section{Supplementary Implementation Details}
\label{appsec:details}

\subsection{Introduction to RAG}
\label{ssec:rag}
AI search engines are designed to provide real-time, contextually relevant responses to user queries. A key technique behind many of these systems is RAG, which integrates pre-trained LLMs with information retrieval to enhance response accuracy and relevance \citep{lewis2020retrieval}. RAG addresses the limitations of static, pre-trained LLMs by incorporating new information from up-to-date, domain-relevant documents (which can be potentially proprietary to the firm). By grounding responses in reliable documents, RAG improves the relevance of AI-generated answers, reduces hallucinations, and mitigates the issue of outdated information that plagues static models \citep{gao2023retrieval}. 

RAG models require two pieces of machinery -- (1) A generative model or LLM that was pre-trained on a large corpus of text, e.g., GPT, Llama, Claude, Gemini, Deepseek. These models can generate coherent general-purpose text, although they are often unable to incorporate proprietary documents and recent news, and (2) a set of documents, $D$, that can be used to provide additional information to the generative model. Depending on the use case, $D$ can take many forms. For example, if the goal is to generate a search engine for news aggregation, then $D$ would consist of a set of licensed news articles from news websites. Alternatively, if the goal is to generate a conversational search chatbot for aiding consumers in e-commerce websites, then $D$ would consist of the platform's own proprietary database, including product details, consumer reviews, etc.  The RAG architecture has three components:

\squishlist

\item \textbf{Retriever (R):} When a search query comes in, the retriever locates and retrieves relevant information by identifying a set of documents that are relevant to the search query. Essentially, given a query $q$ and a set of documents $D$, the retriever's goal is to identify a subset $S_q \subseteq D$ that is most relevant to the query.

\item \textbf{Augmentation (A):} In this phase, the retrieved documents ($S_q$) are integrated with the original input (user query, $q$) to provide additional context for the generative model. This augmentation ensures that the response from the generative model is grounded in retrieved reliable information, enhancing both its accuracy and relevance.

\item \textbf{Generator (G):} The generator, typically an LLM such as GPT or Claude, synthesizes the user's query ($q$) and the retrieved information ($S_q$) to produce a coherent response. By incorporating the retrieved documents, the LLM can generate outputs that go beyond its pre-trained knowledge, delivering more comprehensive and contextually appropriate responses.
\squishend

\begin{figure}[htbp]
    \centering
    \includegraphics[width=0.8\textwidth]{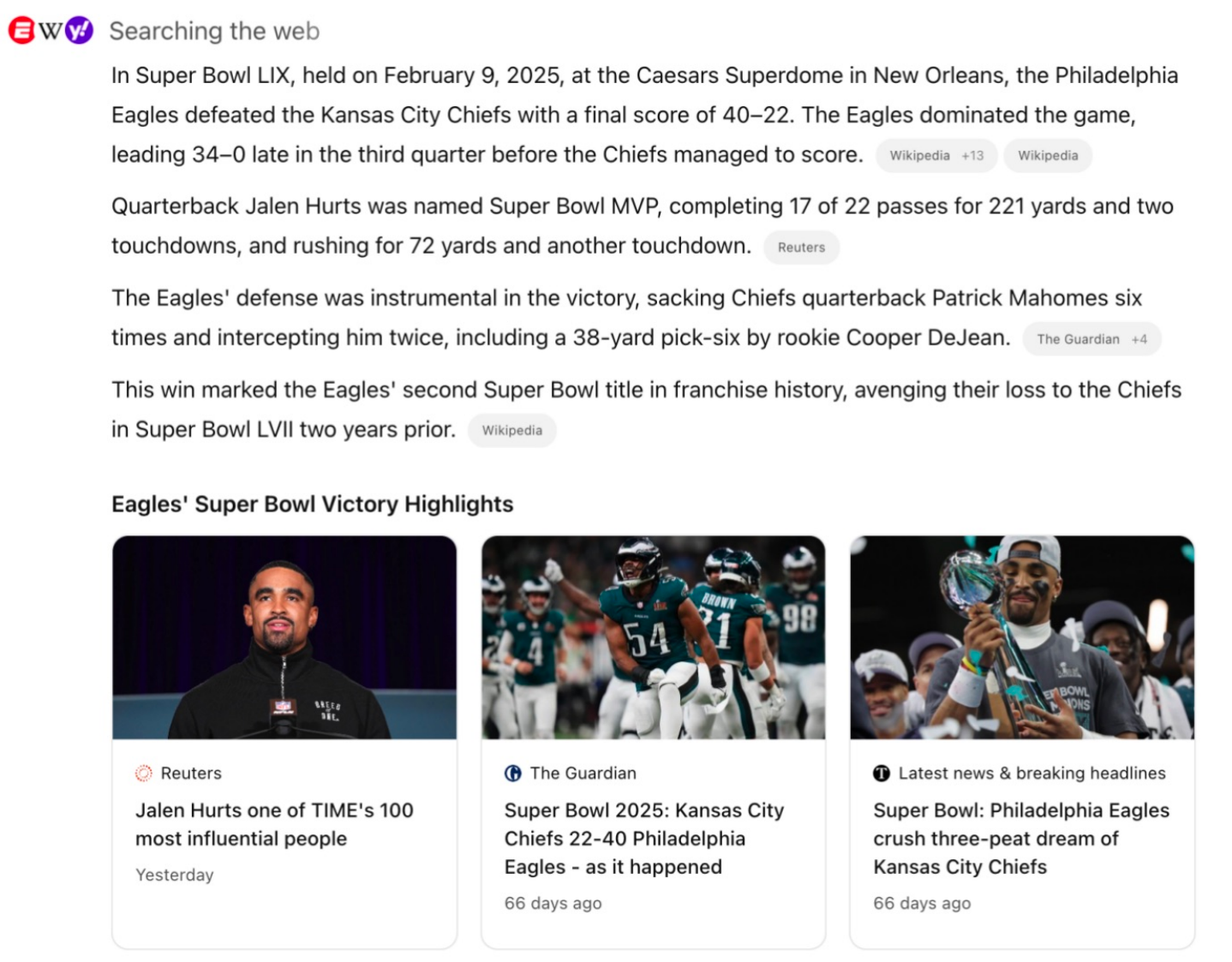}
    \caption{ChatGPT-4o with RAG-Enhanced Web Search}
    \label{fig:gptsearch_demo}
\end{figure}

To illustrate the impact of RAG integration on LLMs, we provide an example with a query: ``Who won the Super Bowl this year?'' in Figure~\ref{fig:gptsearch_demo}.  In this case, the ChatGPT application (powered by GPT-4o) leverages web browsing to retrieve up-to-date information that is not present in the model’s training data (GPT-4o has a knowledge cutoff of September 30, 2023). This example highlights how RAG enables LLMs to access and incorporate real-time information beyond their static training corpus. Note that the RAG system explicitly links to information sources in this example. This is standard in most RAG-based search systems since it allows users to verify response provenance and understand the basis of AI-generated content. Given the effectiveness and scalability of RAG architecture, it now forms the backbone of most of the modern AI-based search and summarization systems, e.g., Amazon's Alexa for Shopping (formerly Rufus) \citep{amazon_rufus}, Google AI Search \citep{google2024overview}, and OpenAI's ChatGPT Search \citep{searchgpt}. 

\subsection{RAG Pipeline for Amazon Review Summarization}
\label{appssec:rag_pipeline}

\begin{figure}[htbp]
    \centering
    \includegraphics[width=1.0\textwidth]{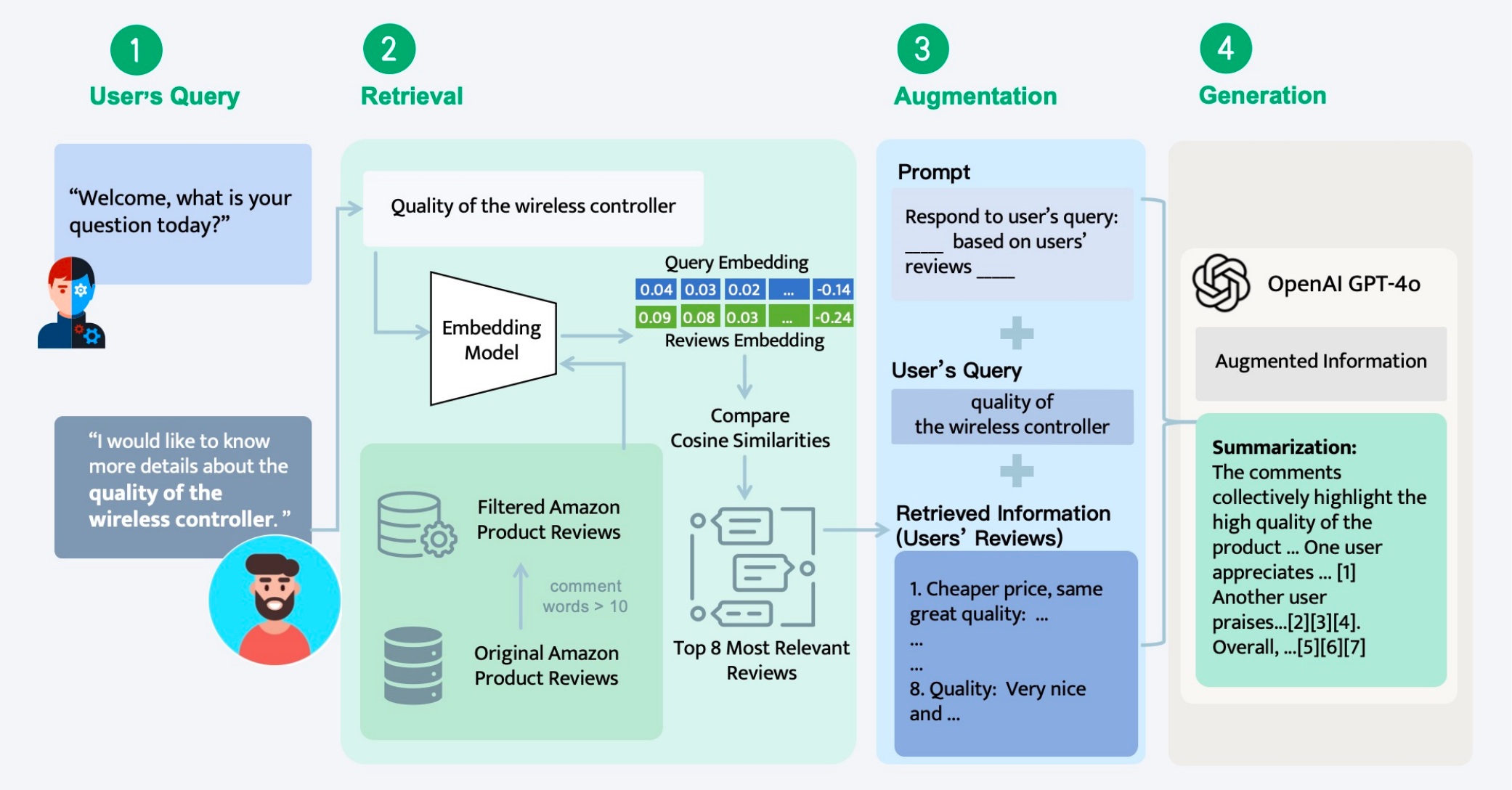}
    \caption{Architecture of our LLM-based search and summarization tool for Amazon Product Reviews. This flowchart illustrates the architecture of an AI-powered search engine designed for processing and summarizing reviews about the quality of \href{https://www.amazon.com/DualShock-Wireless-Controller-PlayStation-Black-4/dp/B01LWVX2RG/ref=sr_1_1?dib=eyJ2IjoiMSJ9.kDKBY50a1BZ2_pYrrwzgF0hgwJNSMhMHDPrv4xRi9ahL7L9SUE8Az33CRClA_p-He7DARMOsDSRCCC1gCPcJ8nIllpg2b9oID4bmQS4-A-dnpenDWJK-1x87nRF2_h1C5ijNpAinsQrCcJKdUA_ZZ2j2W0nLZExErf2Kltm5KrZo8ui3zec5_D4lR6qS-F5CA0TYCcjLJ0j784-rWcSAiiuR0jS3g1RKIELnjOgeRaY.Jn5N6TpEcVe6bowmgfRdrVponsStEgpd7wAqpMiK0Ww\&dib_tag=se\&keywords=DualShock\%2B4\%2BWireless\%2BController\&qid=1735453668\&sr=8-1\&th=1}{DualShock 4 Wireless Controller}. The process starts with the user query, where a specific question about the quality is posed. In the retrieval phase, the query’s key semantic information, ``the quality of the wireless controller'', is embedded and compared to filtered Amazon product reviews using cosine similarity. The system then retrieves the top eight most relevant reviews. During the augmentation phase, these retrieved reviews are combined with the original user query and our designed prompt, guiding the generation process. Finally, the generation phase employs OpenAI's GPT-4o model to summarize the augmented information, providing a concise response that cites the specific product reviews to ensure traceability and relevance to the user's query.}
    \label{fig:architecture_of_AI_engine_for_amz}
\end{figure}

We now describe our RAG-based search and summarization tool for Amazon reviews, which finds the relevant documents $S_q$ and produces the summary $A(q, S_q)$ for any given query $q$. Figure~\ref{fig:architecture_of_AI_engine_for_amz} shows the overview of the four-step procedure.

\squishlist
\item {\bf Step 0: Generate Text Embeddings} \\
The pre-processing step consists of generating text embeddings for all $D$ reviews/documents associated with a product generated using OpenAI's \texttt{text-embedding-3-large} model, which produces embeddings with a default size of 3072 dimensions. These embeddings are based on all the review text, including the title and the main content. In our analysis, we exclude reviews with fewer than 10 words, as they tend to be incomplete or uninformative. Our RAG architecture is agnostic to the exact source of embeddings, and it is possible to use alternative embedding models from open-source LLMs such as Llama, BERT, etc. However, recent research has shown that OpenAI embeddings tend to outperform the embeddings of such earlier models in discriminative tasks \citep{ye2025lola}; hence we use the OpenAI embeddings for our application.

%LLM-based text embeddings have been shown to be effective for downstream classification and are now commonly used in discriminative tasks such as sentiment analysis, scoring the text on how engaging the text is, spam detection \citep{devlin2018bert, liu2019roberta, ye2025lola}.

\item {\bf Step 1: Fetch user query $q$} \\ The process begins with a welcome message from the AI assistant to the user, followed by the user's search query related to some aspect of a product.

\item {\bf Step 2: Retrieval of relevant documents $S_q$} \\
We first process the user query to extract the key semantic information in it using an LLM (in our case \texttt{GPT-4o-2024-08-06}). The goal of this extraction is to identify the core meaning/consumer need expressed in the user's query. For example, in Figure~\ref{fig:architecture_of_AI_engine_for_amz}, the user's query is, ``I would like to know more details about the quality of the wireless controller.'' Here, the key semantic information is, ``quality of the wireless controller,'' which is extracted for further processing. 

Next, we use OpenAI's \texttt{text-embedding-3-large} model to generate the embedding for the processed query. We denote the embedding of the query as $e_q$. For each review $i$ in the set of reviews $D$, we represent its embedding as $e_i$. We then calculate the cosine similarity between the query embedding $e_q$ and the review embedding $e_i$ for each review. The cosine similarity between the document embedding $e_i$ and the query embedding $e_q$ is defined as 
$\frac{e_i \cdot e_q}{\|e_i\| \|e_q\|}$, where $e_i \cdot e_q$ is the dot product of the embeddings, and $\|e_i\|$ and $\|e_q\|$ are their respective Euclidean norms. For a detailed explanation of cosine similarity and its application in text similarity tasks, see Chapter 6 of \citep{schutze2008introduction}. The cosine similarity of a pair of embedding vectors captures the extent to which the two vectors are similar, with higher values indicating greater similarity. Thus, a higher cosine similarity indicates that a given review $i$ has greater relevance to the query $q$. Next, we rank the cosine similarity scores of the query $q$ for all the $D$ reviews and retain the most relevant reviews. 

We restrict ourselves to the eight most relevant reviews ($|S_q|=8$) from $D$ for all queries, though in practice, the retrieval process can be more complex.\footnote{We can either retrieve the top few documents or apply a cutoff based on the cosine similarity of the embedding values, excluding those below a fixed threshold, or combine both approaches by first selecting the top few relevant documents and then filtering those that meet the similarity threshold.} Our choice reflects current industry practice. For example, \cite{perplexity_citation_count} documented that Perplexity AI generated responses citing an average of 5.28 documents per query, and \cite{google_citation_count} mentioned that ``On average, (Google) AI overviews include 4-5 citations (the maximum is 33)'' from analyzing 1 million overviews. In this context, summarizing the top 8 relevant sources aligns well with how real-world LLM summary applications present content today. That said, our proposed Cluster Shapley algorithm is not limited to small-scale applications. It can be readily scaled to larger document sets when needed. %In $\S$\ref{ssec:robustness}, we present additional results for settings where the number of relevant documents is large.

%First, from a computational and monetary standpoint, including a large number of documents can be costly, as the LLM must process all associated tokens as context during generation. Second, including irrelevant or low-relevance documents may degrade summary quality, a concern supported by both anecdotal observations and prior literature on information overload \citep{jacoby1974brand, eppler2004concept}. Third, and most importantly, our choice reflects current industry practice. For example, \cite{perplexity_citation_count} documented that Perplexity AI generated responses citing an average of 5.28 documents per query, and \cite{google_citation_count} mentioned that ``On average, (Google) AI overviews include 4-5 citations (the maximum is 33)'' from analyzing 1 million overviews. In this context, summarizing the top 8 relevant sources aligns well with how real-world LLM summary applications present content today. That said, our proposed Cluster Shapley algorithm is not limited to small-scale applications. It can be readily scaled to larger document sets when needed. %In $\S$\ref{ssec:robustness}, we present additional results for settings where the number of relevant documents is large.

%\zikun{Reviewers asked about the scale; 8 is too small.}

\item {\bf Step 3: Augmentation by putting query $q$ and documents $S_q$ together} \\
In the augmentation phase, we combine the user query with the relevant product reviews using a prompt given to OpenAI's GPT-4o model (\texttt{GPT-4o-2024-08-06}). The detailed prompt is shown in Figure~\ref{fig:complete_summarization_task_prompt} in Web Appendix $\S$\ref{appssec:prompt_design}. The prompt instructs the model to analyze the filtered reviews, exclude irrelevant information, and then generate a summary focusing solely on content related to the query. 

\item {\bf Step 4: Generate the summary $A(q,S_q)$} \\
Given the user query ($q$) and the contextual information from the relevant reviews ($S_q$), the LLM (GPT-4o) produces a grounded summary by retrieving and synthesizing evidence from the provided documents. Although the LLM retains knowledge from its pretraining corpus, the response in this step is guided primarily by the contextual input delivered through the augmentation prompt (see Web Appendix $\S$\ref{appssec:prompt_design}). The output summary cites supporting reviews in square brackets (e.g., [2]), enabling users to trace each statement back to specific source documents. If a review is not relevant, it is explicitly marked as such by the model, e.g., (``[4] is not related to the query'').
\squishend

\subsection{Full List of Products and Designed Queries}
\label{appssec:query_list}

Table~\ref{tab:designed_queries_full} lists the 24 products used in our empirical analysis and, for each product, its number of reviews and the two designed queries described in \S\ref{sec:amazon_data}.

{
\renewcommand{\arraystretch}{0.85} % Reduce row height
\begin{footnotesize} % Or use \scriptsize for even smaller font
\begin{longtable}{|c|p{2.5cm}|c|p{8cm}|}
\hline
\textbf{No.} & \textbf{Product} & \textbf{\begin{tabular}[c]{@{}l@{}}Number of\\ Reviews\end{tabular}} & \textbf{Designed Queries} \\ \hline
\endfirsthead

\hline
\textbf{No.} & \textbf{Product} & \textbf{\begin{tabular}[c]{@{}l@{}}Number of\\ Reviews\end{tabular}} & \textbf{Designed Queries} \\ \hline
\endhead

\hline \multicolumn{4}{|r|}{\textit{Continued on next page}} \\ \hline
\endfoot

%\hline
\endlastfoot

1 & \href{https://www.amazon.com/DualShock-Wireless-Controller-PlayStation-Black-4/dp/B01LWVX2RG/ref=sr_1_1?dib=eyJ2IjoiMSJ9.kDKBY50a1BZ2_pYrrwzgF0hgwJNSMhMHDPrv4xRi9ahL7L9SUE8Az33CRClA_p-He7DARMOsDSRCCC1gCPcJ8nIllpg2b9oID4bmQS4-A-dnpenDWJK-1x87nRF2_h1C5ijNpAinsQrCcJKdUA_ZZ2j2W0nLZExErf2Kltm5KrZo8ui3zec5_D4lR6qS-F5CA0TYCcjLJ0j784-rWcSAiiuR0jS3g1RKIELnjOgeRaY.Jn5N6TpEcVe6bowmgfRdrVponsStEgpd7wAqpMiK0Ww&dib_tag=se&keywords=DualShock%2B4%2BWireless%2BController&qid=1735453668&sr=8-1&th=1}{Wireless Controller} & 15,594 & 1. Does the controller experience unresponsiveness? \newline 2. How would you rate the overall quality of the controller? \\ \hline

2 & \href{https://www.amazon.com/Curly-Co-Collapsible-Hair-Diffuser/dp/B019GBG0IE/ref=sr_1_5?crid=2K93RA7U2NUNF&dib=eyJ2IjoiMSJ9.-SexuJxtuZC91kG78umK3JKIIiihhqg6lFAUflq4JkbNiU5lW5KTWx4ylKknzZEkHWAZ_BEWnj-vGVT-zSxnGMH0x7151LEf0Tv0chyi8EPwrkNldCy2DS0k09zAcg6ABGrSErXP8GZZ-sLZ8cr5FBQRXBA7AhdVZLB5X_1FNYCHI-xATxJlOP0tp51GhZL7pPzYv0jFpmvkx_GoERX8F01Eu_wdAhk0otzJun3z5qZsebx8X6eOhockTimSsfRFPM0JjtlOHZ8WAxLZ1gGAAxBH3T8l7ZdcWCxtNhTX-Vr7jcj5b6pQGu9tGK8E4b_8hTFkVI-sMguoEd0sNQurhP61ZIXv-bMsu5t-PkYaZZfD0joXjxZQi1OMGmQRllt0d1apz_PXD3MKyOVRW2v2ZrIoaYfOeRW8edgyEg5E36bEyLIzaDmiJcb1tOPbTJ7c.VfEDEq4C9QEVEqA9UuDh9U_jnh1_3tqc4hCI4smT14Y&dib_tag=se&keywords=hair+diffuser+Maintain+your+natural+texture+while+you+blow+away+the+frizz%21&qid=1735457424&sprefix=hair+diffuser+maintain+your+natural+texture+while+you+blow+away+the+frizz+%2Caps%2C511&sr=8-5}{Hair Diffuser} & 1,328 & 1. Is the hair diffuser compact enough for travel? \newline 2. How would you describe the quality of the hair diffuser? \\ \hline

3 & \href{https://www.amazon.com/PlayStation-5-Console-CFI-1215A01X/dp/B0BCNKKZ91/ref=sr_1_2?crid=36AP7Y86BNIB&dib=eyJ2IjoiMSJ9.DN24o4RYS97v_6zyZZGG7SKtTT67jkn6NCUWDH30A14lZyU0cwl8UuIwFG190I84BiEcH2oBQImtVbh2KfsCoWkScXIX8CH3GqDCmCk5DQDVmL_vYv8dAuPKf90vulb-D-iiBtgAHwKCT9mcOzVLwIYo9IFGqiL2sEi2JRh6nVDgNX88t2Gtsi8OgnibvNYl-fNsyV_pagb38ytWOa3DWfYFSZO8zcJwQjZQCfPM8W8.27EPJABHrIVhWeN-wLpGVf9twiqxfnaiwfDDrVqowP8&dib_tag=se&keywords=The%2BPS5%2Bconsole%2Bunleashes%2Bnew%2Bgaming%2Bpossibilities%2Bthat%2Byou%2Bnever%2Banticipated.%2BExperience%2Blightning%2Bfast%2Bloading%2Bwith%2Ban%2Bultra-high%2Bspeed%2BSSD%2C%2Bdeeper%2Bimmersion%2Bwith%2Bsupport%2Bfor%2Bhaptic%2Bfeedback%2C%2Badaptive%2Btriggers%2C%2Band%2B3D%2BAudio%2C%2Band%2Ban%2Ball-new%2Bgeneration%2Bof%2Bincredible%2BPlayStation%2Bgames.&qid=1735457232&sprefix=ps5%2Caps%2C437&sr=8-2&th=1}{PlayStation 5} & 2,700 & 1. How’s the quality of the PlayStation? \newline
                     2. How’s the graphics of the PlayStation? \\ \hline

4 & \href{https://www.amazon.com/Turtle-Gaming-Headset-PlayStation-Nintendo-Console/dp/B00YXO5U40/ref=sr_1_1?dib=eyJ2IjoiMSJ9.j8VsOuL9QPbqbpBPsGpb3Q.-lb2Sot0fX60o2pjKNVxw0S3PEuMCaS5ggqnZPqVZtc&dib_tag=se&keywords=Take%2Bgaming%2Baudio%2Band%2Bcomfort%2Bto%2Bthe%2Bnext%2Blevel%2Bwith%2Bthe%2BTurtle%2BBeach%2BRecon%2B50X%2Bofficially%2Blicensed%2Bgaming%2Bheadset%2Bfor%2BXbox.%2BThe%2BRecon%2B50X%2Bfeatures%2Ba%2Blightweight%2Band%2Bcomfortable%2Bdesign%2C%2Bwith%2Bhigh-quality%2B40mm%2Bover-ear%2Bspeakers%2Bthat%2Blet%2Byou%2Bhear%2Bevery%2Bcrisp%2Bhigh%2Band%2Bthundering%2Blow.%2BFor%2Beven%2Bmore%2Bimmersive%2Baudio%2C%2Bthe%2BRecon%2B50X%2Bsupports%2Bspatial%2Bsound%2Btechnologies%2Blike%2BWindows%2BSonic%2C%2BDolby%2BAtmos%2Band%2BDTS%2BHeadphone%3A%2BX*.%2BQuickly%2Band%2Beasily%2Badjust%2Bmaster%2Bvolume%2Band%2Bmic%2Bmute%2Bwith%2Bconvenient%2Bin-line%2Bcontrols.%2BThe%2BRecon%2B50X%2Balso%2Bincludes%2BTurtle%2BBeach%E2%80%99s%2Brenowned%2Bhigh-sensitivity%2Bmic%2C%2Bwhich%2Bcan%2Bbe%2Bremoved%2Bwhen%2Bwatching%2Bmovies%2Band%2Blistening%2Bto%2Bmusic.%2BThe%2Bversatile%2Bmultiplatform%2Bconnection%2Bmakes%2Bit%2Bperfect%2Bfor%2Busing%2Bwith%2BXbox%2BSeries%2BX%2C%2BXbox%2BSeries%2BS%2B%26%2BXbox%2BOne%2C%2Bas%2Bwell%2Bas%2Bwith%2BPS5%2C%2BPS4%2B%26%2BPS4%2BPro%2C%2BNintendo%2BSwitch%2Band%2BPC%2B%26%2BMobile%2Bdevices%2Bwith%2B3.5mm%2Bconnection.&qid=1735457535&sr=8-1&th=1}{Headset} & 6,528 & 1. What’s the overall quality of the headset? \newline
               2. Is the headset comfortable? \\ \hline

5 & \href{https://www.amazon.com/PlayStation-Store-Gift-Card-Digital/dp/B07C438TMN/ref=sr_1_1?crid=11F2XPJ6N4LTW&dib=eyJ2IjoiMSJ9.pm47MMjOzw8JxJ7yDXqHrHdirUEHz8r0_sMoaos3mYSbn4dKhkijc3umtT45MMJPto3ijGyPRlDIqxmw12x0jeNZQtpFwcUbKGTl0h4rKnNbL5Ol7uiNcTHxkg17m1zdHjk8BRhj9u3yYcK1UxxsjvHqwPhvP71_l97loiQ0RNFqZrcUEtk1iW4puVWUhwsW0sQwc5-vV6zrQ9buj9ybf3V2dIVuK0-XdPZePMtRoks.BW-pESfbtbvDSEW7o18wwpG1f1OaT13peWycOo_Myso&dib_tag=se&keywords=gift%2Bcard%2Bplaystation%2BStore&qid=1735457718&sprefix=gift%2Bcard%2Bplaystation%2B%2Caps%2C669&sr=8-1&th=1}{Gift Card} & 4,827 & 1. Does the gift card not work well? \newline
                 2. How quick is the delivery of the gift card? \\ \hline

6 & \href{https://www.amazon.com/Head-TIGI-Hairspray-Extra-Strong/dp/B08X976H6W/ref=sr_1_1?dib=eyJ2IjoiMSJ9.3ziRrrAMWw5a-fcxOzfUWawHv5YteEY_1vL3IKVYV9NcU5HxqNatCD_v9oNNbcor9hxrIrorS0Q83Y3E1AwzptqufJ-EQmkHheojKa4KgrlGHHKqEdLeGLJbWIpkougo-n60PPsP-mAqL4l9vR55sry226ynyCbKgUnzF3Th3sgYRtwHXFjTbZxRfHJ5ujJpO3bt5PKF9Nrw7DLo6vOyrrkrT4lPbNGTamNfef4lpYQJmU37pLxPgHpDpd3Hjvyu8JZDKi_EnHEhJvYmt0ZNosw2YYlcVdnH8Y2W07VEETIdi58ZhRtTuSW458oud86MRN9qoTKnpuK6JGSw2c7ImVvFEHAeofUaBBofGZci-F2pV8r4qBhgrAv1kE1lc8uQqKq9bm6BG-GnULrtxwoo2LhyVMEOaiV3YprHUDp2QW72T17_8yUCIzARiCBWlb8J.FNP2MnVGP98RoHyDYJscRlw4Eu8NagOi75-envwiC3E&dib_tag=se&keywords=hair+styling+agent+TIGI+BH+HARD+HEAD+HAIR+SPRAY+Bed+Head+hair+care+products&qid=1735462411&rdc=1&sr=8-1}{Hair Styling Agent} & 959  & 1. How does the texture of the hair styling agent feel? \newline
                           2. What’s the quality of this hair styling agent? \\ \hline

7 & \href{https://www.amazon.com/Nonbongoy-Turban-Jersey-Headband-Headwraps/dp/B0974SYK4C/ref=sr_1_1_sspa?crid=29SQWJLDT546Y&dib=eyJ2IjoiMSJ9.lwifTeAmgx9Sd0n_wlLt-dQkU1PrreiYNLXPdGbtUoPqaIZcHb39N85G4WYuoehTWXVHyNqD6YMEvWjGzP0vr6vamdYK4Zo_O1g3vUx5Yfi-paiemG_u1Ed0DvM66wORMdBMjxgjkzB8lUhbfBTsrh0dvrWythCv6QBQV2PsCdPSU3iuabdpNoLZplO5oP3zRoYVTQiEX4z84qi_Dnx54YsofwjVScQ2d2wGaQHlrzmtRsiDvwgAjixf1zwIhYSMsZqSrBbZCkWWZ3rAzqy_9FbqiiP6PNbh054817VNol4.ZPSVjYp3AIrTrQisLjFlR6dgXtShg9EDSPIxLpTFJGg&dib_tag=se&keywords=4+pieces+head+wrap+scarves&qid=1735464572&sprefix=4+pieces+head+wrap+scarves+in+different+colors+including+wine+red%2Croyal+blue%2C+black%2Carmy+green%2Caps%2C379&sr=8-1-spons&sp_csd=d2lkZ2V0TmFtZT1zcF9hdGY&psc=1}{Headwrap} & 561 & 1. How stretchy is the headwrap? \newline
                2. Does the headwrap feel durable and high-quality? \\ \hline

8 & \href{https://www.amazon.com/Bed-Head-Curlipops-Tapered-Curling/dp/B008S6INZC/ref=sr_1_1?crid=38GWEUH3I6FNQ&dib=eyJ2IjoiMSJ9.PI5qQUv_bxeZJ0pEktO9q8UR3Q8ZyYIR4RcuuscoPJBlw37hTyttdsUEzu4xaP9xmsf-bQlrq80qf9GHTOwUMSwdHikB36PwCCENtZI59SRwL1ijx5JPpJlYAwYQDjVWgklEDmxkII4EdlOjBZg3JNSFqZ_n5ai-CndIKFsDFhuLvilXKrd3e7JLhbldgfEf95osm3BrCRBZjwLUsIFCcV2TiX209-OnfbbGRtmBBQr0M9OgdLDrYXRPjHzc7VvRxwTHCE2GdFEXJkDZT0XckAX0YHM6tk85xx5kZ238TzW9AUvuhsWwnl5RfadkW1Ls.39gT6ZWDv6Cpfn4LCzaET4cXpaO5MydULmmQqMXWiAI&dib_tag=se&keywords=Bed+Head+Curve+Check+Tourmaline+Ceramic+Styling+Iron&qid=1735464258&sprefix=hair+brush%2Caps%2C1018&sr=8-1}{Hair Curler} & 1,243 & 1. Is it easy and quick to use this hair curler? \newline
                   2. How’s the quality of this hair curler? \\ \hline

9 & \href{https://www.amazon.com/Brush-Bristle-Hairbrush-Smooth-Paddle/dp/B07N4PCYTD/ref=sr_1_1?crid=F5W23V0MW996&dib=eyJ2IjoiMSJ9.wf2JmIecgxlGarEiJIuW0eLl0uo6dfxmLJ8cTxuzcaBDNT4f27DsHa4avV5KFcYokID8E_gRXMJAJAbmNNWf7HBRwadGYisfNe_zN8ZXsnB3L1iQNDkEqrMP8mISGldNmK0wH7-ou48-SxjFrE4M2UrqhMda8KsgKuBG-wtKuLIsqpHcm8FWTVskGP3rBfqjxC-R48DOYCY2pXDsb3tP8kAzCBiKAFK5eJfu_G7Q5ThqzPy6KplSIPTEiwNsAwlywQfMMUktmWw_ai3CcP2fW5ZI8AXLzsG0hbxERGjyPfAPruRWyZuGWTfcwOniVrlgVNGzzVIHQJfpw9VIf9hUtESvR4nfzCUGN2-zwnbIDGoQmPmO1YVsqy9flYY8OXSSZYKmupdRIz3bdBYnffIfmOZb0086sOu1p7ZCroe9CAClrM32ZtZe-bGXXGkt4ZXo.on4sQRUTZvhCMvuYvWA-xbKc-cVoePRWds2QboPK2l4&dib_tag=se&keywords=hair+brush&qid=1735464115&sprefix=hair+brush+reduce+frizz+stimulates+scalp%2Caps%2C2099&sr=8-1}{Hair Brush} & 1,372 & 1. Is the hair brush soft and gentle on hair? \newline
                  2. How's the quality of the hair brush? \\ \hline

10 & \href{https://www.amazon.com/BS-MALL-Brushes-Synthetic-Foundation-Concealers/dp/B07MH1KHJ2/ref=sr_1_1?crid=11MRNLJEJ0K3I&dib=eyJ2IjoiMSJ9.5fa0wVmnph1PJXlrupBaFhajMm-nh_g2agm9x6H22G9qqKKgI6idcmNublnEs_1p6MdaSDAlJ3GJQqmw1J6NxqXv_SvCv0NN4-UXFYINg0w4POWifAK_mwPRbBcVX841fUAENfZ2Y42m50K7xqaZqXot3DXo9L4OQrjXrESNmbsap3ZvWbXk_toQvM5yAAtCEUQ0baFbYfub-30OigrmcqY8IXB5vLWU24NY4aM5psgz2LnC-585gBsh2OtUE5sQlrvwYchqi-5uYo80-z4VyMWyTi8lwssEMwiNxv1fgx9VzgtfZrWMCZ09QbgAAT5YIIdDVwNFyH0iZU1vpkXPpwUsvFtv-_sMkXir_Zh3RJb1mF7FxWFmIRZMYhsnGGP19NpWHiusByxASiNKIdj_2fgFD3RpVS-IuCHCrPU2MQ6qpsaQLZWcOGb2bQ2RdAzQ.jCDVQHD3Tad5XVi4j7nvwdjhTJvcQq1kFSEptT6Pomw&dib_tag=se&keywords=Premium%2BMakeup%2Bbrushes&qid=1735463322&sprefix=premium%2Bmakeup%2Bbrushes%2Caps%2C665&sr=8-1&th=1}{Makeup Brush} & 567  & 1. Does the makeup brush have a smell? \newline
                     2. How's the quality of the makeup brush? \\ \hline

11 & \href{https://www.amazon.com/SALUX-Nylon-Japanese-Beauty-Yellow/dp/B08DT52KKQ/ref=sr_1_5?crid=3CUY61POQXEXT&dib=eyJ2IjoiMSJ9.J6LiEsQ1t7pjPJRwEg3z9dSW7JF-FctxJ9W9fCz3uCywE8BjvznIemXoAdnwmpF4Kk4tCZqHmpmwho85NPkG5rANFJ52VAqbjVRyjy8VtlJW_jpfAhm7iljQvTHm0WkYTcLougrPfbzLeLUpzxvuFF9Ipyh_mWUEKNdkMsl5cZvBg4f310mgFiajETEMNo0eM36mrEZe09L1QrjrmqMp2eOX_xSkzjMpVKbPkKGlra9-27V_BkwNX1n1qZPU2DC4Tg699WZqT2bkue0H9QamQC-s5jrguepkMaqmuI1A8mRORde9xnO_OAoCfI461XwV9BmX6EyT0sdS_PCNCl--vv0JoA3fSv37W6Soqptr90Uxay3T41JJhLn2MBxA-pn0PG38p5EtlxtVWpAzNftxbUsqtBQ0SG4Gj6ElsuIPuieHLYDjseTFizfTOm6HiWR0.sYPjTDDODLTTZZmHTwF1TMwh7O2BOLw1z13RKBpGqmc&dib_tag=se&keywords=Salux+Nylon+Japanese+Beauty+Skin+Bath+Wash+Cloth%2Ftowel&qid=1735463862&sprefix=salux+nylon+japanese+beauty+skin+bath+wash+cloth%2Ftowel%2Caps%2C994&sr=8-5}{Bath Wash} & 1,962  & 1. Does the bath wash make skin softer? \newline
                  2. Is the quality of the bath wash up to par? \\ \hline

12 & \href{https://www.amazon.com/Massager-FReatech-Silicone-Bristles-Exfoliate/dp/B081F9VFF5/ref=sr_1_8?dib=eyJ2IjoiMSJ9.sZY0qATmHkYFqn9Gb3RZNbKffRXaFCgFszVtaPTipG11qnnbNo1ZgcaFgPx0rlJivn4FfhtRcYaapVY-DESX684kbmPSL2iue2alkA0VbX6WKuoq_Lr8QJ3gcKgWOUShLtdZWGgIzcGJluYzuQohN8YEwIAzMsQPZTSw5N3yJQvTphbCERcRVDJ26Y6xk24kwdfqI6mRUuUzZJgCuD2Rp83tClzIp0I6fFX4kLf9jLpSR8Q0jz_mbtAZEd6pAP0L7bgWq-X0dY8ZM_py2dKQzX5obOp1D-GnR-de0JNHRkrh0NCj2IvUWvHd6TXNKcz7AmfQmu7RXyJTeX2kHIDAHodUhoEiL2pgQHBTanfdXSYUKCR5NDzH9VKJHRVyXHMlRVg8dg3JRDRCOJGDSUzrnebz7pQJTgJYWgTgVIXjY6IMRD_vno0eofMtVn0jEJ3l.XSsnky_oMQpt2UCa7q1HUK5UBnfAjug7G6RqCiiL_dA&dib_tag=se&keywords=massager%2Bscalp%2Bsilicone&qid=1735463492&sr=8-8&th=1}{Scalp Massager} & 381 & 1. What size is this scalp massager? \newline
                       2. How’s the quality of this scalp massager? \\ \hline

13 & \href{https://www.amazon.com/Ubisoft-Rocksmith-Cable-Trilingual-playstation-2/dp/B00I0701P4/ref=sr_1_1?crid=31QD5O5ZHQ4OP&dib=eyJ2IjoiMSJ9.dmtzZsdmY9Q38P9P_wUdny63Gh1O-e7E19gpJSJIwJ7Mpz_-QyYl01XnlG2UiPbrT7p1mZbn0csKezXK4Mwhd_44UwO5Bhg3FSQvBNGXnEuDBi2nhiAw4EDEjv7pCGxM7fGjTMvSH481R2c97JEKzhZ7X08IEW2dNQHcYoxj3eMb81_fMZ0PS58u_yFTdQuc0EOOBKthFPu3AUa-xLyIMqeovMY1RfzMZ4KE72ahkMDoVzRT19qi-D6a_djcU41lbaxXIASZm9_yZ2GY_1I8O_mE68ygE12gLYZct2abjSIAVcNHBQXEFI0i_feTC2IRIR-KUonN5tKh18DJbuO9vlBxpjqTk8Jtp1o806s-YIXpCLt6A4d6gVTPiS4ZsJ4W.dK8QeWsRmdoPkmKnHPmHJ8vG3F4e5akG2n6Kq1L3Ask&dib_tag=se&keywords=Connect+any+electric+guitar+or+bass+via+1%2F4%22+output+directly+to+your+console+or+PC+and+get+real-time+note+detection+and+virtual+effects+in+Rocksmith%2B.&qid=1735459210&sprefix=connect+any+electric+guitar+or+bass+via+1%2F4+output+directly+to+your+console+or+pc+and+get+real-time+note+detection+and+virtual+effects+in+rocksmith%2B.+%2Caps%2C440&sr=8-1}{Audio Cable} & 1,511 & 1. How’s the quality of the audio cable? \newline
                   2. How’s the noise level of the audio cable? \\ \hline

14 & \href{https://www.amazon.com/Godefroy-Applications-Tint-Medium-Brown/dp/B019FWSEDI/ref=sr_1_2?dib=eyJ2IjoiMSJ9.zou0BUxEZ-xi8LBPsyz3PY9mS9RxPihJ3Zi9ypqFeE7UcLJXS1b11nsZmjqcaUJhnr1czzJgl5mthuHeZfPh76fHTVHOXeH8te0TC-HfBpq64tDVsfjGC5io-HJt4iYp1NCddf6TKrDPJhHNbuArPRU2PmiPejbNyXlBMMNIGXSOOfzG2oJP6v_hqgFVCwNV.SXfziRA-ca9bQ0dUHeReajgA9YpCR4r8-oGW-w09QEY&dib_tag=se&keywords=Godefroy+Professional+Hair+Color+Tint+Kit+is+ideal+for+coloring+small+areas+of+the+scalp+and+facial+hair.+Pre-measured+in+separate+capsule+units%2C+it+gives+you+the+versatility+to+use+the+exact+amount+of+color+with+no+waste.+Great+for+use+on+beards%2C+mustaches%2C+sideburns%2C+%26+temple%2C+anywhere+you+need+a+little+color.+Also+can+be+used+for+root+touch+ups+in+between+hair+color+treatments.+No+odor+%26+Blends+well+with+existing+color&qid=1735459580&sr=8-2}{Tint Kit} & 1,750  & 1. How good is the quality of the tint kit? \newline
                 2. Is the tint kit effective? \\ \hline

15 & \href{https://www.amazon.com/Super-Mario-3D-All-Stars-Nintendo-Switch/dp/B08G3MN6KP/ref=sr_1_1?crid=1YZ7ZWLA8TVH8&dib=eyJ2IjoiMSJ9.FptZD7PZo83B6hUxvXs6PBvlEXJ9LgQFL4FOyahZcklLAu-TIep0PhdBFE8zcp2SIsRn50eALgHkTPZARj0wOVLQjoGiCBhj9fONmWaPWkNOV_2tGlsV0rtI_j_AB3uEkPAqTvyZu0yUJmhhW3F_5rfOdsmzwILfzdqbiVVvn1cOmZFJ4ad5p9kRB5vL7lBYJvbvFfJeAvNxF2hB07AJWavFz0y1sy3Gc_wJpjR9wO0.eQooCj7CWqhf3PyTf5Lx8LsVCSOOPuiwNGIt8vkYaPw&dib_tag=se&keywords=Super_Mario_3D%2BNintendo&qid=1735459420&sprefix=super_mario_3d%2B%2Caps%2C966&sr=8-1&th=1}{Super Mario} & 1,221 & 1. How’s the quality of Super Mario? \newline
                    2. How's the multiplayer capability of Super Mario? \\ \hline

16 & \href{https://www.amazon.com/Dr-s-Remedy-Enriched-Polish-RESCUE/dp/B00BCMTXO4/ref=sr_1_3?crid=1SIHQ69Y4WU1O&dib=eyJ2IjoiMSJ9.DPwweu-MaCW2EDc2-hIuyG2Y3XEHX4qrtdX0MvD8mtRa4LA8Sz4w310l8j29KVxFX5XuIh6w0IxSC250D59_pf7O93v6p0Lid8007zCkg1SBh2fwzGR6o_LSpRj9E46QfHgiJS-oy8Lj1tw5bjN3VTRUzaTXpcb1r1A4Mq6nKuvqHfhtaajyXwj9RaciOJzHRphIWZQPwj7_p8UmWnQkhf-6kyrlMxzQenOexdjK2PmAHnuonCA6OS_DWsU0_frEkxT91UgDxRRufLK62tl5_5CQd1PZoiYv9Fk8hi5ex6vPVnIQt5eoPnil3vahvLGghs3LfYJtYF56lB7ZSvZH-hvew33N9UwyDX9mN-jkcFxO5mBF-d8rg_q5qVePRDC4dvJ_mrEBw1TIrF1GqvXj1BoHVvTmzNWwsNldFwDcE6Bi0swq0A6h9Xtzl5BEI94O.fKf9hEWiQ3xgDsLZB2RNRRooR2N-S2Jo9zLI009z-W0&dib_tag=se&keywords=Dr.%E2%80%99s+Remedy+nail+polish&qid=1735459920&sprefix=dr.+s+remedy+nail+polish+%2Caps%2C436&sr=8-3}{Nail Polish} & 534 & 1. How’s the quality of the nail polish? \newline
                    2. How's the durability of the nail polish? \\ \hline

17 & \href{https://www.amazon.com/NAIL-AID-Biotin-Ultimate-Strength-Clear/dp/B01DG10XA6/ref=sr_1_3?dib=eyJ2IjoiMSJ9.D0qQhwOtDts_wKvAlxHW-eJ6wK9KIQGKcohQkphxcMMxsZfRU0_XAcYtB63xl0xI_obFXg8IIfpv4QL8bha4RMB8UCEG5b8vhJ9_UjiufVxrnQRsNCWvEK2CbIgC9HOPGn96k32ncYQPWZO9z4uOIg4wzJGGq8mgRfLvpTlOJgPwUPfBEmxiNw848Xnu7Qc_u9mVlA-1h08IoRjMQWcAPeMzLtOHCz3HOWx56EBsLIPtT1BsoC2zYKmeXPILBvuyOWNXCPyNzOwWX97a838hQgBV5REjeZVC15fmKCReDrs5RvBFdbwJ5jjh8hKkL6cdeE6CpIQlpEdDVVctBv1Z2vxMvhtw5vZ0Uff2h9F7xqQRIxCpliknD_Vh2K1d6O74RigEhYuWvEol2VNll1EAIjvaLGsAZzYjmVG1EGjoy1nFht1p_UF1wo677lZOwOF5.549vlb7eFkvgE-87WNtDyR7e3-BbgQVgjU_hPA8QB2s&dib_tag=se&keywords=NAIL+STRENGTHENER%3A+A+nail+binding+treatment+rich+in+ceramide+that+provides+nails+with+strength+to+reinforce+the+thinnest+nail+layers+to+prevent+bending+and+breaking.&qid=1735460078&sr=8-3}{Nail Aid} & 323 & 1. How’s the quality of the nail aid? \newline
                 2. How effective is the nail aid? \\ \hline

18 & \href{https://www.amazon.com/Hairingrid-Mannequin-Hairdresser-Cosmetology-Training/dp/B07G87C9PP/ref=sr_1_1?crid=DBKN14ZBDY0K&dib=eyJ2IjoiMSJ9.HYsdm3k_l3WQTLo1IGTyVGRI31W1tFWttuOKknk3MyJFNDuc_v3mzXT-rOfKeqSJOoI6nuo5TTeRSDQ7Y9VysX_VwkDbEbA9uh7Qn2-VrBMmSP3q1FPIErTSkvMdZ8v2H_IjMd1ajHrOezq_Gqm0e1UMkSjUocECGaliv2RZVqtb9zdRE7lATtHVcoSGMzhc7sC5qiSYR0y7YD-l6Ihe3CaCqcoI7Ex8Xeb3aCcfKolInT02a-LD2WzZ3YJxXH4MjiF21FU9lK_3I9waFVB5KG23aojePElVRVRK4nbILKQiIChdVsatBPfO4te1_hJMhmx7EOwCC7UM2RPVUA0kWg7yqFeZQu-ZZGROBf83_iN2c9DWbh3Ipzvd9jZW5h9bzllCnhPI2Gz7H3RWpykH86zbi7fw-BgRJl3n1o216Egu6Jedb5J51MBlYtJ-ySCw.4zdL8x7WcuxXSekN2_7usLMUasege0ZRgaxT31fauO8&dib_tag=se&keywords=mannequin+Individual+Hair+Strand%3A14+inches&qid=1735460255&sprefix=mannequin+individual+hair+strand+14+inches%2Caps%2C457&sr=8-1}{Mannequin} & 881 & 1. How would you describe the quality of the mannequin? \newline
                  2. Is it a good mannequin for practicing braiding? \\ \hline

19 & \href{https://www.amazon.com/Headband-Birthday-Supplies-costume-Decoration/dp/B0DL55M4ZB/ref=sr_1_1?crid=8A4YIAU5P1VU&dib=eyJ2IjoiMSJ9.Vd97NL0FAsbTBoj_jadpbV3C7bHDrUjgh_0vHg2OoYl-2RVsXP220CBWAZeYceBugnGfxATzbZ0MLilkVmIa3NuPu_ZEk_-uzLYDM4-Lncj24dw1wCE5C282O7KujAXrCM2XT_FnAFyuqLOTplp-nahQ_KhL5CqYrv6sUFcSPxeKzkCRgpGWIEmpMTCv8XxZOtWn-UQiodixdc4F-ZticnfabYdubyOWAMgm5IWHM3ZTSHOxAPE-aaJr1Nu73-hyDNQUaSTOfWg98_Bbm_WhuWe23iG89nhtV3wczkltMcKMvU-UlMKM28vhetMge9-yqFjpM-Lu1PPn0q4QZPd2GseG1GKj7Vt55r6CabH_bml90nwjq4ovEtR5sIy_9u_pjVh-KqPM2T08GB9vS5mN82On4j5Vy72uu-saErhewd5WSseumVrU-rBnqcpo7AtM.4-NFK-GZDapKvhVW2i65CkTjW1dYtadSOiSYnpwMigg&dib_tag=se&keywords=Mouse+Ears+Headbands&qid=1735464725&sprefix=mouse+ears+headbands+in+2+style%2Caps%2C1070&sr=8-1}{Headbands} & 1,153 & 1. How good is the quality of the headbands? \newline
                  2. Are the headbands comfortable to wear? \\ \hline

20 & \href{https://www.amazon.com/Gauge-Stretching-Cream-Tapers-Expanders/dp/B00GGZVPFW/ref=sr_1_1?dib=eyJ2IjoiMSJ9.f8WbZiGbieVMSbYbouh1hE1isGA1CzTHgfqiQLnFhL3dIdFZV7jvMXuBdRH3XYgLRTcu0QtvAIccknNZ7VxoUnz6p7hr8KxB7zPfSCBaEXDugGswVUfBCxEvg5PM6v95VXYxn5cNGPYDm2K9zZvFKHx5T65fi_Q5QnTiIfzm12lNv2RfqHlz_Y75RNx5ymstHlvNWvntmicm51W5HSwuRA.v6Xp04FtDWFkgsuABh8EUMFbCAMIWKmKee52HQSwq2c&dib_tag=se&keywords=LUBRICANT+-+Gauge+Gear+lubricates+your+skin+while+you+stretch%21+Helps+Insertion+of+Tapers+and+Plugs+Double+Flared%2FOversized+Plugs.+Also+perfect+to+restart+a+closed+piercing+with+an+earring&qid=1735460616&sr=8-1}{Gauge Gear} & 785 & 1. What’s the quality of the gauge gear? \newline
                   2. Does the gauge gear help with healing? \\ \hline

21 & \href{https://www.amazon.com/Yes-Cucumbers-Soothing-Hypoallergenic-Facial/dp/B079NV9PBG/ref=sr_1_1?dib=eyJ2IjoiMSJ9.FztzjLL7brOuOUnWqOUf5A.sdrvJ6TMgi1KcHtS4tdThf-4-gJuKwykRISMsgTC9W8&dib_tag=se&keywords=Yes+To+Cucumbers+Hypoallergenic+Facial+Wipes+gently+remove+dirt%2C+sweat+and+makeup+%28eye+makeup+too%21%29+without+the+need+to+rinse.+Whether+you%27re+out+and+about%2C+going+for+a+run%2C+or+having+a+late+night%2C+these+wipes+make+it+a+snap+to+refresh+on-the-go.+Packed+with+green+superfoods+these+all-natural%2C+biodegradable+wipes+naturally+refresh+and+rejuvenate+all-in-one.+Stay+gorgeous%2C+and+go+from+zero+to+clean+in+seconds%21+Snap+up+the+lid+and+snatch+out+a+towelette%21+Gently+glide+across+your+face+and+neck.+Feel+refreshed%3F+You+should-+your+skin+is+clean%2C+revitalized+and+nourished-+in+just+a+snap%21+These+95%25+natural%2C+oil-free%2C+non-comedogenic%2C+hypoallergenic+wipes+are+made+with+compostable%2C+FSC+certified+fabric.+They+are+petroleum+free%2C+SLS+free%2C+paraben+free%2C+and+cruelty-free.+Say+Yes+to+Cucumbers+Hypoallergenic+Facial+Wipes%21+These+2+pack+wipes+are+specifically+sold+by+the+manfucaturer.+We+can%26rsquo.+INGREDIENTS+Water%2C+Glycerin%2C+Cucumis+Sativas+%28Cucumber%29+Extract%2C+Aloe+Barbadensis+%28Aloe%29+Extract%2C+Camellia+Sinensis+%28Green+Tea%29+Extract%2C+Sodium+Chloride+%28Sea+Salt%29%2C+Carpylyl%2FCapryl+Glucoside%2C+Guar+Hydroxypropyltrimonium+Chloride%2C+Polyglyceryl-4+Caprate%2C+Caprylic%2FCapric+Triglyceride%2C+Sodium+Benzoate%2C+Potassium+Sorbate%2C+Citric+Acid%2C+Phenoxyethanol%2C+Fragrance.&qid=1735460684&sr=8-1}{Facial Wipe} & 446 & 1. How’s the quality of the facial wipes? \newline
                  2. How effective is the facial wiped at cleaning?\\ \hline

22 & \href{https://www.amazon.com/Dental-Tools-Cleaning-Stainless-Remover/dp/B078R7ZX1W/ref=sr_1_1?crid=1BKL1KXBW6LNQ&dib=eyJ2IjoiMSJ9.9-Zzy2HRujv2UptDSitVHT5n2_R1p7YkaFBH_vMeCbNTkwLCxIxMFQnzTBo9_IJhBTWb0gV1ybOidGfrRBpYVdcdKk-clljUC_LU3NVSF0-XTgYAtLCm1qUUEI_tYcr7BcGSn1eTBOfzWvaKlJv1QuxPiUYlvhx0xnKwfh9cSIBbVHjDtjDPzXypE3D9FazK9cOtYirpj-nA8eL6F-YbAPTnFLlSPyBB5TcIMV3mjYGpJNnZFV9u0lqafbaWO0uYr-dsVrCB1FfZKMNioe_8STvIgCH0xUpUBpJH0c3FNTS0EdISIh4PSSLYc3nih_Bd7rfriIVIpffclVk5TTBwXwgy5qlblg82zWEfjHeIh0xiTPHaeWJG_4ynWf2sPknCuW2O-573jrcSnSZTqpuEQslM3vq4WFNSQ-MxEnI4NLnGOiJk5kP8Nx4ka1O6sS1b.1nymvyVGrratG0vUA1jjYncoEoZDQ_6kU_e_fFSEkjY&dib_tag=se&keywords=dental+tool&qid=1735463256&sprefix=dualshock%2B4%2Bwireless%2Bcontroller%2Caps%2C990&sr=8-1}{Dental Tool} & 1,343  & 1. How’s the quality of the dental tool? \newline
                    2. Does the dental tool do its job effectively? \\ \hline

23 & \href{https://www.amazon.com/Kamirola-Princess-Crystal-Headbands-Wedding/dp/B092Z1K9DP/ref=sr_1_4?crid=19UL6P6UAVATA&dib=eyJ2IjoiMSJ9.r20xm8VM5JHYvysFRXbbaokmWVnGhKw7CrypeP5OoXKLKvMWhFjf1skO4P3NGU0iSsWk09hZS2w_K_T1Nb_x4k-RAhuDOeZ4o4pnnDNK0gJLv9250Y7Eu9Y2mOvzu_VMP8i7aS_Ew7yHIOJM8QosGfa6k8v1woz8y-cZjC6yLzAVln-8NiFr50_tnyMofl4Lh2LZtZSfyNnuutwQJZiIxKYtR4kIWtaDNDdzrVUTm1n3NQ1WGSD46YT1F2ktKlD0ip8KudklogsKGMgy6Ouc6QZqwiNcCoAhTdJSlojfbhrB0qnJp7veDa0v08w91151puHZ0yvao0dywrC8grrPtZzJUaOphKyuNFhPujkxSXI1KpeXOcSF0o5G9kTovxisEbaHGDyVRsh7_fz0lXZFy34iQ72xYdpaWj5yXYdOSxhCFR-QNdB4g_zvBP8NyjZY.Y1On7l-Xo6oWWa6umjHhy470mNCzckNL_c71SiIeqh4&dib_tag=se&keywords=crystal%2Bcrown&qid=1735463056&sprefix=crystal%2Bcrown%2Band%2Ba%2Bexquisite%2Bgift%2Bbox%2Caps%2C873&sr=8-4&th=1}{Crystal Crowns} & 1,374 & 1. What’s the quality of the crystal crowns? \newline
                       2. How beautiful are the crystal crowns? \\ \hline

24 & \href{https://www.amazon.com/Niacinamide-Serum-Women-Refreshing-Facial/dp/B0CPPQB8RM/ref=sr_1_1?dib=eyJ2IjoiMSJ9.DWQpsyG4Gm-rcg361gRardjFuoYClE3KJmx3RGnCTBBShof5BWgFGBcQOkkBaEGVO3XRvAtfbFujluOEiA_0cBD2SkDIc2bROEmzuPsC8JY6hMXwr1Egxao4PjdpNnOJyS6jeBUbdGeljgQFqoJWA9pZ_LXAfHhuGaVi5gpjNUooj62WMN7wp2s8cvVZaA1BoN1z3E4hFO_C-lTvcblJ3977MnmJhrPVSVYfGvuLmP6c6FhTWXvYJ6DGvIOV9iupXgBe4wORIJeWWV7UUIVgt1KiFpYKMgjRNdbMdvyPm7cviJz8puxXl8Z_1_fuByPx_qCcPeNND2nY-q88LHBTuve5X-s5Yp-f9TD9Dftps2SkwJLCumipbjS0gMGGe23SBbgvXezfSiLxzdW90gWO0vbskJIul4XlnwipHvcXGHZzwXqitbBXJnmZmN9P_51F.Tk3HgbjsaW6WwyoaWRFXAp6Pjd0bQj-CeDHonV7IdR0&dib_tag=se&keywords=THE+ORDINARY+Niacinamide+10%25+%2B+Zinc+1%25+30ml%2C+1.01+Fl+Oz+%28Pack+of+1%29&qid=1735465788&sr=8-1}{Blemish Formula} & 449 & 1. What’s the quality of the blemish formula? \newline
                        2. How effective is the blemish formula? \\ \hline
\caption{Designed queries for selected products. We assume each product has an equal probability of appearing (i.e., uniform exposure), and each query within a product contributes equally (50\%) to the valuation.} \label{tab:designed_queries_full}
\end{longtable}
\end{footnotesize}
}

\subsection{Prompt Designs}
\label{appssec:prompt_design}

We now provide our designed GPT-4o prompts for both summarization ($A(q,S)$) and evaluation ($v(q,A)$) steps, outlined in Figure~\ref{fig:complete_summarization_task_prompt} and Figure~\ref{fig:complete_evaluation_task_prompt} respectively.

\begin{center}
\begin{minipage}{1\textwidth}
\begin{lstlisting}[basicstyle=\ttfamily\scriptsize\itshape, breaklines=true, breakatwhitespace=true, frame=single]
You are tasked with generating a high-quality summary based on user comments. Follow these steps to ensure that your summary is accurate, relevant, and well-structured.

1. Carefully Analyze the Comments:
   - Read through all the comments provided in the context.
   - Identify the key points that are related to the topic '{original_query}'.

2. Select Relevant Information:
   - Only include information in your summary that is relevant to the topic '{original_query}'.
   - For comments marked as "not relevant", simply state "[X] is not related to the query." Replace '[X]' with the corresponding comment number.

3. Construct a Coherent Summary:
   - Use an unbiased and journalistic tone in your summary.
   - Ensure that the summary is medium to long in length and that it covers the key points effectively.

4. Cite the Source of Information:
   - For each part of the summary, include a citation in the form '[NUMBER]', where 'NUMBER' corresponds to the comment's index.
   - Start numbering from '0' and continue sequentially, making sure not to skip any numbers.
   - The citation should be placed at the end of the sentence or clause that it supports.
   - If a sentence in your summary is derived from multiple comments, cite each relevant comment, e.g., '[0][1]'.

5. Final Review:
   - Double-check your citations to ensure they accurately correspond to the comments used.
   - Make sure that every sentence in the summary is cited and that irrelevant comments are correctly identified and excluded after the initial irrelevant statement.
   - Make sure every comment is cited. For example, if comment [0], [1], and [2] are all not related to the topic, then just summarize: '[0] is not related to the query. [1] is not related to the query. [2] is not related to the query.' If comment [0] is relevant, while [1], [2], and [3] are irrelevant, then summarize like this: provide a summary of [0], and then state '[1] is not related to the query. [2] is not related to the query. [3] is not related to the query.' Do not miss any comment even though they are irrelevant.
   - Ensure that your response is structured in JSON format with the following fields:
     - "key": A string that represents the indices of the comments used to generate this summary, e.g., "012" for comments 0, 1, and 2.
     - "summary": The final generated summary text, with citations included.

6. Key Reminders:
   - Do not include any irrelevant information in your summary. If a comment is not related to the topic, state it as described and move on.
   - Ensure that your summary is comprehensive, accurate, and clearly tied to the topic '{original_query}'.
\end{lstlisting}
\end{minipage}
\captionof{figure}{Prompt for GPT-4o to analyze the relevant reviews to the original query and generate a summary. The prompt specifies citation rules and explicitly requires noting if reviews are irrelevant. The \href{https://platform.openai.com/docs/guides/structured-outputs}{structured output} is generated and formatted in JSON, consisting of two fields: ``key'' for indexing the source reviews and ``summary'' for the final generated text, complete with appropriate citations.}
\label{fig:complete_summarization_task_prompt} 
\end{center}

\begin{center}
\begin{minipage}{1\textwidth}
\begin{lstlisting}[basicstyle=\ttfamily\scriptsize\itshape, breaklines=true, breakatwhitespace=true, frame=single]
You are an AI model trained to evaluate summaries. Below, you will find several summaries identified by their labels. Your task is to rate each summary on one metric. Please make sure you read and understand every single word of these instructions.

Evaluation Criteria:
Information Coverage MUST be an integer from 0 to 10 - How well the summary captures and clearly describes one or several key characteristics of the product. A high-quality summary should convey the important features, benefits, or drawbacks of the product as highlighted in the reviews. It should provide a rich and accurate depiction of key points.

Pay attention: The most important consideration is how effectively the summary communicates the product's key characteristics. The clearer and more richly it conveys these characteristics, the higher the score. If it fails to adequately describe the product's features, it should receive a low score.

Evaluation Steps:
1. Read all summaries provided and compare them carefully. Ensure the summary clearly and richly describes the key points relevant to the product without including irrelevant information.
2. Identify any important details or characteristics of the product that are missing from the summary.
3. Rate each of the summary based on how well it covers and conveys the important information from the reviews. The MORE comprehensively the summary covers the relevant information, the HIGHER the score it should receive. Pay attention: The primary focus should be on the topic {original_query}. If the summary deviates from the topic, it should receive a low score, regardless of the amount of information it contains.
4. If a summary contains only the sentence "[X] is not related to the query." where X is a number, then give it a score of 0. However, if the summary contains other content besides this sentence, just ignore it when scoring.

Your response should be in JSON format, with an array of objects. Each object should have two properties:
1. "key": The key of the summary (e.g., "0", "1", "01", etc.)
2. "score": The score for that summary (an integer from 0 to 10)
\end{lstlisting}
\end{minipage}
\captionof{figure}{Prompt for GPT-4o to evaluate the extent to which a summary accurately and comprehensively captures the key product attributes as requested in the query.}
\label{fig:complete_evaluation_task_prompt} 
\end{center}

\subsection{Temperature Selection in GPT-4o for Summarization and Evaluation Prompts} 
\label{appsec:temp_effect}

Due to the inherent stochastic nature of LLM outputs, we now investigate how the temperature setting in GPT-4o can affect the outcomes from summarization and evaluation prompts.

To gauge the temperature's impact, we conduct two numerical experiments examining the summarization and evaluation outcomes, respectively. For both experiments, we randomly sample five queries from the Amazon dataset and retrieve the five most relevant reviews per query. This filtering step helps mitigate overfitting, as our goal is to select an appropriate temperature setting and evaluate its effect using samples distinct from those used in the numerical experiment for testing algorithms. We test four temperature levels (0.0, 0.1, 0.5, and 1.0), and for each setting, we replicate the prompt 10 times to compute output variance. 

%Detailed results for the summarization and evaluation experiments are presented in Web Appendix~\ref{appssec:temp_summ} and Web Appendix~\ref{appssec:temp_eval}, respectively. For the summarization, we select a temperature of 0.1 to balance output consistency and expression richness. For the evaluation, we also choose a temperature of 0.1.

%\subsection{Temperature for Summarization Prompt}
%\label{appssec:temp_summ}

\begin{figure}[!t]
    \centering
    \begin{subfigure}[b]{0.48\textwidth}
        \centering
        \includegraphics[width=\textwidth]{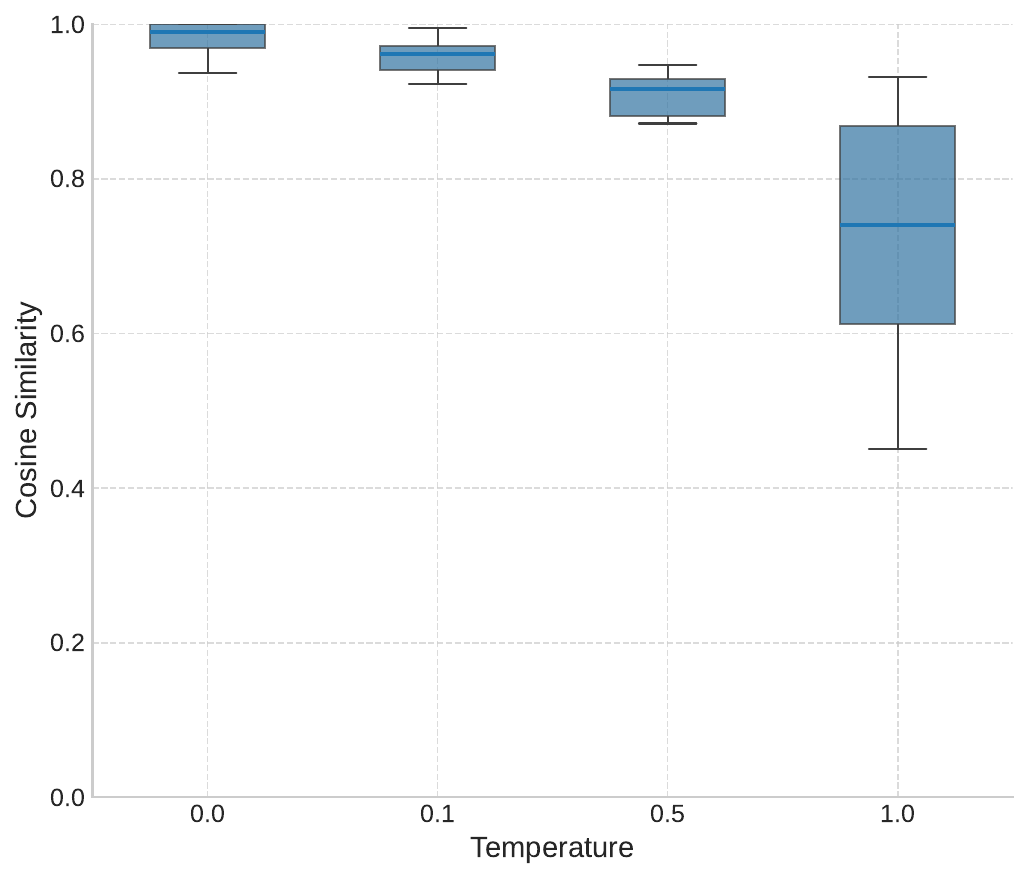}
        \caption{Effect of Temperature on Summarization Semantic Consistency}
        \label{fig:cosine}
    \end{subfigure}
    \hfill
    \begin{subfigure}[b]{0.48\textwidth}
        \centering
        \includegraphics[width=\textwidth]{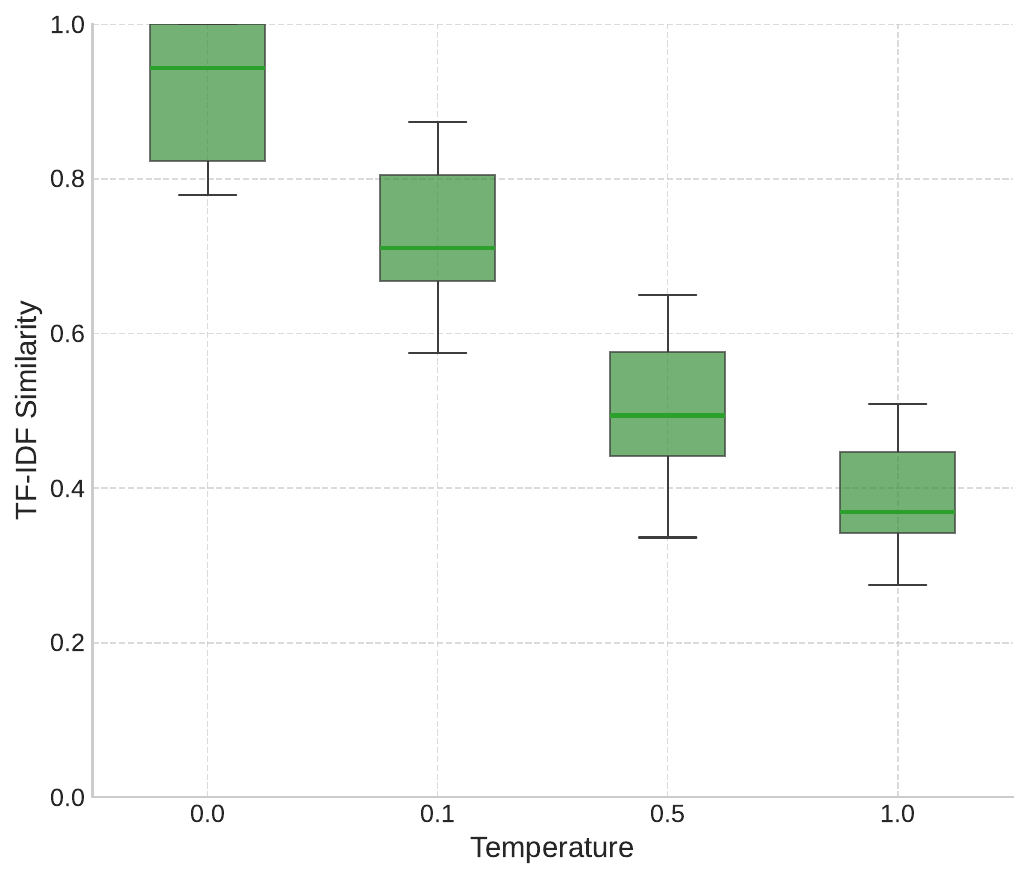}
        \caption{Effect of Temperature on Lexical Diversity}
        \label{fig:tfidf}
    \end{subfigure}
    \caption{Analysis of Temperature's Effect on Summary Generation. (a) Cosine similarity of embeddings measures semantic consistency, where higher values indicate stronger preservation of meaning across generated summaries. (b) TF-IDF similarity reflects lexical choice patterns, where lower values indicate more diverse vocabulary usage in the generated text, demonstrating the trade-off between consistency and diversity at different temperatures.}
    \label{fig:similarity_analysis}
\end{figure}

Because the outputs of summarization prompts are text information, to gauge the performance, we use two metrics: (1) We assess semantic consistency through embedding-based cosine similarity, which measures the degree to which different summaries preserve the same underlying meaning regardless of specific word choices; (2) We examine lexical diversity using TF-IDF (Term Frequency–Inverse Document Frequency) similarity measures, which quantify the variation in vocabulary and phrasing across multiple generations. While semantic consistency measures reliability in meaning preservation, lexical diversity reflects the model's creativity in expression. Figure~\ref{fig:similarity_analysis} visualizes the effect of temperature settings on both metrics.

\FloatBarrier 

As shown in Figure~\ref{fig:similarity_analysis}(a), even at temperature 0.0, where theoretically deterministic behavior is expected, the model exhibits slight variations in output (mean cosine similarity = 0.9820, std = 0.0217), confirming the presence of hardware-level computational variability. As temperature increases, we observe a decrease in semantic consistency, with mean similarities of 0.9576 (temp = 0.1), 0.8545 (temp = 0.5), and 0.7339 (temp = 1.0). While the decline in semantic consistency from temperature 0.0 to 0.1 is modest (approximately 2.5\%), this minor trade-off shows significant improvements in lexical diversity, as demonstrated in the TF-IDF analysis in Figure~\ref{fig:similarity_analysis}(b). Specifically, when transitioning from temperature 0.0 to 0.1, we observe a beneficial decrease in TF-IDF similarity from 0.9102 to 0.7244, indicating substantially more diverse vocabulary usage while maintaining semantic integrity. This optimal balance point at temperature 0.1 enables richer and more nuanced expression through varied word choices, while preserving the essential meaning of the content with high semantic consistency. However, at higher temperatures, both metrics indicate potential instability in the generation process. The substantial increase in semantic standard deviation (from 0.0217 at temp = 0.0 to 0.1696 at temp = 1.0) suggests increasingly unpredictable semantic variations, while the further decrease in TF-IDF similarity (0.5013 at temp = 0.5 and 0.3845 at temp = 1.0) indicates excessive vocabulary variation. These patterns at higher temperatures could potentially compromise both semantic reliability and textual coherence, reinforcing our choice of temperature 0.1 as the optimal setting for balancing semantic preservation with expressive diversity.

Choosing the temperature for evaluation is similar and straightforward, as the output is a numerical score ranging from 0 to 10. Since our ultimate goal is to obtain stable Shapley values, which are computed as weighted averages of these scores, we focus directly on how temperature affects the variance of the Shapley values.

\begin{figure}[htbp]
    \centering
    \includegraphics[width=0.8\textwidth]{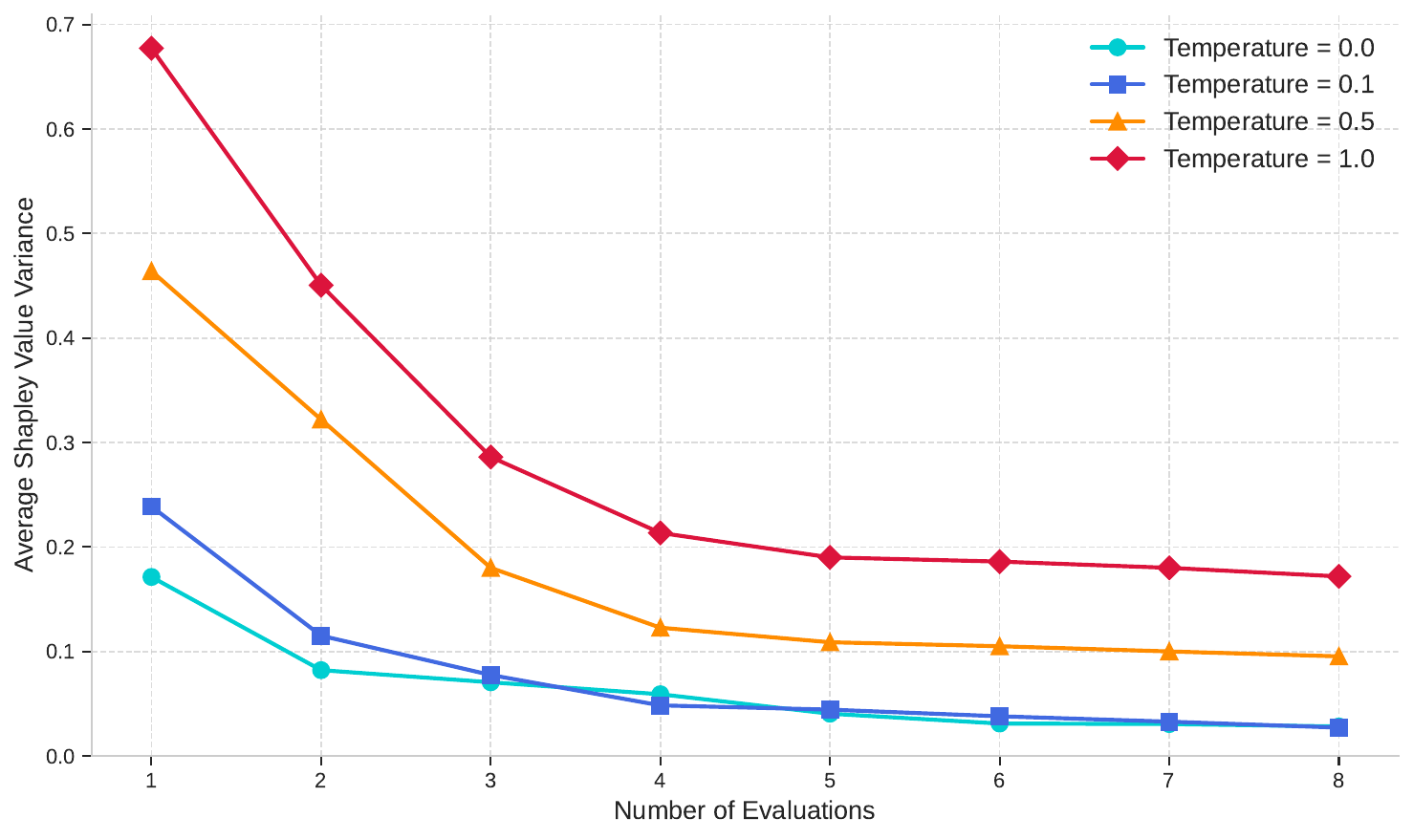}
    \caption{Shapley value variance under different temperatures and different replications of evaluation prompts. The $x$-axis represents the number of replications of evaluations. The $y$-axis is the variance of the average Shapley value over replications.}
    \label{fig:evaluation_temp_variance}
\end{figure}

Using the same setup as in our summarization analysis, we fix the generated summaries and repeat the evaluation process under different temperature settings. This isolates the variance caused solely by the evaluation prompt. As shown in Figure~\ref{fig:evaluation_temp_variance}, even at temperature 0.0, evaluation outputs exhibit variability due to hardware-level computational noise, similar to what we observed in summarization.

Among the tested settings, temperature 0.0 yields the lowest Shapley variance but still shows noticeable fluctuations. Temperature 0.1 displays slightly higher variance but achieves similar stability after averaging the evaluation score over multiple replications, making it a practical choice that balances consistency and alignment with our summarization process. In contrast, higher temperatures (0.5 and 1.0) result in substantially larger variances, suggesting diminished reliability.

We therefore adopt a temperature of 0.1 for evaluation prompts to ensure stable Shapley values without introducing excessive rigidity or randomness.

\subsection{Variance Analysis of Summarization, Evaluation, and Shapley Value}
\label{appssec:var_analysis}

Due to the stochastic nature of LLM prompts, we now analyze how this randomness can finally affect the variance of Shapley values, and propose a way to reduce the variance in this appendix.

The variance comes from both the summarization prompt \( A(S) \) and the evaluation prompt \( v(A(S)) \). We omit the notation $q$ in functions for better clarity, as $q$ is fixed and its omission does not introduce ambiguity. 

Formally, we can write down the random summarization and evaluation process as:
\begin{equation}\label{eqn:var_1}
    v(A(S)) = \mu(A(S)) + \varepsilon,
\end{equation}

where \( \varepsilon \) is white noise in the evaluation process, and \( \mu(A(S)) \) represents the expected performance score of the summarization \( A(S) \). Note that \( A(S) \) is a random object (a text-valued random variable), as the LLM may generate different summaries even under the same set of reviews \( S \) due to intrinsic randomness in GPT responses. This formula reflects that the randomness in the observed evaluation score \( v(A(S)) \) originates from two sources: the randomness in \( A \) and the evaluation noise.

We assume that the random summarizations \( \{A(S)\}_{S \subseteq D} \) are mutually independent. Given this assumption and independent white noise, and because the Shapley value (Equation~\eqref{eqn:shapley}) is a linear combination of the subset-level scores \( v \), its variance is governed by the variances and covariances of these scores; under our mutual-independence assumption the cross-subset covariances vanish, so it reduces to a weighted sum of the individual subset-score variances. Thus, in the following, we focus on analyzing the variance of \( v(A(S)) \):

\begin{equation}\nonumber
    \text{Var}(v(A(S))) = \mathbb{E}[v(A(S))^2] - \left(\mathbb{E}[v(A(S))]\right)^2.
\end{equation}

Substituting Equation \eqref{eqn:var_1} into the above variance expression yields:

\begin{equation}\nonumber
    \text{Var}(v(A(S))) = \mathbb{E}_{A,\varepsilon}\left[(\mu(A(S)) + \varepsilon)^2\right] - \left(\mathbb{E}_{A,\varepsilon}[\mu(A(S)) + \varepsilon]\right)^2.
\end{equation}

Expanding the squared terms and leveraging the linearity of expectation, we can get:

\begin{equation}\nonumber
    \text{Var}(v(A(S))) = \mathbb{E}_{A}[\mu(A(S))^2] + 2\mathbb{E}_{A,\varepsilon}[\mu(A(S))\varepsilon] + \mathbb{E}_{\varepsilon}[\varepsilon^2] - \left(\mathbb{E}_{A}[\mu(A(S))] + \mathbb{E}_{\varepsilon}[\varepsilon]\right)^2.
\end{equation}

Given that \( \varepsilon \) is independent of \( \mu(A(S)) \) and has zero mean, we have \( \mathbb{E}[\mu(A(S))\varepsilon] = \mathbb{E}[\mu(A(S))]\mathbb{E}[\varepsilon] = 0 \) and \( \mathbb{E}[\varepsilon] = 0 \). Therefore, the expression simplifies to:

\begin{equation}\nonumber
    \text{Var}(v(A(S))) = \mathbb{E}[\mu(A(S))^2] + \mathbb{E}[\varepsilon^2] - \left(\mathbb{E}[\mu(A(S))]\right)^2.
\end{equation}

Recognizing that \( \text{Var}(\mu(A(S))) = \mathbb{E}[\mu(A(S))^2] - \left(\mathbb{E}[\mu(A(S))]\right)^2 \) and \( \text{Var}(\varepsilon) = \mathbb{E}[\varepsilon^2] \), we can rewrite the variance of \( v(A(S)) \) as:

\begin{equation}\label{eqn:var_2}
    \text{Var}(v(A(S))) = \text{Var}(\mu(A(S))) + \text{Var}(\varepsilon).
\end{equation}

This result demonstrates that the total variance of the evaluation score \( v(A(S)) \) is the sum of the variance due to the summarization process \( \text{Var}(\mu(A(S))) \) and the variance due to the evaluation noise \( \text{Var}(\varepsilon) \). %Essentially, the observed \( v(A(S)) \) is noisy, and we can reduce this noise by averaging scores from multiple summarization and/or evaluation processes. 

\subsubsection{Empirical Variance in Summarization and Evaluation}
\label{appsssec:var_sum_eval}

To validate this variance decomposition and gauge the magnitudes of these variances, we conduct an experiment to quantify the variance contributions from both the summarization and evaluation stages. We select five distinct queries from Amazon product reviews, each containing five reviews, which is the same setting as in Web Appendix $\S$\ref{appsec:temp_effect}. For each query, we calculate the Shapley value to explore how the summarization and evaluation processes contribute to the overall variance. Specifically, for each subset of reviews in a query, we replicate both summarization and evaluation prompts three times to calculate the following empirical variance. 

\begin{itemize}
    \item \textbf{Total Variance} \( \text{Var}(v(A(S))) \) - This variance represents the overall variability of the evaluation scores for a given subset, considering all summarization and evaluation rounds. For each subset of reviews, we generate multiple summarizations and perform several evaluation rounds for each summarization. The total variance is calculated as the variance of all the evaluation scores across these rounds, capturing the combined effects of both summarization and evaluation.
        
    \item \textbf{Summarization Variance} \( \text{Var}(\mu(A(S))) \) - This variance reflects the variability introduced during the summarization process. After generating multiple summaries for each subset, we compute the mean evaluation score for each summary. The summarization process variance is then calculated as the variance of these mean evaluation scores across different summarizations. This captures how much the content of the summaries themselves contributes to the overall variability in evaluation scores, independent of the evaluation noise.

    \item \textbf{Evaluation Variance} \( \text{Var}(\varepsilon) \) - This variance isolates the variability introduced during the evaluation process. For each summarization, we evaluate the subset multiple times. The evaluation noise variance is computed as the average variance of the scores within each summarization round, reflecting the inconsistency of GPT-based evaluations across the same summary. In other words, it measures how much the scores fluctuate due to noise in the evaluation model rather than changes in the summaries themselves.

\end{itemize}

The results of this experiment, presented in Table \ref{tab:shapley_variance_comparison}, compare the total variance, evaluation noise variance, and summarization process variance for each subset. Consistent with the variance decomposition analysis, the total variance equals the sum of the variances from both the evaluation noise and the summarization process. On average, the summarization process variance accounts for approximately 53.08\% of the total variance, while evaluation noise contributes around 46.92\%. %It is important to note that while reducing both summary and evaluation variances could lead to more consistent evaluation scores, the current levels of variance are moderate and do not pose a significant threat to score stability. Therefore, improvements in consistency-whether by refining the summarization process or enhancing the evaluation protocol-would be beneficial but are not critical for maintaining the accuracy of the scores at this stage.

\begin{table}[ht!]
%\footnotesize
\scriptsize
\centering
\begin{tabular}{lccc}
\toprule
\textbf{Subset $S$} & \textbf{Total Variance} & \textbf{Evaluation Variance} & \textbf{Summarization Variance} \\
\midrule
\{1\}     & 0.3729  & 0.0741  & 0.2988  \\
\{2\}     & 0.2654  & 0.0741  & 0.1914  \\
\{3\}     & 0.1173  & 0.0741  & 0.0432  \\
\{4\}     & 0.1358  & 0.0371  & 0.0988  \\
\{5\}     & 0.3642  & 0.1667  & 0.1975  \\
\{1, 2\}  & 0.3519  & 0.0926  & 0.2593  \\
\{1, 3\}  & 0.3933  & 0.1111  & 0.2822  \\
\{1, 4\}  & 0.2037  & 0.0371  & 0.1667  \\
\{1, 5\}  & 0.1975  & 0.1482  & 0.0494  \\
\{2, 3\}  & 0.1543  & 0.0741  & 0.0802  \\
\{2, 4\}  & 0.1605  & 0.0741  & 0.0864  \\
\{2, 5\}  & 0.2099  & 0.1296  & 0.0802  \\
\{3, 4\}  & 0.1975  & 0.0926  & 0.1049  \\
\{3, 5\}  & 0.2778  & 0.1482  & 0.1296  \\
\{4, 5\}  & 0.2778  & 0.1482  & 0.1296  \\
\{1, 2, 3\} & 0.2469  & 0.1111  & 0.1358  \\
\{1, 2, 4\} & 0.2963  & 0.1111  & 0.1852  \\
\{1, 2, 5\} & 0.2099  & 0.1482  & 0.0617  \\
\{1, 3, 4\} & 0.1790  & 0.1111  & 0.0679  \\
\{1, 3, 5\} & 0.2654  & 0.1667  & 0.0988  \\
\{1, 4, 5\} & 0.3599  & 0.1852  & 0.1747  \\
\{2, 3, 4\} & 0.1481  & 0.0926  & 0.0556  \\
\{2, 3, 5\} & 0.2160  & 0.1667  & 0.0494  \\
\{2, 4, 5\} & 0.2346  & 0.2037  & 0.0309  \\
\{3, 4, 5\} & 0.2778  & 0.1914  & 0.0864  \\
\{1, 2, 3, 4\} & 0.2407  & 0.0556  & 0.1852  \\
\{1, 2, 3, 5\} & 0.2346  & 0.0741  & 0.1605  \\
\{1, 2, 4, 5\} & 0.2716  & 0.1482  & 0.1235  \\
\{1, 3, 4, 5\} & 0.2407  & 0.1296  & 0.1111  \\
\{2, 3, 4, 5\} & 0.3086  & 0.0741  & 0.2346  \\
\{1, 2, 3, 4, 5\} & 0.0432  & 0.0185  & 0.0247  \\
\midrule
\textbf{Average Variance} & 0.2457  & 0.1153  & 0.1304  \\
\bottomrule
\end{tabular}
\caption{Performance score variance for different subsets. There are a total of \(2^5-1=31\) different subsets.}
\label{tab:shapley_variance_comparison}
\end{table}

\subsubsection{Reduce the Variance of Shapley Value} 
\label{appsssec:reduce_var_shap}

The variance decomposition shown in Equation \eqref{eqn:var_2} provides a foundation for understanding how variance arises in our system, guiding our approach to measuring and managing variance when calculating the Shapley value. Specifically, we generate multiple instances of the summarization \( A(S) \) and take the average score across these summaries to reduce variance introduced by \( A \), and/or evaluate each summarization multiple times to obtain an averaged score across evaluations, thereby reducing variance introduced by \( \varepsilon \).

However, more summarization and evaluation replications mean higher computation cost, although with lower variance in Shapley. To determine the most cost-effective approach, we test variance under various configurations, including multiple evaluations and summarization. Specifically, we conduct the experiment using a single query, ``the delivery speed of the card,'' with four selected reviews. For each combination of summarization counts (ranging from 1 to 4) and evaluation counts (also from 1 to 4), we repeat the process six times to calculate the variance in Shapley value. We measure computational cost in terms of the total response time of the GPT-4o API required for each configuration.

\begin{figure}[htbp]
    \centering
    \includegraphics[width=0.8\textwidth]{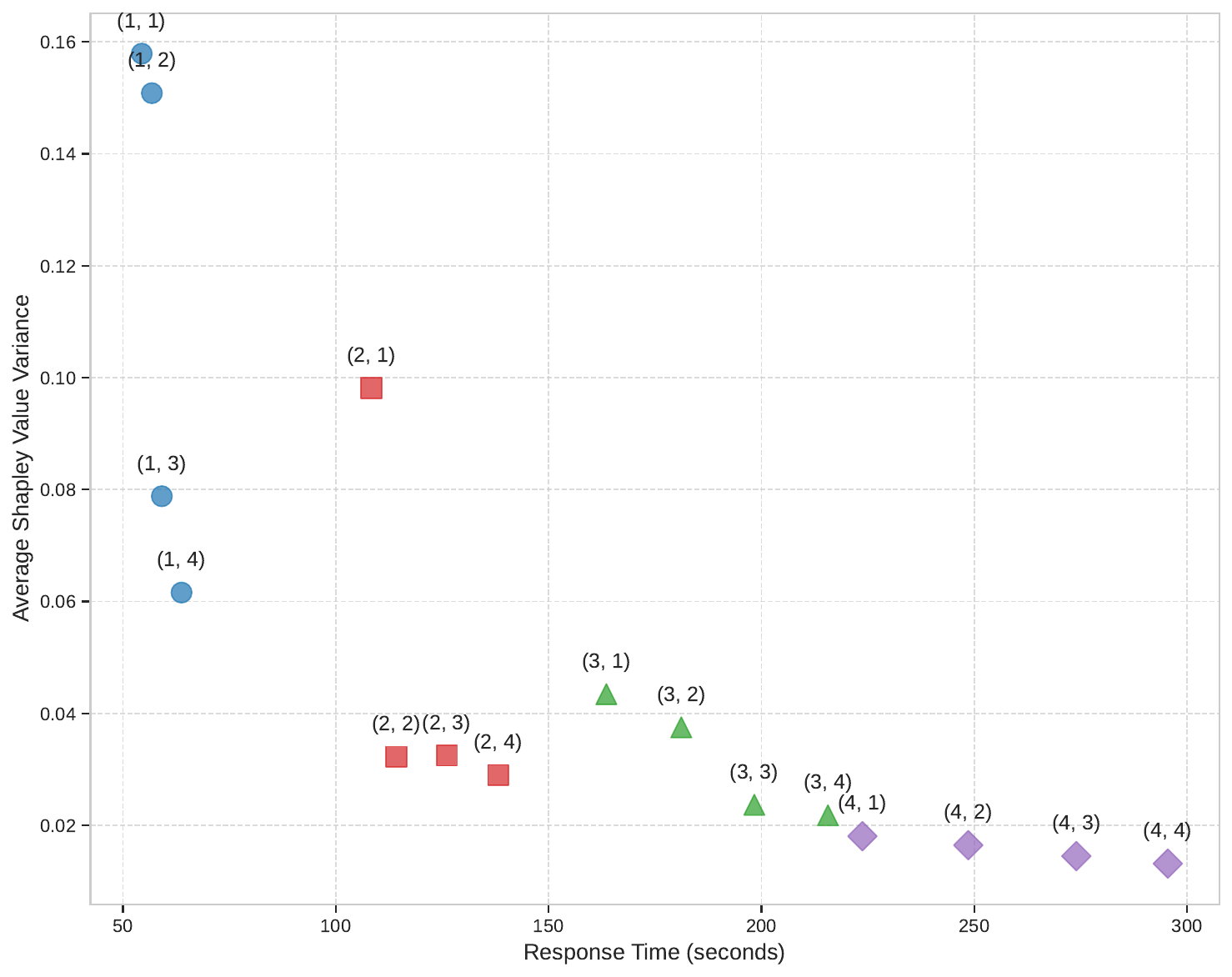}
    \caption{Cost-Effectiveness Analysis. The $x$-axis is the API response time (in seconds), and the $y$-axis is the average Shapley value variance. Each labeled point represents a configuration, with the first number indicating the number of summarizations and the second the number of evaluations. %Point size reflects the number of evaluation steps, while color and shape represent the summarization count.
    } 
    \label{fig:sum-eval-freq}
\end{figure}

The results are displayed in Figure~\ref{fig:sum-eval-freq}. We can observe that increasing both the summarization and evaluation counts reduces the average variance of the Shapley values. Notably, the most significant decrease in variance occurs when moving from lower to moderate counts of summarizations and evaluations. For example, increasing the evaluation count from 1 to 3 while keeping the summarization count at 1 reduces the average variance from approximately 0.1579 to 0.0788. Similarly, increasing the summarization count from 1 to 2 with an evaluation count of 2 decreases the average variance from approximately 0.1508 to 0.0322.

However, the rate of variance reduction diminishes with higher counts. Beyond certain thresholds, additional reductions in variance become marginal. For instance, increasing the evaluation count from 3 to 4 with a summarization count of 1 results in a variance reduction of only about 0.0172, from 0.0788 to 0.0616. These findings indicate a trend of diminishing returns, suggesting that conducting more than three evaluations or more than two summarizations provides limited benefits in terms of variance reduction.

Also, notice that one more summarization incurs more computation cost than one more evaluation because summarization outputs have more tokens than only one integer score out of the evaluation prompt. Considering practical applications where computational resources and time are constrained, a configuration with 1 summarization and 3 or 4 evaluations achieves substantial variance reduction while maintaining reasonable computation times.

To further examine whether evaluations should be replicated three or four times, we conduct an experiment using the test dataset consisting of five queries, each associated with five reviews, the same setup as in Web Appendices~\ref{appsec:temp_effect} and~\ref{appsssec:var_sum_eval}. We fix the summarization to a single run and vary the number of evaluation repetitions from 1 to 8. For each configuration, we compute Shapley values across 10 independent replications to estimate the variance.

\begin{figure}[htbp]
    \centering
    \includegraphics[width=0.7\textwidth]{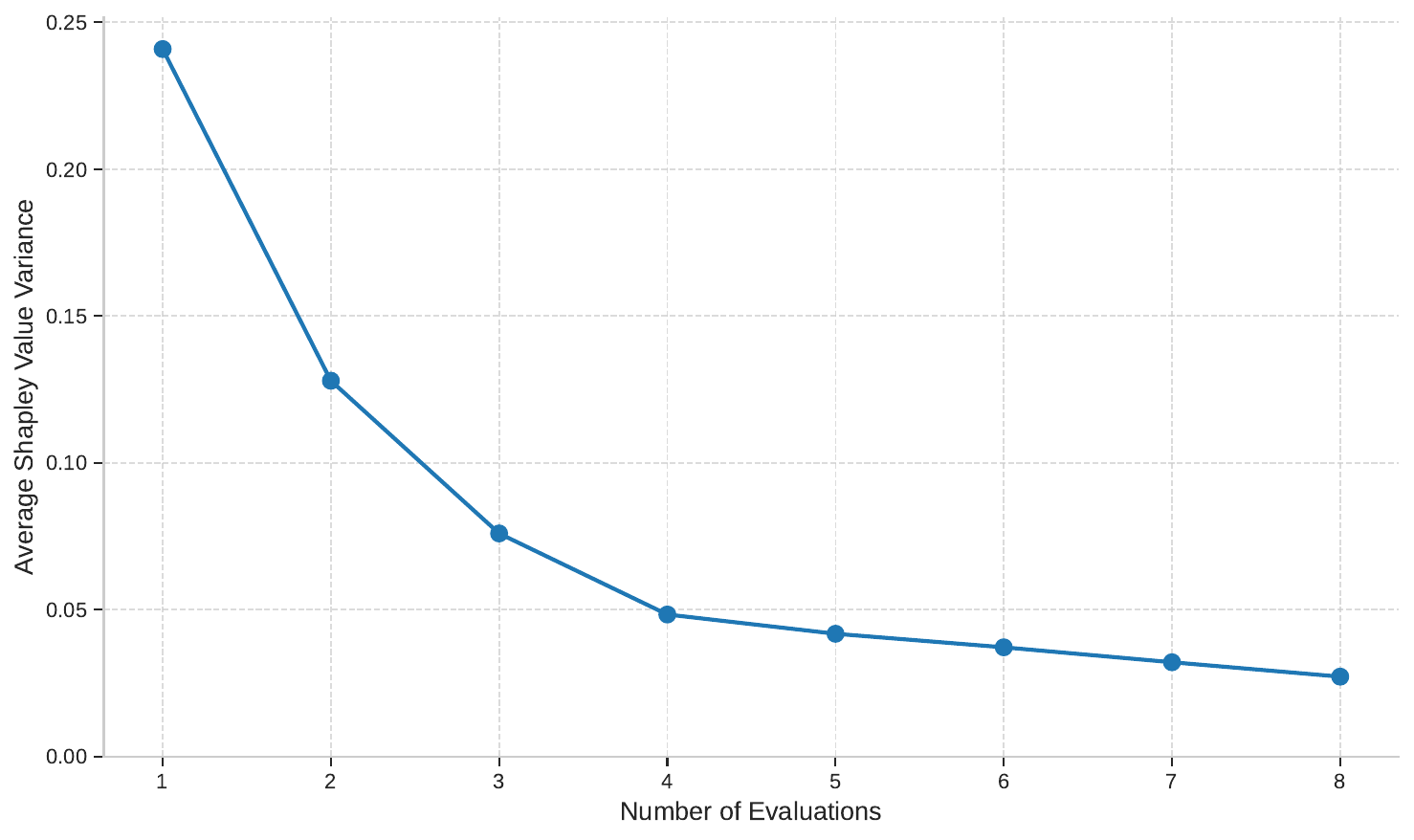}
    \caption{Reduction in Shapley Value Variance with Increased Evaluations}
    \label{fig:evaluation-freq}
\end{figure}

As shown in Figure~\ref{fig:evaluation-freq}, the variance in Shapley values decreases as the number of evaluation repetitions increases, but the rate of reduction diminishes over time. The most notable improvement occurs between one and four evaluations, where the average variance drops from approximately 0.25 to 0.05. Beyond four repetitions, additional variance reduction becomes marginal. These findings suggest that averaging over four evaluations provides an effective trade-off between computational cost and variance reduction. Accordingly, we adopt this configuration in our main numerical experiments.

\subsection{Computation time of the Clustering Step}
\label{appssec:cluster_epsilon_iteration_time}

We now analyze our adaptive clustering (i.e., Algorithm \ref{alg:dbscan}) using the same experimental setup as our benchmark comparisons: 48 test queries, each evaluated under 41 different $\epsilon$ settings ranging from 0 to 1. The results demonstrate efficient convergence behavior, with 71.19\% of cases converging immediately without requiring iterations. When iterations are needed, the process requires an average of 1.8 iterations, with a maximum of 19 iterations observed in extreme scenarios.

Each iteration of the adaptive clustering involves only pairwise distance comparisons over the retrieved documents, so its cost is orders of magnitude smaller than the LLM operations in Step 2 of our Cluster Shapley algorithm ($\S$\ref{ssec:cluster_shapley}), which require approximately 3.5 seconds per subset evaluation. The computational cost of the adaptive clustering step is therefore negligible in the overall Shapley value calculation pipeline.

\section{Robustness Checks and Extensions}

We now present detailed results and discussion for several aspects of the robustness checks and extensions described in $\S$\ref{ssec:robustness}.

\subsection{Robustness Check using Claude for Evaluation}  
\label{appssec:claude_evaluation}  

In the main study, we use the same GPT-4o for both summarization and for evaluating the summaries, which may introduce biases because LLMs tend to favor their own summaries. To address this, we conduct a robustness check here using \texttt{Claude-3.5-Sonnet} as an alternative evaluation model to compare against the results obtained from \texttt{GPT-4o-2024-08-06}, aiming to examine whether the evaluation outcomes remain consistent using different LLMs. 

To ensure a fair comparison, we use the same prompt (as shown in Figure~\ref{fig:complete_evaluation_task_prompt}) and the same temperature setting of $0.1$ for evaluation in Claude-3.5. For each summary generated by GPT-4o, we perform four independent evaluations using both GPT-4o and Claude-3.5, and take the average over four replications as the final evaluation score, to mitigate the variance caused by the inherent stochasticity of LLM outputs. To quantify the alignment between the two LLM models' evaluations, we conduct the Pearson correlation test on the relative rankings of summaries assigned by the two models, as this ranking metric captures whether the models agree on the comparative quality of summaries even if their absolute scores differ.  We also do the correlation test on Shapley values to see the impact of using different LLM models for evaluation on the final Shapley values.

The correlation between evaluation rankings obtained from two different LLM evaluation models is 0.788, and the correlation between the resulting Shapley values is 0.915. Both results are statistically significant at the 0.05 level, suggesting that our evaluation framework yields consistent outcomes across different evaluation approaches, and the validity of using the same GPT-4o for both summarization and evaluation.

%For the MAD, we calculate both the evaluation scores of subsets and the Shapley values of documents to understand the differences at both evaluation and attribution levels caused by different evaluation LLM models.  The results show mean MAD is 0.812 for subset scores, suggesting the differences between Claude-3.5-Sonnet and GPT-4o. These differences are  attributable to the intrinsic randomness in LLM outputs, rather than systematic biases in evaluation. Despite these variations, the correlation coefficient for subset scores demonstrated strong alignment, with a mean of 0.788 and a median of 0.837. For Shapley values, the observed differences were notably smaller, with a mean MAD of 0.179 and a median MAD of 0.168. The Pearson correlation coefficient for Shapley values was substantially higher, with a mean of 0.915 and a median of 0.936, indicating a very strong consistency between the two models in evaluating contributions.  These results underscore the robustness of our evaluation framework. Although small differences in absolute scores were observed, these differences are within the range expected from the stochastic nature of LLMs and do not compromise the relative rankings and the computed Shapley values. The high correlation coefficients across both subset scores and Shapley values suggest that our evaluation framework is generalizable and reliable across different state-of-the-art LLMs, such as GPT-4o and Claude. This consistency strengthens the validity of the proposed methodology and ensures its applicability to a broader range of evaluation settings.  

\subsection{Robustness to Retrieval Depth}
\label{appssec:vary_k}

Document valuations are computed conditional on the retrieved set $S_q$, which is determined by the retrieval system and, in particular, by the top-$K$ retrieval depth (the main analysis fixes $K=8$). Because changing $K$ alters the coalition structure (and because a different retrieval system would likewise yield a different $S_q$), we verify that our valuations are not an artifact of a particular retrieval depth. This check therefore speaks both to the sensitivity of the results to the top-$K$ choice and, more broadly, to their dependence on the retrieval step that selects $S_q$.

Top-$K$ retrieval returns the $K$ documents most similar to the query, so the retrieved sets are nested ($S_q^{(4)} \subseteq S_q^{(6)} \subseteq S_q^{(8)}$). We recompute exact Shapley values at $K \in \{4, 6, 8\}$ for all 48 query--product pairs. For each pair we compare the exact Shapley values of the documents shared across depths and report their rank correlation in Table~\ref{tab:vary_k}. The valuations are highly stable: the median Spearman and Kendall correlations equal $1.00$ for every comparison, so for most products the document ranking is \emph{identical} regardless of $K$, and the mean correlations remain at or above $0.91$. As expected, the largest changes arise when moving from the shallow $K=4$ to $K=8$, whereas adjacent depths agree almost perfectly. We conclude that the relative valuation of documents is robust to the retrieval depth and, more generally, to moderate changes in the retrieved set.

\begin{table}[htbp]
\centering
\caption{Robustness to retrieval depth $K$. Rank correlation, across the 48 query--product pairs, between the exact Shapley values of the shared documents computed at different top-$K$ retrieval depths. Higher indicates greater stability; a value of $1$ denotes an identical ranking.}
\label{tab:vary_k}
\begin{tabular}{lcc}
\toprule
Comparison & Spearman (mean / median) & Kendall (mean / median) \\
\midrule
$K=4$ vs.\ $K=8$ & $0.93$ / $1.00$ & $0.91$ / $1.00$ \\
$K=6$ vs.\ $K=8$ & $0.98$ / $1.00$ & $0.97$ / $1.00$ \\
\bottomrule
\end{tabular}
\end{table}

\subsection{Robustness to the Summarizer Model}
\label{appssec:vary_summarizer}

The main analysis uses GPT-4o as the summarization model. To verify that the document valuations are not an artifact of this choice, we re-run the valuation pipeline using Gemini~2.5~Flash-Lite, a substantially smaller and cheaper model, as the summarizer, holding the evaluator (Gemini~2.5~Flash, which scores the summaries from both summarizers), retrieval procedure, and evaluation prompt fixed. For each of the 48 query--product pairs we recompute the document Shapley values from the Flash-Lite summaries and compare them with the GPT-4o-based values, averaging four evaluation passes as in $\S$\ref{appssec:claude_evaluation}. The document Shapley values are broadly preserved in magnitude: the mean absolute difference between the Shapley values under the two summarizers has a median of $0.28$ across the 48 pairs (mean $0.42$; Table~\ref{tab:vary_summarizer}). A different summarizer produces different summaries, and hence a different realized value function $v(\cdot)$, so the values are not identical; this typical disagreement is nonetheless small relative to the spread of Shapley values across documents within a query (whose median within-query range is about $1.4$). This is consistent with the summarization-agnostic design of the framework: the valuation procedure is unchanged, and only the input summaries it values differ.

\begin{table}[ht]
\centering
\caption{Robustness to the summarizer model. Distribution, across the 48 query--product pairs, of the mean absolute difference between the document Shapley values computed from GPT-4o summaries and from Gemini~2.5~Flash-Lite summaries (common Gemini~2.5~Flash evaluator, four evaluation passes).}
\label{tab:vary_summarizer}
\begin{tabular}{cccc}
\toprule
median & mean & 25th pct. & 75th pct. \\
\midrule
$0.28$ & $0.42$ & $0.18$ & $0.50$ \\
\bottomrule
\end{tabular}
\end{table}

\subsection{Performance under MAPE and MSE Metrics}
\label{appssec:mse_mape}

\begin{figure}[htbp]
    \centering
    \begin{subfigure}[b]{0.48\textwidth}
        \centering
        \includegraphics[width=\textwidth]{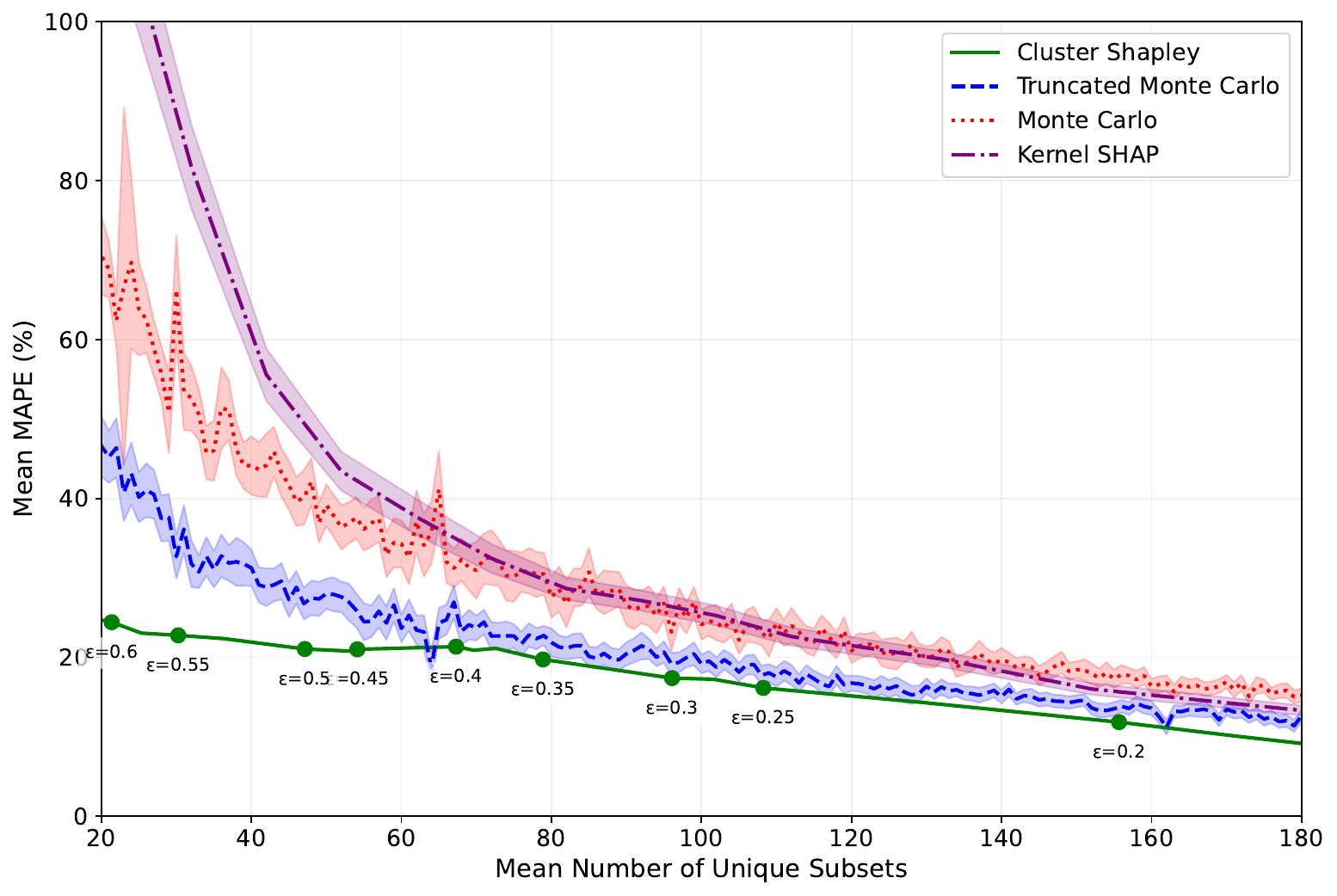}
        \caption{MAPE Performance}
        \label{fig:performance_mape}
    \end{subfigure}
    \hfill
    \begin{subfigure}[b]{0.48\textwidth}
        \centering
        \includegraphics[width=\textwidth]{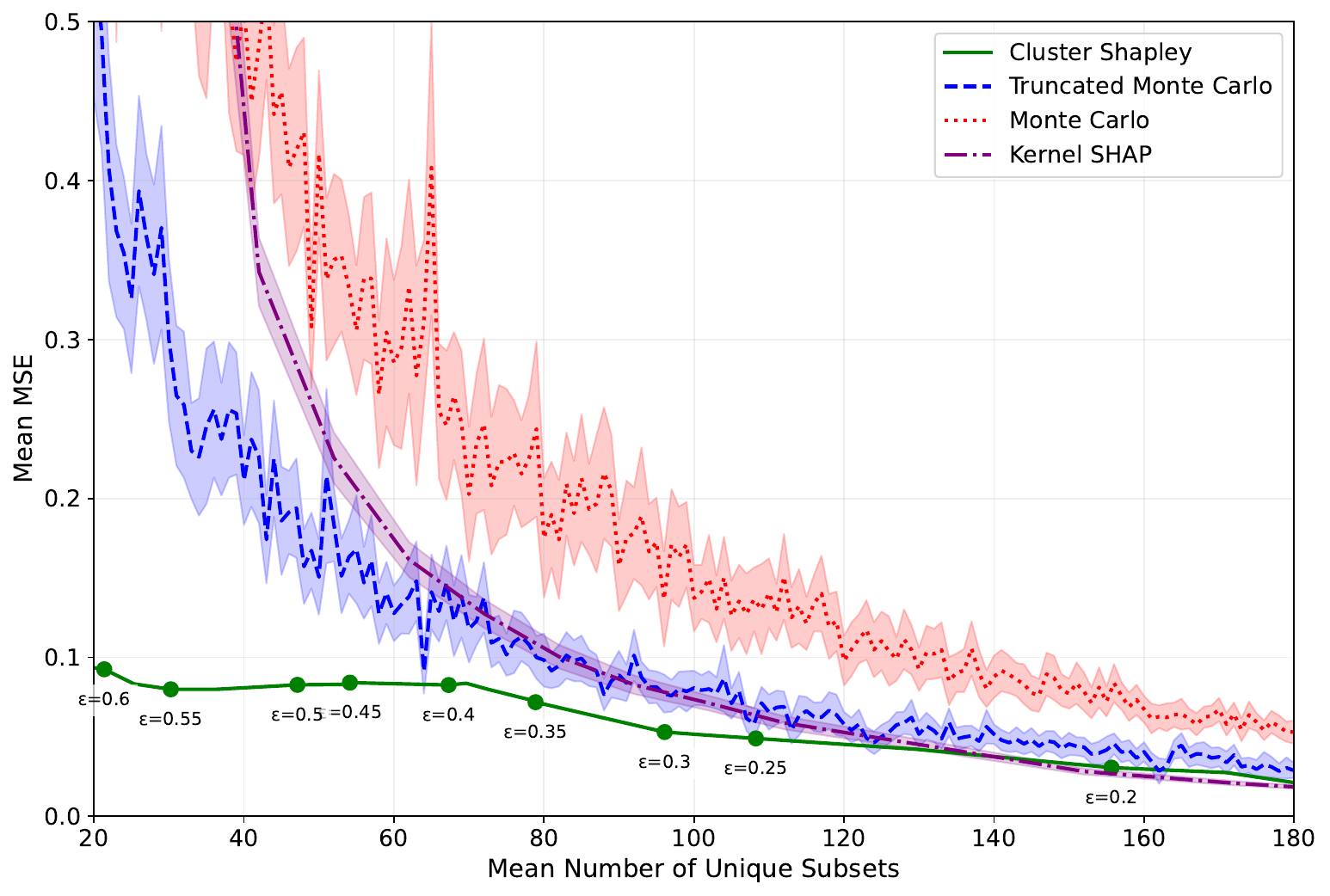}
        \caption{MSE Performance}
        \label{fig:performance_mse}
    \end{subfigure}

    \vspace{1em} % Add some vertical space between rows
    
    % Second row, single figure
    \begin{subfigure}[b]{0.48\textwidth} % Wider figure
        \centering
        \includegraphics[width=\textwidth]{figures/algorithm_comparison_mae_with_ci_strict_DBSCAN.pdf}
        \caption{MAE Performance}
        \label{fig:performance_mae}
    \end{subfigure}
    
    \caption{Efficient frontiers of algorithms under MAPE, MSE, and MAE measures. Subfigure (c) replicates Figure~\ref{fig:benchmark_comparison}. The $x$-axis represents the number of unique subsets used by the algorithms, averaged across all test queries and reviews. The $y$-axis represents the Mean error measures of the Shapley values, averaged across all test queries and reviews.}
    \label{fig:performance_comparison_strict_mse_mape}
\end{figure}

For completeness, we evaluate the performance of the Cluster Shapley algorithm using alternative error metrics, MAPE and MSE, and compare it to other benchmark methods in Figure~\ref{fig:performance_comparison_strict_mse_mape}. For ease of comparison, we replicate the results from Figure~\ref{fig:benchmark_comparison} as panel (c) in Figure~\ref{fig:performance_comparison_strict_mse_mape}. As shown, Cluster Shapley consistently outperforms the benchmarks across all three metrics.

\subsection{Cluster Shapley using the Standard DBSCAN}
\label{appssec:standard_DBSCAN}

\begin{figure}[htbp]
    \centering
    % First row, first figure
    \begin{subfigure}[b]{0.48\textwidth}
        \centering
        \includegraphics[width=\textwidth]{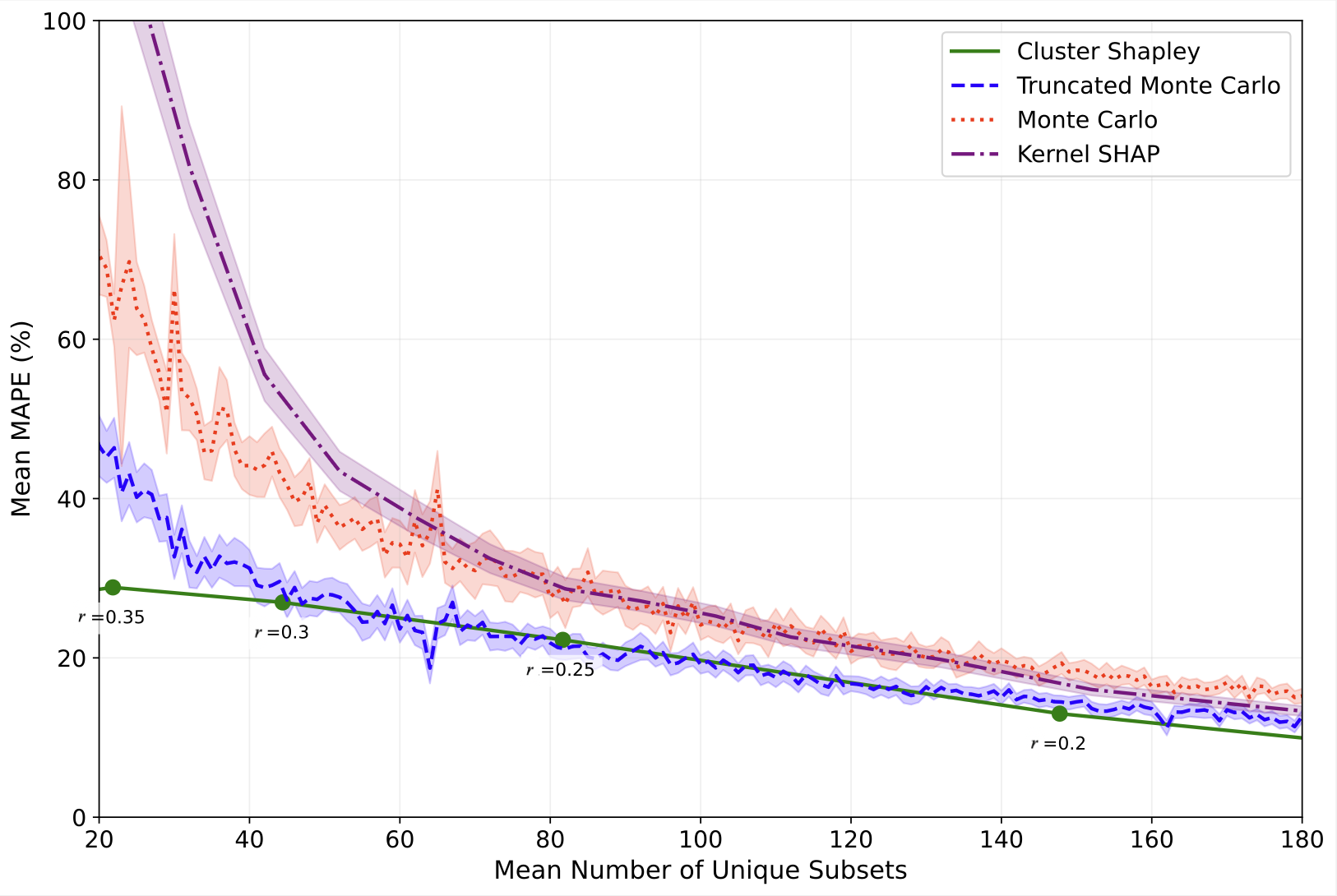}
        \caption{MAPE Performance}
        \label{fig:standard_mae_comparison}
    \end{subfigure}
    \hfill
    % First row, second figure
    \begin{subfigure}[b]{0.48\textwidth}
        \centering
        \includegraphics[width=\textwidth]{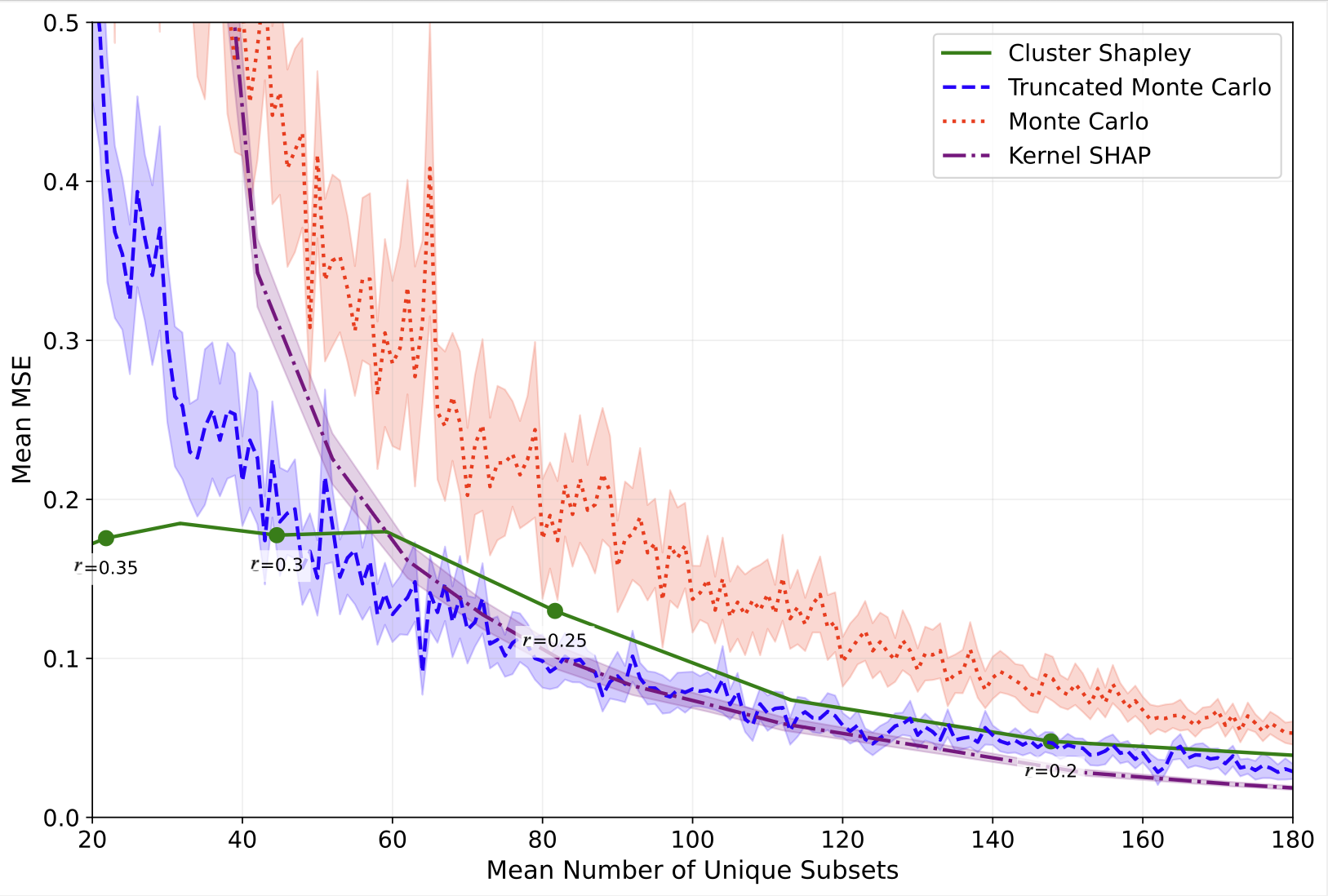}
        \caption{MSE Performance}
        \label{fig:standard_mape_comparison}
    \end{subfigure}
    
    \vspace{1em} % Add some vertical space between rows
    
    % Second row, single figure
    \begin{subfigure}[b]{0.48\textwidth} % Wider figure
        \centering
        \includegraphics[width=\textwidth]{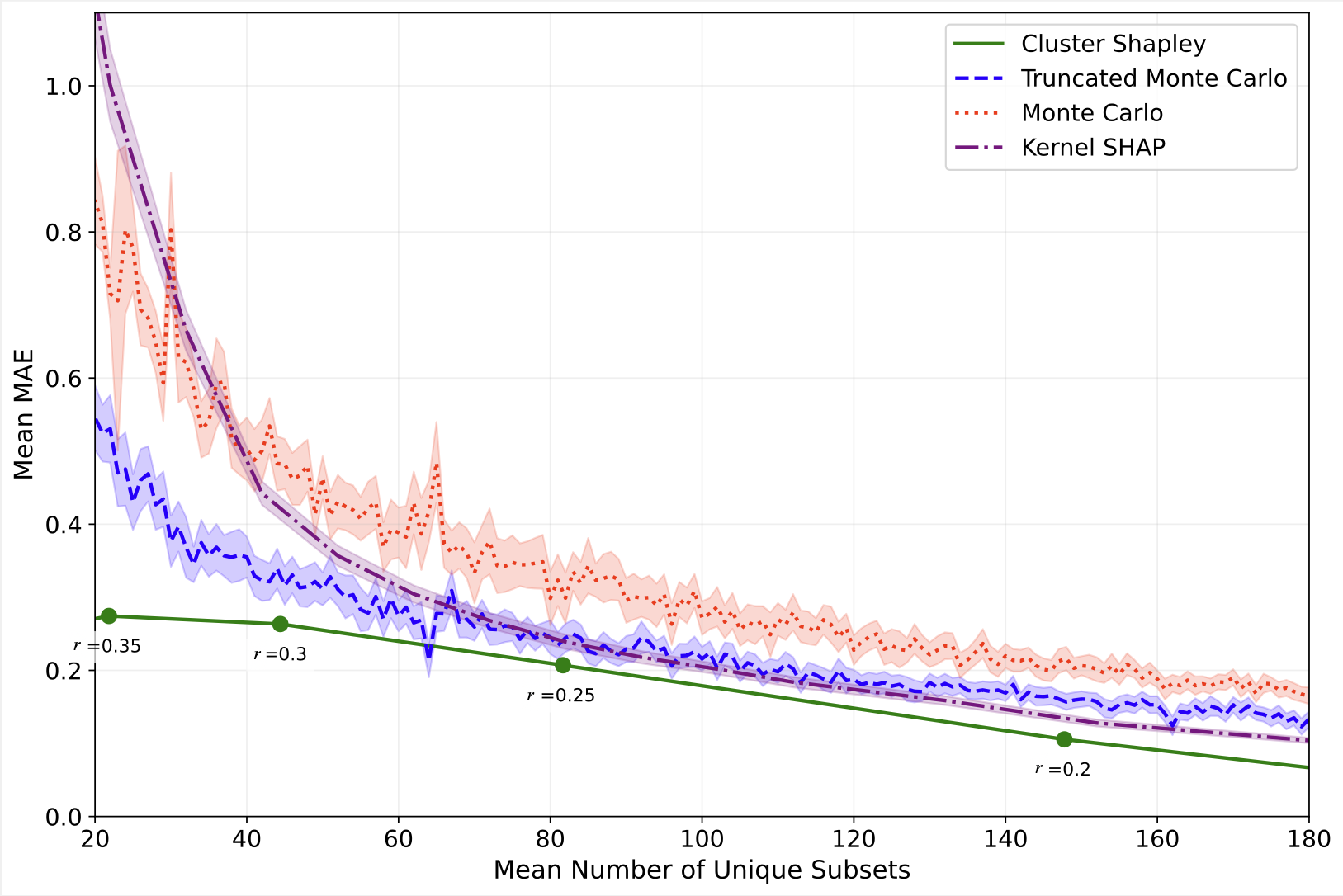}
        \caption{MAE Performance}
        \label{fig:standard_mse_comparison}
    \end{subfigure}
    
    \caption{Efficient frontiers of Cluster Shapley using the \textbf{standard DBSCAN} in Step 1 (clustering). Note that the performance curves for the other three algorithms are identical to those in Figure~\ref{fig:performance_comparison_strict_mse_mape}. Each point on the Cluster Shapley curve corresponds to a different neighborhood radius \(r\), which is the same as the clustering diameter \(\epsilon\) in the standard DBSCAN.}
    \label{fig:performance_comparison_standard}
\end{figure}

We now report the performance of the Cluster Shapley algorithm using the standard DBSCAN algorithm, instead of our proposed adaptive variant, in Figure~\ref{fig:performance_comparison_standard}. As discussed in \S\ref{ssec:cluster_shapley}, the standard DBSCAN does not guarantee that all pairs of documents within the same cluster have embedding distances strictly smaller than \(\epsilon\), which may affect approximation quality.

Comparing Figure~\ref{fig:performance_comparison_standard} (which uses standard DBSCAN for clustering) with Figure~\ref{fig:performance_comparison_strict_mse_mape} (which uses our proposed adaptive DBSCAN, i.e., Algorithm~\ref{alg:dbscan}), we observe that Cluster Shapley achieves consistently better performance with adaptive clustering across all three approximation error measures. Moreover, when using more than 60 unique subsets, Cluster Shapley with standard DBSCAN performs significantly worse than both Truncated Monte Carlo and Kernel SHAP, highlighting the importance of enforcing tighter clustering constraints through the adaptive procedure.%When fewer unique subsets are used (corresponding to a larger clustering diameter \(\epsilon\)), the standard DBSCAN leads to slightly worse performance. However, this performance gap narrows as the number of unique subsets increases. This pattern aligns with our objective of reducing computation time while maintaining accuracy.

\subsection{Sample Splitting Robustness Check}
\label{appssec:split_sample}

To assess the robustness of our results to the choice of test queries and the stability of hyperparameter tuning, we conduct a sample-splitting analysis. Specifically, we randomly divide the 48 test queries used in the main analysis into two equal subsets, referred to as Split 1 and Split 2, each containing 24 queries. We then replicate the algorithm comparison analysis (shown in Figure~\ref{fig:benchmark_comparison} of the main text) separately for each split.

\begin{figure}[htbp]
    \centering
    \begin{subfigure}[b]{0.48\textwidth}
        \centering
        \includegraphics[width=\textwidth]{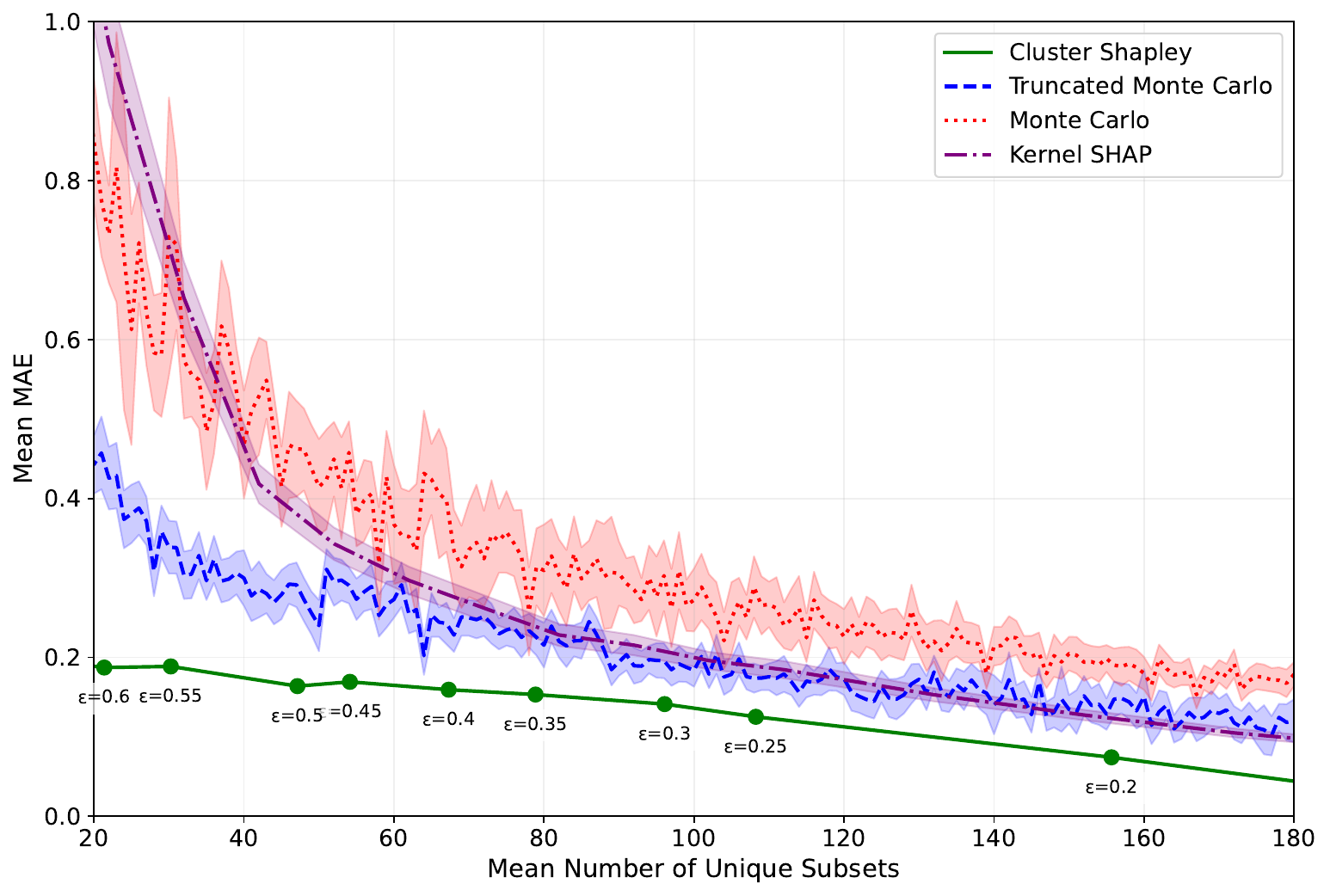}
        \caption{Split 1 (24 queries)}
        \label{fig:robustness_split1}
    \end{subfigure}
    \hfill
    \begin{subfigure}[b]{0.48\textwidth}
        \centering
        \includegraphics[width=\textwidth]{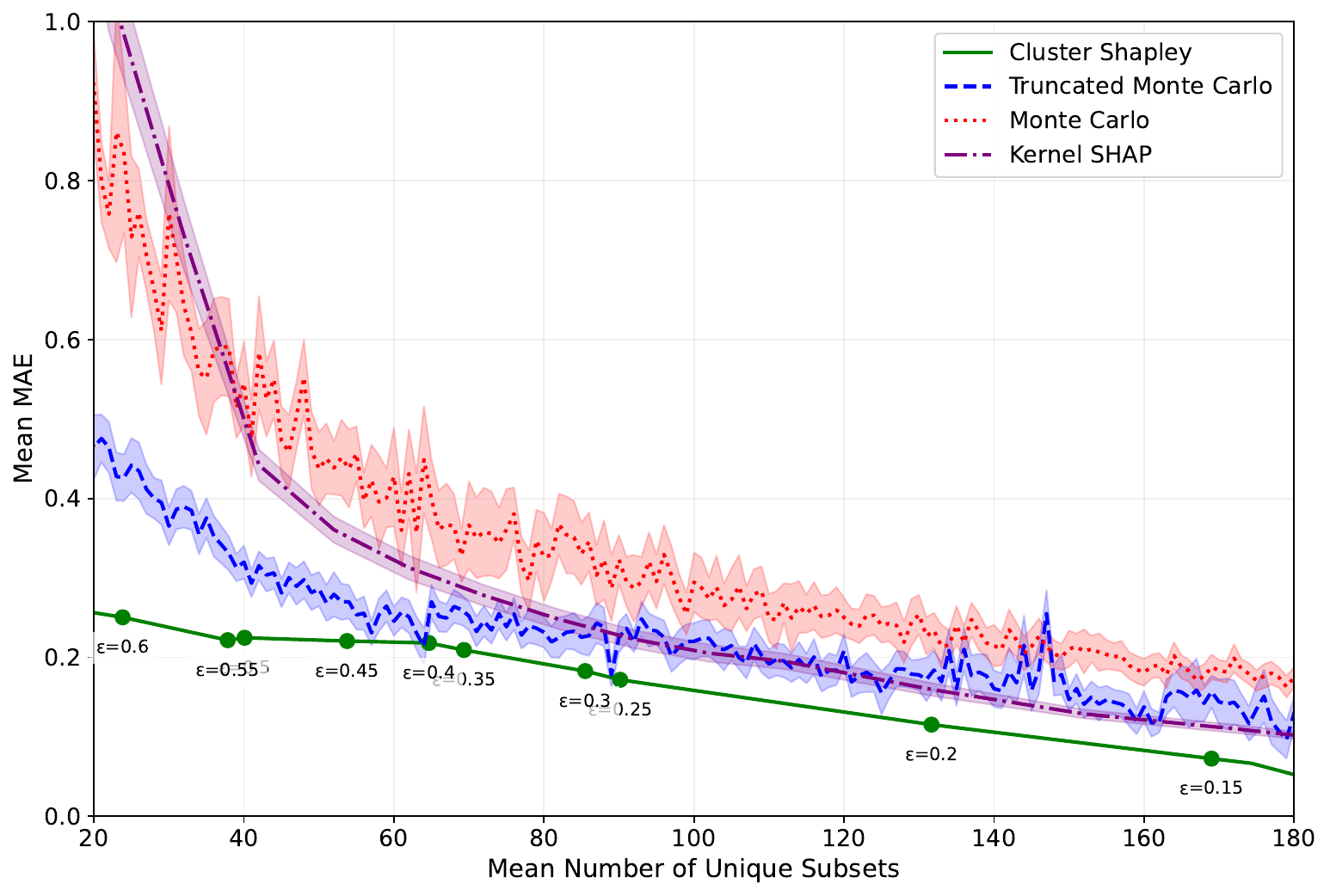}
        \caption{Split 2 (24 queries)}
        \label{fig:robustness_split2}
    \end{subfigure}
\caption{Robustness check based on random query splitting. Algorithm comparison results for Split 1 and Split 2 remain consistent, confirming that the main findings in Figure~\ref{fig:benchmark_comparison} are not sensitive to the choice of test queries.}
\label{fig:robustness_splits}
\end{figure}

The results, presented in Figure~\ref{fig:robustness_splits}, yield two key findings. First, the comparative performance of different algorithms remains consistent across the two query subsets, indicating that the results are not driven by specific queries. Second, the choice of the clustering diameter $\epsilon$ is stable across splits, suggesting that one subset can be reliably used for hyperparameter tuning without compromising performance on the other subset.

\subsection{Cluster Shapley with Cluster-Level Approximation Algorithms}
\label{appssec:cluster_approx}

While the proposed Cluster Shapley algorithm substantially reduces computational complexity by limiting the number of unique subsets requiring evaluation, certain real-world applications may involve substantially larger document sets. To further enhance efficiency in such settings, we explore the integration of approximation algorithms within Cluster Shapley to approximate cluster-level Shapley values. Specifically, in this appendix, we evaluate Algorithm~\ref{alg:cluster shap} by incorporating the Monte-Carlo approximation algorithm in Step 2.

We present two experiments. The first experiment (10 documents) replicates the main benchmark setting but increases the number of documents per query to reflect more computationally intensive scenarios. The second experiment scales up to 30 documents per query to assess the algorithm’s behavior under much larger document sets, where exact Shapley computation is infeasible.

\paragraph{Experiment 1: 10 Documents with LLM-Based Evaluation.}  
We randomly sample two queries from our benchmark dataset (“How quick is the delivery of the gift card?” and “How’s the quality of the PlayStation?”), each associated with 10 reviews (instead of 8 in the main analysis). This increase results in a total of $2^{10}-1 = 1{,}023$ coalitions per query, leading to nearly 300,000 tokens processed by LLMs.

For the standard Cluster Shapley algorithm, we use the implementation described in the main paper. For the Monte Carlo–based variant, we vary two hyperparameters: the clustering diameter $\epsilon \in [0.01, 1.0]$ and the number of Monte Carlo permutations used in Step 2. We evaluate all combinations of these hyperparameters and report the efficient frontier, i.e., the lowest approximation error achieved for each level of computational cost (measured by the number of unique subset evaluations).

\begin{figure}[htbp]
    \centering
    \includegraphics[width=0.7\textwidth]{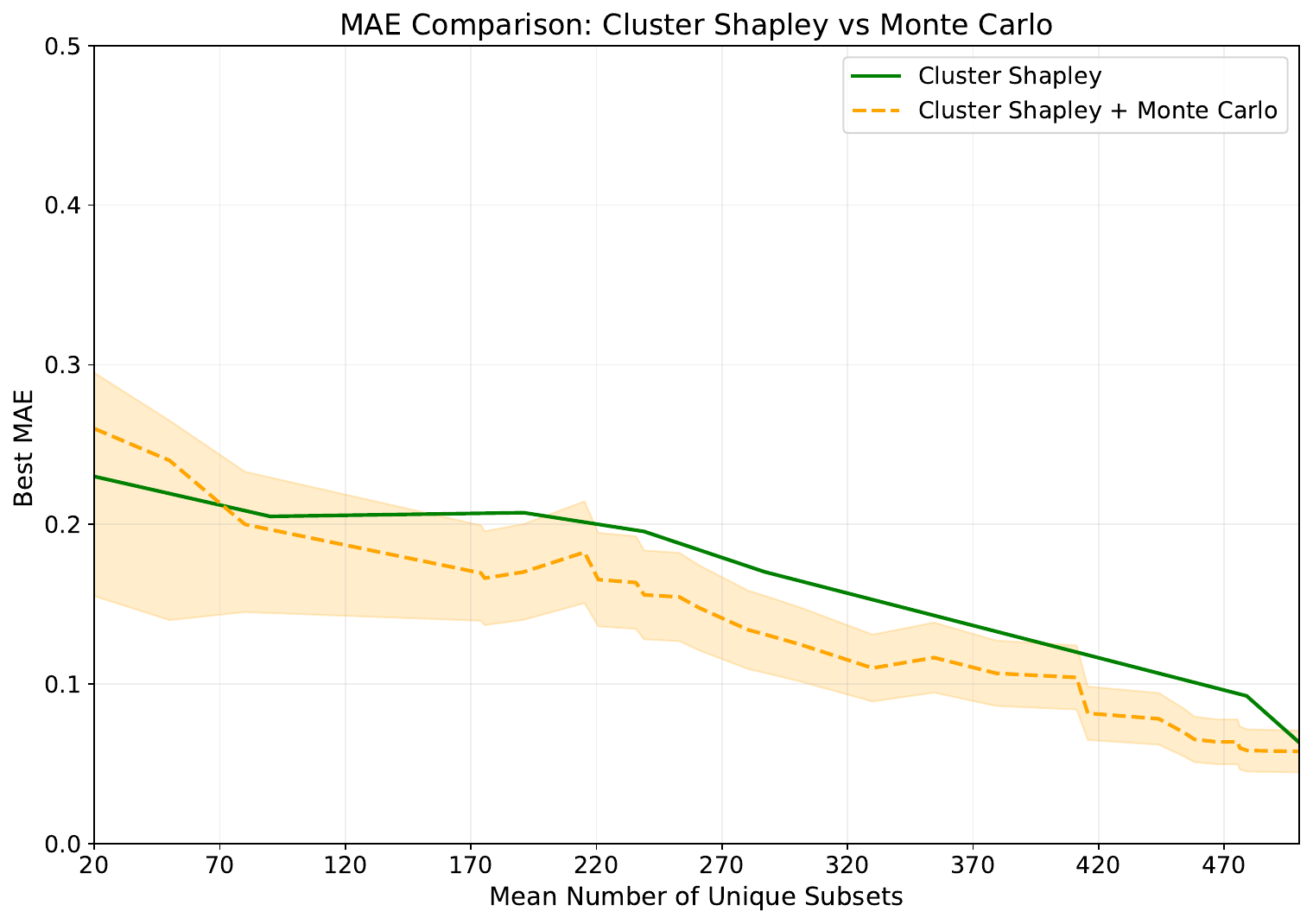}
    \caption{Efficient frontiers of Cluster Shapley and Cluster Shapley with Monte Carlo sampling (10 documents). The $x$-axis shows the number of unique subsets evaluated (averaged over 10 replications), and the $y$-axis reports the Mean Absolute Error (MAE) relative to exact Shapley values computed with LLM evaluation. 95\% confidence intervals are shown.}
    \label{fig:cluster_shap+approx}
\end{figure}

Figure~\ref{fig:cluster_shap+approx} shows that Cluster Shapley with Monte Carlo consistently achieves lower approximation error than the standard version for the same computational cost (though the difference is not statistically significant below 170 subset evaluations), suggesting that introducing Monte Carlo sampling in Step 2 improves the efficiency–accuracy trade-off.

\paragraph{Experiment 2: 30 Documents with Synthetic Evaluation.}  
To evaluate Cluster Shapley in larger-scale settings, we extend the experiment to queries with 30 documents. Exact Shapley value computation becomes intractable at this scale ($2^{30} \approx 10^9$ subsets), so we generate synthetic summarization scores for evaluation.

Specifically, we assign each document a true Shapley value linearly proportional to its cosine similarity with the query embedding, normalized to sum to 10. The value function $v(q, A(q, S))$ for any subset $S$ is then computed as the sum of the true Shapley values of documents in $S$, plus additive Gaussian noise. We simulate four levels of noise with standard deviation $\sigma \in \{0.05, 0.10, 0.15, 0.20\}$, all with mean zero. This setup allows us to evaluate the robustness and scalability of Cluster Shapley with Monte Carlo sampling under noisy yet controlled evaluation.

\begin{figure}[htbp]
    \centering
    \includegraphics[width=1\textwidth]{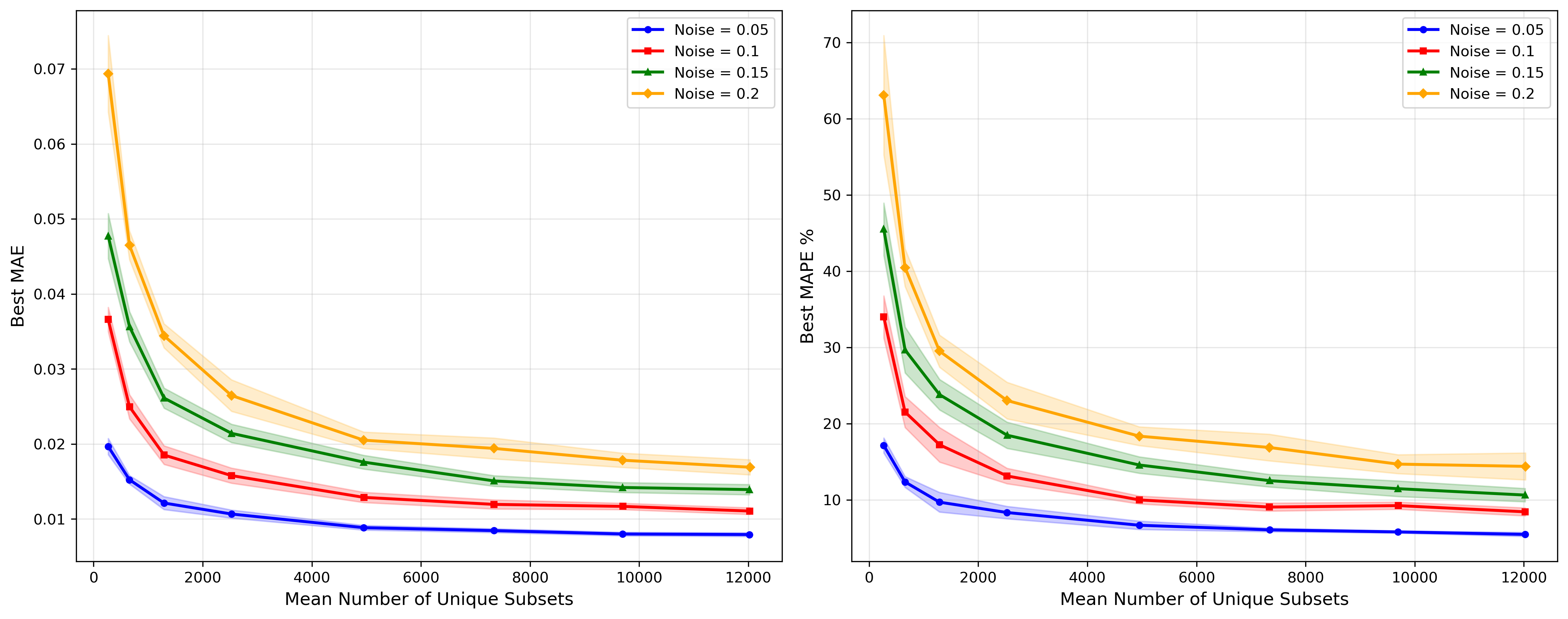}
    \caption{MAE and MAPE of Cluster Shapley with Monte Carlo sampling under four noise levels (30 documents). The $x$-axis shows the number of unique subsets evaluated. The $y$-axes report the Mean Absolute Error (MAE) and Mean Absolute Percentage Error (MAPE) relative to the true synthetic Shapley values. Each line corresponds to a different level of Gaussian noise added to the evaluation.}
    \label{fig:cluster_shap_MC_noise}
\end{figure}

As shown in Figure~\ref{fig:cluster_shap_MC_noise}, Cluster Shapley with Monte Carlo remains effective even in large-scale settings with noisy evaluation. As expected, higher noise levels slightly degrade performance, but the method still achieves low MAE and MAPE with a reasonable number of subset evaluations.

% ==== arXiv/SSRN: bibliography inlined from bu2.bbl (no bibtex / no .bbl needed) ====

% ==== end inlined bu2 ====

\end{appendices}

\end{bibunit}


\begin{thebibliography}{84}
\providecommand{\natexlab}[1]{#1}
\providecommand{\url}[1]{\texttt{#1}}
\expandafter\ifx\csname urlstyle\endcsname\relax
  \providecommand{\doi}[1]{doi: #1}\else
  \providecommand{\doi}{doi: \begingroup \urlstyle{rm}\Url}\fi

\bibitem[{AI/ML API}(2025)]{aimlapi2025}
{AI/ML API}.
\newblock {AI/ML API} inference pricing, 2025.
\newblock \href{https://aimlapi.com/ai-ml-api-pricing}{Link}. Accessed on May,
  2025.

\bibitem[Amaldoss and Du(2023)]{amaldoss2023can}
Wilfred Amaldoss and Jinzhao Du.
\newblock How can publishers collaborate and compete with news aggregators?
\newblock \emph{Journal of Marketing Research}, 60\penalty0 (6):\penalty0
  1114--1134, 2023.

\bibitem[Athey et~al.(2021)Athey, Mobius, and Pal]{athey2021impact}
Susan Athey, Markus Mobius, and Jeno Pal.
\newblock The impact of aggregators on internet news consumption.
\newblock Technical report, National Bureau of Economic Research, 2021.

\bibitem[Bachrach et~al.(2010)Bachrach, Markakis, Resnick, Procaccia,
  Rosenschein, and Saberi]{bachrach2010approximating}
Yoram Bachrach, Evangelos Markakis, Ezra Resnick, Ariel~D Procaccia, Jeffrey~S
  Rosenschein, and Amin Saberi.
\newblock Approximating power indices: theoretical and empirical analysis.
\newblock \emph{Autonomous Agents and Multi-Agent Systems}, 20\penalty0
  (2):\penalty0 105--122, 2010.

\bibitem[Barshan et~al.(2020)Barshan, Brunet, and
  Dziugaite]{barshan2020relatif}
Elnaz Barshan, Marc-Etienne Brunet, and Gintare~Karolina Dziugaite.
\newblock Relatif: Identifying explanatory training samples via relative
  influence.
\newblock In \emph{International Conference on Artificial Intelligence and
  Statistics}, pages 1899--1909. PMLR, 2020.

\bibitem[Bhargava(2022)]{bhargava2022creator}
Hemant~K Bhargava.
\newblock The creator economy: Managing ecosystem supply, revenue sharing, and
  platform design.
\newblock \emph{Management Science}, 68\penalty0 (7):\penalty0 5233--5251,
  2022.

\bibitem[Calzada and Gil(2020)]{calzada2020news}
Joan Calzada and Ricard Gil.
\newblock What do news aggregators do? evidence from google news in spain and
  germany.
\newblock \emph{Marketing Science}, 39\penalty0 (1):\penalty0 134--167, 2020.

\bibitem[Carroll(2025)]{carroll2025aioverview}
Sarah Carroll.
\newblock Will google’s ai overview kill web traffic?, January 2025.
\newblock
  \href{https://www.forumone.com/insights/blog/will-googles-ai-overview-kill-web-traffic/}{Link}.
  Accessed on May, 2025.

\bibitem[Castro et~al.(2009)Castro, G{\'o}mez, and
  Tejada]{castro2009polynomial}
Javier Castro, Daniel G{\'o}mez, and Juan Tejada.
\newblock Polynomial calculation of the shapley value based on sampling.
\newblock \emph{Computers \& operations research}, 36\penalty0 (5):\penalty0
  1726--1730, 2009.

\bibitem[Chang(2023)]{chang2023yelp}
Ashley Chang.
\newblock How i became yelp elite in 48 days, August 2023.
\newblock
  \href{https://ashleychangg.medium.com/how-i-became-yelp-elite-in-48-days-367abaa3879}{Link}.
  Accessed on May, 2025.

\bibitem[Chee(2025)]{ipa2025google}
Foo~Yun Chee.
\newblock Exclusive: Google’s {AI} {O}verviews hit by {EU} antitrust
  complaint from independent publishers, 2025.
\newblock
  \href{https://www.reuters.com/legal/litigation/googles-ai-overviews-hit-by-eu-antitrust-complaint-independent-publishers-2025-07-04/}{Link}.
  Accessed on Nov, 2025.

\bibitem[Cheng et~al.(2025)Cheng, Ofek, and Yoganarasimhan]{cheng_etal_2025}
Magie Cheng, Elie Ofek, and Hema Yoganarasimhan.
\newblock Balancing engagement and polarization: Multi-objective alignment of
  news content using llms.
\newblock \emph{arXiv preprint arXiv:2504.13444}, 2025.

\bibitem[Chevalier and Mayzlin(2006)]{chevalier_mayzlin_2006}
Judith~A Chevalier and Dina Mayzlin.
\newblock The effect of word of mouth on sales: Online book reviews.
\newblock \emph{Journal of marketing research}, 43\penalty0 (3):\penalty0
  345--354, 2006.

\bibitem[{CNBC}(2026)]{cnbc2026alexashopping}
{CNBC}.
\newblock Amazon ditches {Rufus} chatbot, launches {Alexa} shopping agent in
  {AI} strategy pivot, 2026.
\newblock
  \href{https://www.cnbc.com/2026/05/13/amazon-ditches-rufus-ai-chatbot-in-favor-of-alexa-shopping-agent.html}{Link}.
  Accessed on June, 2026.

\bibitem[Cook(1977)]{cook1977detection}
R~Dennis Cook.
\newblock Detection of influential observation in linear regression.
\newblock \emph{Technometrics}, 19\penalty0 (1):\penalty0 15--18, 1977.

\bibitem[Corder and Decker(2019)]{corder2019shapley}
Kevin Corder and Keith Decker.
\newblock Shapley value approximation with divisive clustering.
\newblock In \emph{2019 18th IEEE International Conference On Machine Learning
  And Applications (ICMLA)}, pages 234--239. IEEE, 2019.

\bibitem[Cripps et~al.(2004)Cripps, Mailath, and Samuelson]{cripps_etal_2004}
Martin~W Cripps, George~J Mailath, and Larry Samuelson.
\newblock Imperfect monitoring and impermanent reputations.
\newblock \emph{Econometrica}, 72\penalty0 (2):\penalty0 407--432, 2004.

\bibitem[Dellarocas et~al.(2013)Dellarocas, Katona, and
  Rand]{dellarocas2013media}
Chrysanthos Dellarocas, Zsolt Katona, and William Rand.
\newblock Media, aggregators, and the link economy: Strategic hyperlink
  formation in content networks.
\newblock \emph{Management science}, 59\penalty0 (10):\penalty0 2360--2379,
  2013.

\bibitem[Edge et~al.(2024)Edge, Trinh, Cheng, Bradley, Chao, Mody, Truitt, and
  Larson]{edge2024local}
Darren Edge, Ha~Trinh, Newman Cheng, Joshua Bradley, Alex Chao, Apurva Mody,
  Steven Truitt, and Jonathan Larson.
\newblock From local to global: A graph rag approach to query-focused
  summarization.
\newblock \emph{arXiv preprint arXiv:2404.16130}, 2024.

\bibitem[Ester et~al.(1996)Ester, Kriegel, Sander, and Xu]{ester1996density}
Martin Ester, Hans-Peter Kriegel, J{\"o}rg Sander, and Xiaowei Xu.
\newblock A density-based algorithm for discovering clusters in large spatial
  databases with noise.
\newblock In \emph{Proceedings of the Second International Conference on
  Knowledge Discovery and Data Mining (KDD)}, pages 226--231. AAAI Press, 1996.

\bibitem[Famà et~al.(2024)Famà, Myftiu, Pagnottoni, and
  Spelta]{fama2024explainable}
Angelo Famà, Jurgena Myftiu, Paolo Pagnottoni, and Andrea Spelta.
\newblock Explainable machine learning for financial risk management: Two
  practical use cases.
\newblock \emph{Statistics}, 2024.

\bibitem[Fan et~al.(2024)Fan, Ding, Ning, Wang, Li, Yin, Chua, and
  Li]{fan2024survey}
Wenqi Fan, Yujuan Ding, Liangbo Ning, Shijie Wang, Hengyun Li, Dawei Yin,
  Tat-Seng Chua, and Qing Li.
\newblock A survey on rag meeting llms: Towards retrieval-augmented large
  language models.
\newblock In \emph{Proceedings of the 30th ACM SIGKDD Conference on Knowledge
  Discovery and Data Mining}, pages 6491--6501, 2024.

\bibitem[Fatima et~al.(2008)Fatima, Wooldridge, and Jennings]{fatima2008linear}
Shaheen~S Fatima, Michael Wooldridge, and Nicholas~R Jennings.
\newblock A linear approximation method for the {S}hapley value.
\newblock \emph{Artificial Intelligence}, 172\penalty0 (14):\penalty0
  1673--1699, 2008.

\bibitem[Ghorbani and Zou(2019)]{ghorbani2019data}
Amirata Ghorbani and James Zou.
\newblock Data shapley: Equitable valuation of data for machine learning.
\newblock In \emph{International conference on machine learning}, pages
  2242--2251. PMLR, 2019.

\bibitem[Goodwin(2024{\natexlab{a}})]{Goodwin2024}
Danny Goodwin.
\newblock Google ceo says ai overviews are increasing search usage, April
  2024{\natexlab{a}}.
\newblock
  \href{https://searchengineland.com/google-ceo-says-ai-overviews-are-increasing-search-usage-439983}{Link}.
  Accessed on May, 2025.

\bibitem[Goodwin(2024{\natexlab{b}})]{perplexity_citation_count}
Danny Goodwin.
\newblock 60\% of perplexity citations overlap with top 10 google organic
  results, 2024{\natexlab{b}}.
\newblock
  \href{https://searchengineland.com/perplexity-citations-top-10-google-organic-results-439029}{Link}.
  Accessed on July, 2025.

\bibitem[{Google}(2025)]{google2025ai-overviews}
{Google}.
\newblock {AI} overviews – search anything, effortlessly, 2025.
\newblock \href{https://search.google/ways-to-search/ai-overviews/}{Link}.
  Accessed on May, 2025.

\bibitem[Gosiewska and Biecek(2019)]{gosiewska2019not}
Alicja Gosiewska and Przemyslaw Biecek.
\newblock Do not trust additive explanations.
\newblock \emph{arXiv preprint arXiv:1903.11420}, 2019.

\bibitem[Grynbaum and Mac(2023)]{nyt_openai_lawsuit_2023}
Michael~M. Grynbaum and Ryan Mac.
\newblock The new york times sues {OpenAI} and {Microsoft} over content use,
  2023.
\newblock
  \href{https://www.nytimes.com/2023/12/27/business/media/new-york-times-open-ai-microsoft-lawsuit.html}{Link}.
  Accessed on May, 2025.

\bibitem[Guo et~al.(2020)Guo, Rajani, Hase, Bansal, and Xiong]{guo2020fastif}
Han Guo, Nazneen~Fatema Rajani, Peter Hase, Mohit Bansal, and Caiming Xiong.
\newblock Fastif: Scalable influence functions for efficient model
  interpretation and debugging.
\newblock \emph{arXiv preprint arXiv:2012.15781}, 2020.

\bibitem[Gupta(2025)]{google_citation_count}
Pragati Gupta.
\newblock 40.58\% of {AI} citations come from google’s top 10 results (study
  of 1m+ ai overviews), 2025.
\newblock
  \href{https://writesonic.com/blog/ai-citations-from-serp-results-study}{Link}.
  Accessed on July, 2025.

\bibitem[Han et~al.(2020)Han, Wallace, and Tsvetkov]{han2020explaining}
Xiaochuang Han, Byron~C Wallace, and Yulia Tsvetkov.
\newblock Explaining black box predictions and unveiling data artifacts through
  influence functions.
\newblock \emph{arXiv preprint arXiv:2005.06676}, 2020.

\bibitem[Haque et~al.(2018)Haque, Saber, and Shah]{haque2018sentiment}
Tanjim~Ul Haque, Nudrat~Nawal Saber, and Faisal~Muhammad Shah.
\newblock Sentiment analysis on large scale amazon product reviews.
\newblock In \emph{2018 IEEE international conference on innovative research
  and development (ICIRD)}, pages 1--6. IEEE, 2018.

\bibitem[Hou et~al.(2024)Hou, Li, He, Yan, Chen, and McAuley]{hou2024bridging}
Yupeng Hou, Jiacheng Li, Zhankui He, An~Yan, Xiusi Chen, and Julian McAuley.
\newblock Bridging language and items for retrieval and recommendation.
\newblock \emph{arXiv preprint arXiv:2403.03952}, 2024.

\bibitem[Jeon and Nasr(2016)]{jeon2016news}
Doh-Shin Jeon and Nikrooz Nasr.
\newblock News aggregators and competition among newspapers on the internet.
\newblock \emph{American Economic Journal: Microeconomics}, 8\penalty0
  (4):\penalty0 91--114, 2016.

\bibitem[Jia et~al.(2019)Jia, Dao, Wang, Hubis, Hynes, G{\"u}rel, Li, Zhang,
  Song, and Spanos]{jia2019towards}
Ruoxi Jia, David Dao, Boxin Wang, Frances~Ann Hubis, Nick Hynes, Nezihe~Merve
  G{\"u}rel, Bo~Li, Ce~Zhang, Dawn Song, and Costas~J Spanos.
\newblock Towards efficient data valuation based on the shapley value.
\newblock In \emph{The 22nd International Conference on Artificial Intelligence
  and Statistics}, pages 1167--1176. PMLR, 2019.

\bibitem[Kang et~al.(2023)Kang, Ni, Mehta, Sathiamoorthy, Hong, Chi, and
  Cheng]{kang2023llms}
Wang-Cheng Kang, Jianmo Ni, Nikhil Mehta, Maheswaran Sathiamoorthy, Lichan
  Hong, Ed~Chi, and Derek~Zhiyuan Cheng.
\newblock Do llms understand user preferences? evaluating llms on user rating
  prediction.
\newblock \emph{arXiv preprint arXiv:2305.06474}, 2023.

\bibitem[Khosravi and Yoganarasimhan(2026)]{khosravi2026impact}
Mehrzad Khosravi and Hema Yoganarasimhan.
\newblock Impact of ai search summaries on website traffic: Evidence from
  google ai overviews and wikipedia.
\newblock \emph{Available at SSRN 6164926}, 2026.

\bibitem[Kokalj et~al.(2021)Kokalj, Škrlj, Lavrač, Pollak, and
  Robnik-Šikonja]{kokalj2021bert}
Enja Kokalj, Blaž Škrlj, Nada Lavrač, Senja Pollak, and Marko
  Robnik-Šikonja.
\newblock Bert meets shapley: Extending shap explanations to transformer-based
  classifiers.
\newblock In \emph{EACL Hackashop on Explainability for NLP}, 2021.

\bibitem[Kong et~al.(2023)Kong, Li, Nassif, Fiez, Henao, and
  Chakrabarti]{kong2023neural}
Fanjie Kong, Yuan Li, Houssam Nassif, Tanner Fiez, Ricardo Henao, and Shreya
  Chakrabarti.
\newblock Neural insights for digital marketing content design.
\newblock In \emph{Proceedings of the ACM SIGKDD Conference on Knowledge
  Discovery and Data Mining (KDD)}, 2023.

\bibitem[Kroll et~al.(2017)Kroll, Huey, Barocas, Felten, Reidenberg, Robinson,
  and Yu]{kroll_etal_2017}
Joshua~A Kroll, Joanna Huey, Solon Barocas, Edward~W Felten, Joel~R Reidenberg,
  David~G Robinson, and Harlan Yu.
\newblock Accountable algorithms.
\newblock \emph{UNIVERSITY of PENNSYLVANIA LAW REVIEW}, pages 633--705, 2017.

\bibitem[Lewis et~al.(2020)Lewis, Perez, Piktus, Petroni, Karpukhin, Goyal,
  K{\"u}ttler, Lewis, Yih, Rockt{\"a}schel, et~al.]{lewis2020retrieval}
Patrick Lewis, Ethan Perez, Aleksandra Piktus, Fabio Petroni, Vladimir
  Karpukhin, Naman Goyal, Heinrich K{\"u}ttler, Mike Lewis, Wen-tau Yih, Tim
  Rockt{\"a}schel, et~al.
\newblock Retrieval-augmented generation for knowledge-intensive nlp tasks.
\newblock \emph{Advances in Neural Information Processing Systems},
  33:\penalty0 9459--9474, 2020.

\bibitem[Li et~al.(2022)Li, Ong, Oei, Lian, Phua, Htet, Lim, and
  Motani]{li2022unified}
Anthony Li, Ming~Lun Ong, Chien~Wei Oei, Weixiang Lian, Hwee~Pin Phua, Lin~Htun
  Htet, Wei~Yen Lim, and Mehul Motani.
\newblock Unified auto clinical scoring (uni-acs) with interpretable ml models.
\newblock In \emph{Machine Learning for Healthcare Conference (MLHC)}, 2022.

\bibitem[Liu et~al.(2009)]{liu_2009}
Tie-Yan Liu et~al.
\newblock Learning to rank for information retrieval.
\newblock \emph{Foundations and Trends{\textregistered} in Information
  Retrieval}, 3\penalty0 (3):\penalty0 225--331, 2009.

\bibitem[Lloyd(1982)]{lloyd1982least}
Stuart Lloyd.
\newblock Least squares quantization in pcm.
\newblock \emph{IEEE transactions on information theory}, 28\penalty0
  (2):\penalty0 129--137, 1982.

\bibitem[Lundberg and Lee(2017)]{lundberg2017unified}
Scott~M. Lundberg and Su-In Lee.
\newblock A unified approach to interpreting model predictions.
\newblock In \emph{The 31st International Conference on Neural Information
  Processing Systems}, pages 4768--4777. PMLR, 2017.

\bibitem[Maleki et~al.(2013)Maleki, Tran-Thanh, Hines, Rahwan, and
  Rogers]{maleki2013bounding}
Sasan Maleki, Long Tran-Thanh, Greg Hines, Talal Rahwan, and Alex Rogers.
\newblock Bounding the estimation error of sampling-based shapley value
  approximation.
\newblock \emph{arXiv preprint arXiv:1306.4265}, 2013.

\bibitem[Mann and Shapley(1960)]{mann1960values}
Irwin Mann and Lloyd~S. Shapley.
\newblock Values of large games, {IV}: {E}valuating the {E}lectoral {C}ollege
  by {M}ontecarlo {T}echniques.
\newblock Technical Report RM-2651, RAND Corporation, Santa Monica, CA, 1960.

\bibitem[Mayzlin and Yoganarasimhan(2012)]{mayzlin2012link}
Dina Mayzlin and Hema Yoganarasimhan.
\newblock Link to success: How blogs build an audience by promoting rivals.
\newblock \emph{Management Science}, 58\penalty0 (9):\penalty0 1651--1668,
  2012.

\bibitem[Mehta and Chilimbi(2024)]{amazon_rufus}
Rajiv Mehta and Trishul Chilimbi.
\newblock Amazon announces {Rufus}, a new generative {AI}-powered
  conversational shopping experience, 2024.
\newblock \href{https://www.aboutamazon.com/news/retail/amazon-rufus}{Link}.
  Accessed on May, 2025.

\bibitem[{Microsoft}(2025)]{microsoft2025copilotsearch}
{Microsoft}.
\newblock Copilot search in bing, 2025.
\newblock
  \href{https://www.microsoft.com/en-us/bing/copilot-search/?form=MA13XW&cs=82041551}{Link}.
  Accessed on May, 2025.

\bibitem[Mihalcea and Tarau(2004)]{mihalcea2004textrank}
Rada Mihalcea and Paul Tarau.
\newblock Textrank: Bringing order into text.
\newblock In \emph{Proceedings of the 2004 conference on empirical methods in
  natural language processing}, pages 404--411, 2004.

\bibitem[Mikolov et~al.(2013)Mikolov, Chen, Corrado, and
  Dean]{mikolov2013efficient}
Tomas Mikolov, Kai Chen, Greg Corrado, and Jeffrey Dean.
\newblock Efficient estimation of word representations in vector space.
\newblock \emph{arXiv preprint arXiv:1301.3781}, 2013.

\bibitem[Mohammadi(2024)]{mohammadi2024wait}
Behnam Mohammadi.
\newblock Explaining large language models decisions using shapley values.
\newblock \emph{arXiv preprint arXiv:2404.01332}, 2024.

\bibitem[OpenAI(2024)]{searchgpt}
OpenAI.
\newblock Introducing {ChatGPT} search, 2024.
\newblock \href{https://openai.com/index/introducing-chatgpt-search/}{Link}.
  Accessed on May, 2025.

\bibitem[Patel et~al.(2025)Patel, Zhou, and Fanti]{patel2025maxshapley}
Sara Patel, Mingxun Zhou, and Giulia Fanti.
\newblock {MaxShapley}: Towards incentive-compatible generative search with
  fair context attribution.
\newblock \emph{arXiv preprint arXiv:2512.05958}, 2025.

\bibitem[Patil et~al.(2023)Patil, Boit, Gudivada, and
  Nandigam]{patil2023survey}
Rajvardhan Patil, Sorio Boit, Venkat Gudivada, and Jagadeesh Nandigam.
\newblock A survey of text representation and embedding techniques in {NLP}.
\newblock \emph{IEEE Access}, 2023.

\bibitem[Pennington et~al.(2014)Pennington, Socher, and
  Manning]{pennington2014glove}
Jeffrey Pennington, Richard Socher, and Christopher~D Manning.
\newblock Glove: Global vectors for word representation.
\newblock In \emph{Proceedings of the 2014 conference on empirical methods in
  natural language processing (EMNLP)}, pages 1532--1543, 2014.

\bibitem[{Perplexity AI}(2025)]{perplexityai2025}
{Perplexity AI}.
\newblock Perplexity {AI}, 2025.
\newblock \href{https://www.perplexity.ai}{Link}. Accessed on May, 2025.

\bibitem[{ProRata AI}(2025)]{prorata2025}
{ProRata AI}.
\newblock {ProRata AI}, 2025.
\newblock \href{https://prorata.ai}{Link}. Accessed: Aug, 2025.

\bibitem[Ragodos et~al.(2024)Ragodos, Wang, Feng, et~al.]{ragodos2024model}
Ronilo Ragodos, Tong Wang, Lu~Feng, et~al.
\newblock From model explanation to data misinterpretation: Uncovering the
  pitfalls of post hoc explainers in business research.
\newblock \emph{arXiv preprint arXiv:2408.16987}, 2024.

\bibitem[{Reddit Help}(2024)]{reddit_karma}
{Reddit Help}.
\newblock What is karma?, November 2024.
\newblock
  \href{https://support.reddithelp.com/hc/en-us/articles/204511829-What-is-karma}{Link}.
  Accessed on May, 2025.

\bibitem[Resnick et~al.(2000)Resnick, Kuwabara, Zeckhauser, and
  Friedman]{resnick_etal_2000}
Paul Resnick, Ko~Kuwabara, Richard Zeckhauser, and Eric Friedman.
\newblock Reputation systems.
\newblock \emph{Communications of the ACM}, 43\penalty0 (12):\penalty0 45--48,
  2000.

\bibitem[Robertson(2024)]{robertson2024openai}
Katie Robertson.
\newblock Openai strikes a deal to license news corp content, May 2024.
\newblock
  \href{https://www.nytimes.com/2024/05/22/business/media/openai-news-corp-content-deal.html}{Link}.
  Accessed on May, 2025.

\bibitem[Sax and Wang(2025)]{Sax_Wang_2025}
Marijn Sax and Hao Wang.
\newblock From threat to opportunity: Gaming the algorithmic system as a
  service.
\newblock \emph{Internet Policy Review}, 14\penalty0 (2), 2025.

\bibitem[Schermerhorn(2023)]{amazon_gen_ai_reviews}
Vaughn Schermerhorn.
\newblock How amazon continues to improve the customer reviews experience with
  generative {AI}, 2023.
\newblock
  \href{https://www.aboutamazon.com/news/amazon-ai/amazon-improves-customer-reviews-with-generative-ai}{Link}.
  Accessed on May, 2025.

\bibitem[Shankar et~al.(2024)Shankar, Zamfirescu-Pereira, Hartmann,
  Parameswaran, and Arawjo]{shankar2024validates}
Shreya Shankar, JD~Zamfirescu-Pereira, Bj{\"o}rn Hartmann, Aditya Parameswaran,
  and Ian Arawjo.
\newblock Who validates the validators? aligning llm-assisted evaluation of llm
  outputs with human preferences.
\newblock In \emph{Proceedings of the 37th Annual ACM Symposium on User
  Interface Software and Technology}, pages 1--14, 2024.

\bibitem[Shapley(1953)]{shapley1953value}
Lloyd~S Shapley.
\newblock A value for n-person games.
\newblock \emph{Contribution to the Theory of Games}, 2, 1953.

\bibitem[Sherrer(2025)]{sherrer2025misleading}
Kara Sherrer.
\newblock Google: It’s ‘misleading’ for websites to blame ai overviews
  for lost traffic, April 2025.
\newblock
  \href{https://www.eweek.com/news/google-ai-overviews-smb-impact/}{Link}.
  Accessed on May, 2025.

\bibitem[Srinivas(2025)]{srinivas2025}
Aravind Srinivas.
\newblock Perplexity has crossed \$100m in annualized revenue, March 2025.
\newblock
  \href{https://www.linkedin.com/posts/aravind-srinivas-16051987_perplexity-has-crossed-100m-in-annualized-activity-7310678232079495170-38ik/}{Link}.
  Accessed on May, 2025.

\bibitem[{Universal Music Group}(2024)]{prorataUMG2025}
{Universal Music Group}.
\newblock {ProRata} invents generative ai attribution technology to compensate
  and credit content owners while facilitating fairness and fact, with support
  from universal music group, 2024.
\newblock
  \href{https://www.universalmusic.com/prorata-invents-generative-ai-attribution-technology-to-compensate-and-credit-content-owners-while-facilitating-fairness-and-fact/}{Link}.
  Accessed: Aug, 2025.

\bibitem[Wang et~al.(2024)Wang, Deng, Chiba-Okabe, Barak, and
  Su]{wang2024economic}
Jiachen~T Wang, Zhun Deng, Hiroaki Chiba-Okabe, Boaz Barak, and Weijie~J Su.
\newblock An economic solution to copyright challenges of generative ai.
\newblock \emph{arXiv preprint arXiv:2404.13964}, 2024.

\bibitem[Wang et~al.(2025)Wang, Mittal, Song, and Jia]{wang2025datashapley}
Jiachen~T. Wang, Prateek Mittal, Dawn Song, and Ruoxi Jia.
\newblock Data {Shapley} in one training run.
\newblock In \emph{International Conference on Learning Representations
  (ICLR)}, 2025.

\bibitem[Wu and Zhu(2022)]{wu2022competition}
Yanhui Wu and Feng Zhu.
\newblock Competition, contracts, and creativity: Evidence from novel writing
  in a platform market.
\newblock \emph{Management Science}, 68\penalty0 (12):\penalty0 8613--8634,
  2022.

\bibitem[Xu et~al.(2023)Xu, Feng, and Chen]{xu2023chatgpt}
Ruiyun Xu, Yue Feng, and Hailiang Chen.
\newblock Chatgpt vs. google: A comparative study of search performance and
  user experience.
\newblock \emph{arXiv preprint arXiv:2307.01135}, 2023.

\bibitem[Ye et~al.(2025)Ye, Yoganarasimhan, and Zheng]{ye2025lola}
Zikun Ye, Hema Yoganarasimhan, and Yufeng Zheng.
\newblock {LOLA}: Llm-assisted online learning algorithm for content
  experiments.
\newblock \emph{Forthcoming in Marketing Science}, 2025.

\bibitem[Yoganarasimhan(2013)]{yoganarasimhan_2013}
Hema Yoganarasimhan.
\newblock The value of reputation in an online freelance marketplace.
\newblock \emph{Marketing Science}, 32\penalty0 (6):\penalty0 860--891, 2013.

\bibitem[Yoganarasimhan(2020)]{yoganarasimhan_2020}
Hema Yoganarasimhan.
\newblock Search personalization using machine learning.
\newblock \emph{Management Science}, 66\penalty0 (3):\penalty0 1045--1070,
  2020.

\bibitem[Zeff(2025)]{cloudflare2025}
Maxwell Zeff.
\newblock Cloudflare launches a marketplace that lets websites charge {AI} bots
  for scraping, 2025.
\newblock
  \href{https://techcrunch.com/2025/07/01/cloudflare-launches-a-marketplace-that-lets-websites-charge-ai-bots-for-scraping/}{Link}.
  Accessed: Aug, 2025.

\bibitem[Zhang et~al.(2024{\natexlab{a}})Zhang, Peng, Zhao, Hu, Zhu, Zeng, and
  Hu]{zhang2024llasa}
Shuo Zhang, Boci Peng, Xinping Zhao, Boren Hu, Yun Zhu, Yanjia Zeng, and Xuming
  Hu.
\newblock Llasa: Large language and e-commerce shopping assistant.
\newblock \emph{arXiv preprint arXiv:2408.02006}, 2024{\natexlab{a}}.

\bibitem[Zhang et~al.(2024{\natexlab{b}})Zhang, Wang, and
  Bhargava]{zhang2024if}
Xingyue~Luna Zhang, Kitty Wang, and Hemant~K Bhargava.
\newblock If platforms are exploiting producers, is platform competition the
  solution?
\newblock \emph{Available at SSRN 4728676}, 2024{\natexlab{b}}.

\bibitem[Zhang et~al.(2025)Zhang, Ma, Wang, He, Wang, He, Wang, and
  Liu]{zhang2025lexical}
Zhange Zhang, Yuqing Ma, Yulong Wang, Shan He, Tianbo Wang, Siqi He, Jiakai
  Wang, and Xianglong Liu.
\newblock Lexical diversity-aware relevance assessment for retrieval-augmented
  generation.
\newblock In \emph{Proceedings of the 63rd Annual Meeting of the Association
  for Computational Linguistics (Volume 1: Long Papers)}, pages 27758--27781,
  2025.

\bibitem[Zheng et~al.(2023)Zheng, Chiang, Sheng, Zhuang, Wu, Zhuang, Lin, Li,
  Li, Xing, Zhang, Gonzalez, and Stoica]{zheng2023judging}
Lianmin Zheng, Wei-Lin Chiang, Ying Sheng, Siyuan Zhuang, Zhanghao Wu, Yonghao
  Zhuang, Zi~Lin, Zhuohan Li, Dacheng Li, Eric~P. Xing, Hao Zhang, Joseph~E.
  Gonzalez, and Ion Stoica.
\newblock Judging {LLM}-as-a-judge with {MT}-bench and chatbot arena.
\newblock In \emph{Advances in Neural Information Processing Systems
  (NeurIPS)}, 2023.

\bibitem[Zhu(2025)]{zhu2025generative}
Yuting Zhu.
\newblock Generative search: Evidence from a large-scale field experiment.
\newblock \emph{Available at SSRN 5297194}, 2025.

\end{thebibliography}

\begin{thebibliography}{12}
\providecommand{\natexlab}[1]{#1}
\providecommand{\url}[1]{\texttt{#1}}
\expandafter\ifx\csname urlstyle\endcsname\relax
  \providecommand{\doi}[1]{doi: #1}\else
  \providecommand{\doi}{doi: \begingroup \urlstyle{rm}\Url}\fi

\bibitem[Gao et~al.(2023)Gao, Xiong, Gao, Jia, Pan, Bi, Dai, Sun, and
  Wang]{gao2023retrieval}
Yunfan Gao, Yun Xiong, Xinyu Gao, Kangxiang Jia, Jinliu Pan, Yuxi Bi, Yi~Dai,
  Jiawei Sun, and Haofen Wang.
\newblock Retrieval-augmented generation for large language models: A survey.
\newblock \emph{arXiv preprint arXiv:2312.10997}, 2023.

\bibitem[Ghorbani and Zou(2019)]{ghorbani2019data}
Amirata Ghorbani and James Zou.
\newblock Data shapley: Equitable valuation of data for machine learning.
\newblock In \emph{International conference on machine learning}, pages
  2242--2251. PMLR, 2019.

\bibitem[Goodwin(2024)]{perplexity_citation_count}
Danny Goodwin.
\newblock 60\% of perplexity citations overlap with top 10 google organic
  results, 2024.
\newblock
  \href{https://searchengineland.com/perplexity-citations-top-10-google-organic-results-439029}{Link}.
  Accessed on July, 2025.

\bibitem[Gupta(2025)]{google_citation_count}
Pragati Gupta.
\newblock 40.58\% of {AI} citations come from google’s top 10 results (study
  of 1m+ ai overviews), 2025.
\newblock
  \href{https://writesonic.com/blog/ai-citations-from-serp-results-study}{Link}.
  Accessed on July, 2025.

\bibitem[Lewis et~al.(2020)Lewis, Perez, Piktus, Petroni, Karpukhin, Goyal,
  K{\"u}ttler, Lewis, Yih, Rockt{\"a}schel, et~al.]{lewis2020retrieval}
Patrick Lewis, Ethan Perez, Aleksandra Piktus, Fabio Petroni, Vladimir
  Karpukhin, Naman Goyal, Heinrich K{\"u}ttler, Mike Lewis, Wen-tau Yih, Tim
  Rockt{\"a}schel, et~al.
\newblock Retrieval-augmented generation for knowledge-intensive nlp tasks.
\newblock \emph{Advances in Neural Information Processing Systems},
  33:\penalty0 9459--9474, 2020.

\bibitem[Mehta and Chilimbi(2024)]{amazon_rufus}
Rajiv Mehta and Trishul Chilimbi.
\newblock Amazon announces {Rufus}, a new generative {AI}-powered
  conversational shopping experience, 2024.
\newblock \href{https://www.aboutamazon.com/news/retail/amazon-rufus}{Link}.
  Accessed on May, 2025.

\bibitem[OpenAI(2024)]{searchgpt}
OpenAI.
\newblock Introducing {ChatGPT} search, 2024.
\newblock \href{https://openai.com/index/introducing-chatgpt-search/}{Link}.
  Accessed on May, 2025.

\bibitem[Reddit(2025)]{reddit2025answers}
Reddit.
\newblock Reddit answers (currently in beta), 2025.
\newblock
  \href{https://support.reddithelp.com/hc/en-us/articles/32026729424916-Reddit-Answers-Currently-in-Beta}{Link}.
  Accessed on May, 2025.

\bibitem[Reid(2024)]{google2024overview}
Elizabeth Reid.
\newblock Generative {AI} in search: Let google do the searching for you, May
  2024.
\newblock
  \href{https://blog.google/products/search/generative-ai-google-search-may-2024/}{Link}.
  Accessed on May, 2025.

\bibitem[Sch{\"u}tze et~al.(2008)Sch{\"u}tze, Manning, and
  Raghavan]{schutze2008introduction}
Hinrich Sch{\"u}tze, Christopher~D Manning, and Prabhakar Raghavan.
\newblock \emph{Introduction to information retrieval}, volume~39.
\newblock Cambridge University Press Cambridge, 2008.

\bibitem[Shapley(1953)]{shapley1953value}
Lloyd~S Shapley.
\newblock A value for n-person games.
\newblock \emph{Contribution to the Theory of Games}, 2, 1953.

\bibitem[Ye et~al.(2025)Ye, Yoganarasimhan, and Zheng]{ye2025lola}
Zikun Ye, Hema Yoganarasimhan, and Yufeng Zheng.
\newblock {LOLA}: Llm-assisted online learning algorithm for content
  experiments.
\newblock \emph{Forthcoming in Marketing Science}, 2025.

\end{thebibliography}
\end{document}